\documentclass{article}

\PassOptionsToPackage{square,numbers,sort&compress}{natbib}
\usepackage[nonatbib,final]{neurips_2020}

\usepackage[utf8]{inputenc} %
\usepackage[T1]{fontenc}    %
\usepackage{hyperref}       %
\usepackage{xurl}            %
\usepackage{booktabs}       %
\usepackage{amsfonts}       %
\usepackage{nicefrac}       %
\usepackage{microtype}      %
\usepackage{natbib}
\usepackage{graphicx}
\usepackage{subcaption}
\usepackage{amsmath}
\usepackage[dvipsnames]{xcolor}
\usepackage{enumitem}

\definecolor{correctcolor}{RGB}{5,113,176}
\definecolor{wrongcolor}{RGB}{202,0,32}

\title{Beyond accuracy: quantifying trial-by-trial behaviour of CNNs and humans by measuring error consistency}
\author{%
  Robert Geirhos\textsuperscript{$\ast$}\\
   University of Tübingen \& IMPRS-IS\\
   \texttt{robert.geirhos@uni-tuebingen.de}
   \And
   Kristof Meding\textsuperscript{$\ast$}\\
   University of Tübingen\\
   \texttt{kristof.meding@uni-tuebingen.de}
   \AND
   Felix A. Wichmann\\
   University of Tübingen\\
   \texttt{felix.wichmann@uni-tuebingen.de}
}

\begin{document}

\maketitle

\begin{abstract}
A central problem in cognitive science and behavioural neuroscience as well as in machine learning and artificial intelligence research is to ascertain whether two or more decision makers---be they brains or algorithms---use the same strategy. Accuracy alone cannot distinguish between strategies: two systems may achieve similar accuracy with very different strategies. The need to differentiate beyond accuracy is particularly pressing if two systems are at or near ceiling performance, like Convolutional Neural Networks (CNNs) and humans on visual object recognition.
Here we introduce trial-by-trial \emph{error consistency}, a quantitative analysis for measuring whether two decision making systems systematically make errors on the same inputs. Making consistent errors on a trial-by-trial basis is a necessary condition if we want to ascertain similar processing strategies between decision makers. Our analysis is applicable to compare algorithms with algorithms, humans with humans, and algorithms with humans.\\
When applying error consistency to visual object recognition we obtain three main findings: (1.) Irrespective of architecture, CNNs are remarkably consistent with one another. (2.) The consistency between CNNs and human observers, however, is little above what can be expected by chance alone---indicating that humans and CNNs are likely implementing very different strategies. (3.) CORnet-S, a recurrent model termed the ``current best model of the primate ventral visual stream'', fails to capture essential characteristics of human behavioural data and behaves essentially like a standard purely feedforward ResNet-50 in our analysis; highlighting that certain behavioural failure cases are not limited to feedforward models. Taken together, error consistency analysis suggests that the strategies used by human and machine vision are still very different---but we envision our general-purpose error consistency analysis to serve as a fruitful tool for quantifying future progress.
\end{abstract}

\section[Introduction]{Introduction\footnote{Blog post summary: \url{https://medium.com/@robertgeirhos/are-all-cnns-created-equal-d13a33b0caf7}}}
Complex systems are notoriously difficult to understand---be they Convolutional Neural Networks (CNNs) or the human mind or brain. Paradoxically, for CNNs, we have access to every single model parameter, know exactly how the architecture is formed of stacked convolution layers, and we can inspect every single pixel of the training data---yet understanding the behaviour emerging from these primitives has proven surprisingly challenging \cite{lillicrap2019does}, leaving us continually struggling to reconcile the success story of CNNs with their brittleness \cite{szegedy2013intriguing,ilyas2019adversarial, geirhos2020shortcut}.\footnote{Note again the parallel in neuroscience, even for very simple brains: The nervous system of the nematode \textit{C.elegans} is basically known in its entirety---
still it is not fully understood how the (comparatively) complex behaviour of \textit{C.elegans} is brought about by the biological ``hardware'' \cite{Mausfeld_2003a}.} In response to the need to better understand the internal mechanisms, a number of visualisation methods have been developed \cite{zeiler2014visualizing, zintgraf2017visualizing, olah2020zoom}. And while many of them have proven helpful in fuelling intuitions, some have later been found to be misleading \cite{nie2018theoretical, adebayo2018sanity}; moreover, most visualisation analyses are qualitative at nature. On the other hand, quantitative comparisons of different algorithms like benchmarking model accuracies have led to a lot of progress across deep learning, but reveal little about the internal mechanism: two models may reach similar levels of accuracy with very different internal processing strategies, an aspect that is gaining importance as CNNs are rapidly approaching ceiling performance across tasks and datasets. In order to understand whether two algorithms are implementing a similar or a different strategy, we need analyses that are quantitative \emph{and} allow for drawing conclusions about the internal mechanism.

\begin{figure}[t]
    \centering
    \begin{subfigure}{0.33\textwidth}
        \centering
        \includegraphics[width=\linewidth]{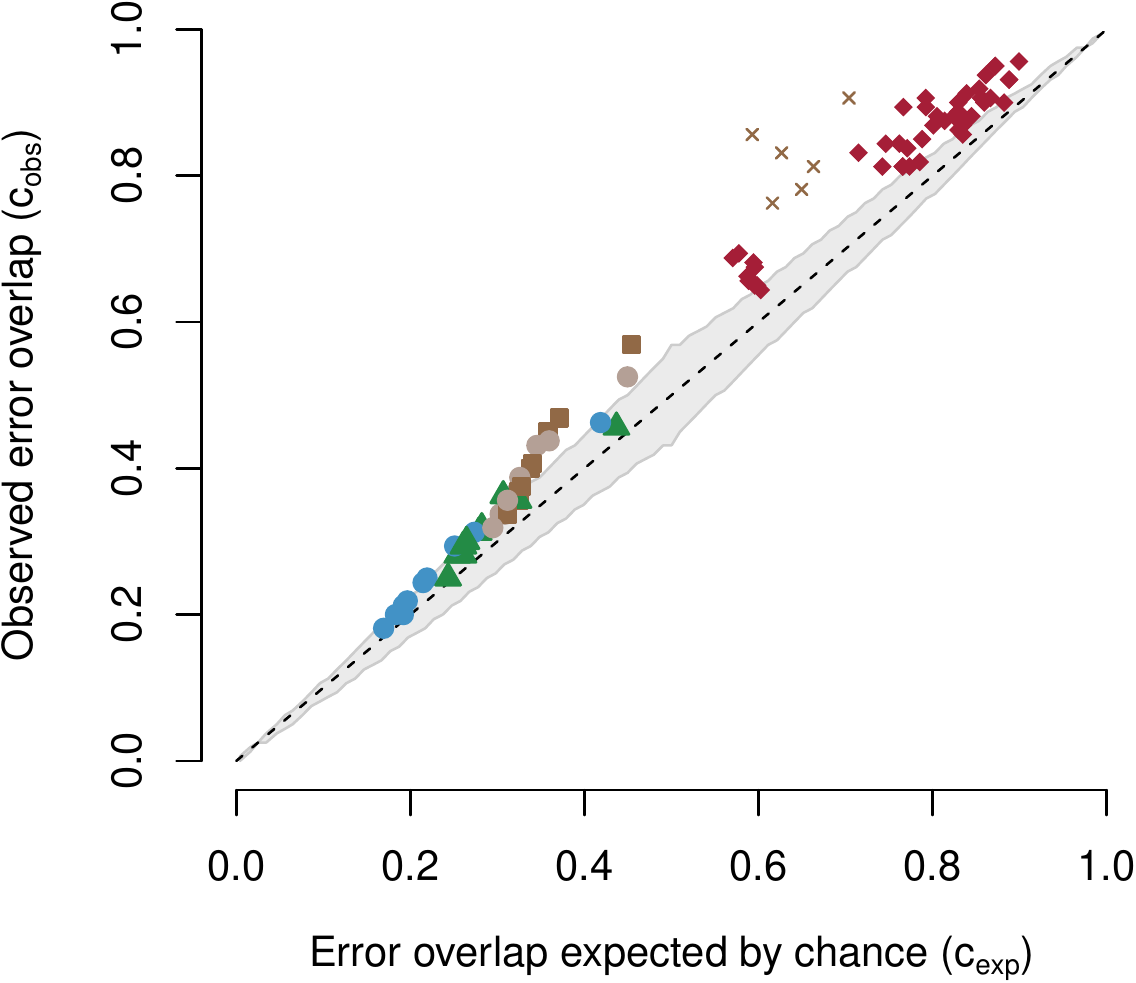}
        \caption{}
        \label{subfig:concept_a}
    \end{subfigure}%
    \begin{subfigure}{0.33\textwidth}
        \centering
        \includegraphics[width=\linewidth]{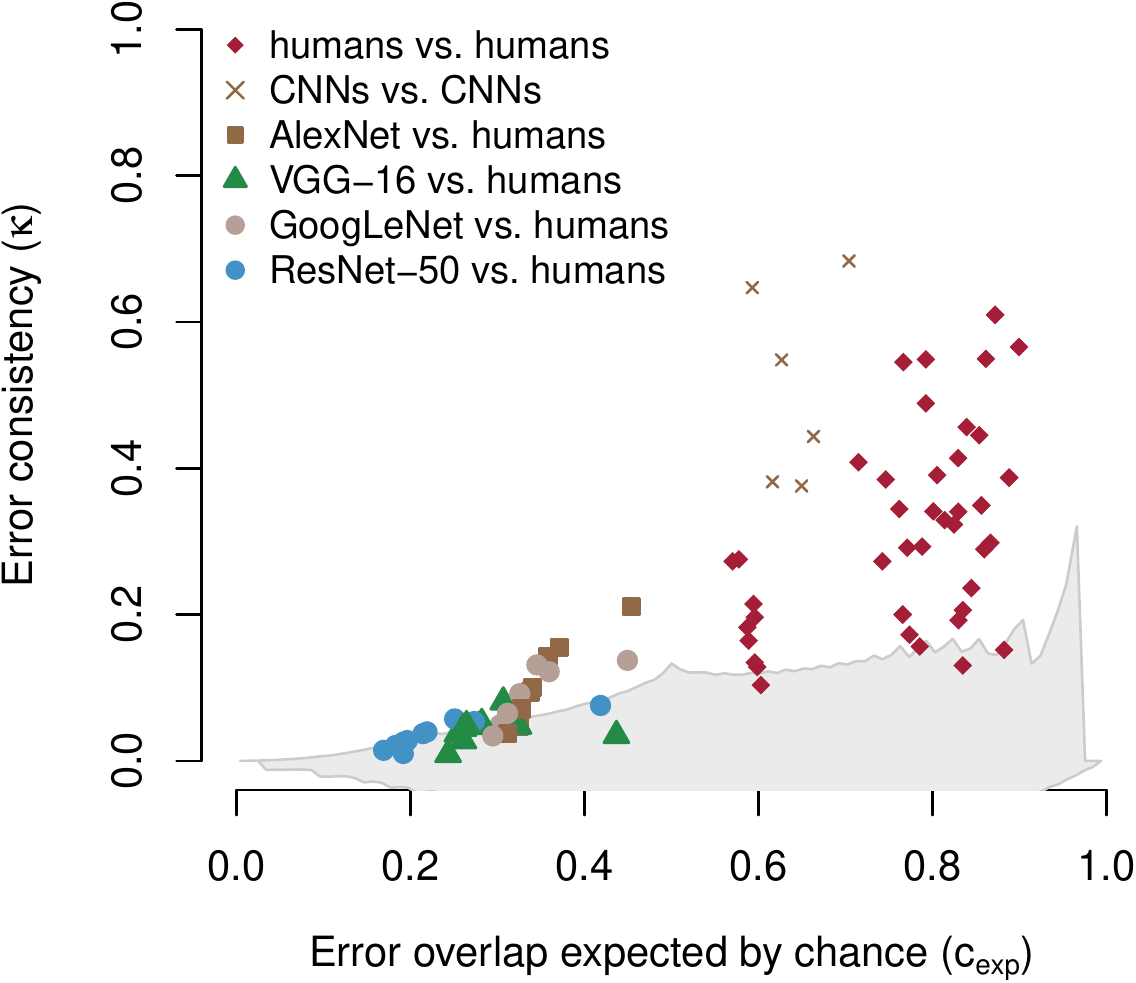}
        \caption{}
         \label{subfig:concept_b}
    \end{subfigure}
    \begin{subfigure}{0.33\textwidth}
    \centering
    \includegraphics[width=\linewidth]{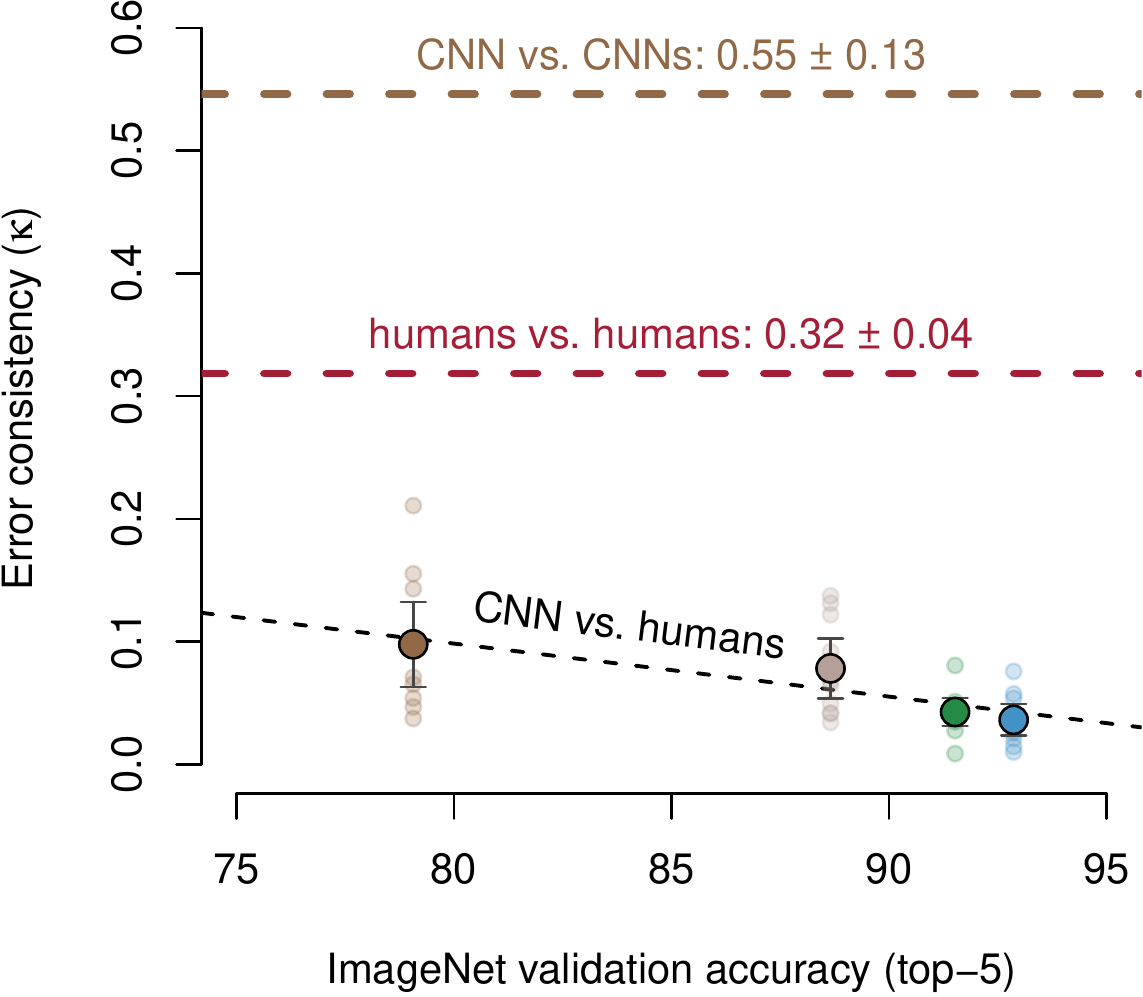}
    \caption{}
      \label{subfig:concept_c}
    \end{subfigure}
    \caption{Do humans and CNNs make consistent errors? From left to right three steps for analysing this question are visualised. For a detailed description of these steps please see the intuition~(\ref{subsec:intuition}). \textbf{(a)} Observed vs. expected error overlap (errors on the same trials) for a classification experiment where humans and CNNs classified the same images \cite{geirhos2019imagenettrained}. Values above the diagonal indicate more overlap than expected by chance. \textbf{(b)} Same data as on the left but measured by error consistency ($\kappa$). Higher values indicate greater consistency; shaded areas correspond to a simulated 95\% percentile for chance-level consistency. \textbf{(c)} Error consistency vs.\ ImageNet accuracy.}
    \label{fig:concept}
\end{figure}

We here introduce \emph{error consistency}\footnote{For a discussion of this terminology we refer to Section~\ref{app:sec:terminology} in the appendix}, a quantitative analysis for measuring whether two black-box perceptual systems systematically make errors on the same inputs. Irrespective of any potential differences at Marr's implementational level \cite{marr1982vision} (which may be quite large, e.g. between two different neural network architectures or even larger between a CNN and a human observer), one can only conclude that two systems use a similar strategy if these systems make similar errors: not just a similar number of errors (as measured by accuracy), but also errors on the same inputs, i.e.\ if two systems find the same \textit{individual} stimuli difficult or easy (as measured by error consistency). An agreement can be considered inverse to the Reichenbach-principle \cite{reichenbach1956direction} of correlation: correlation between variables does not imply a direct causal relationship. However, correlation does imply \emph{at least} an indirect causal link through other variables. For error consistency, zero error consistency implies that two decision makers are not using the same strategy.
While error consistency can be applied across fields, tasks and domains (including vision, auditory processing, etc.), we believe it to be of particular relevance at the intersection of deep learning, neuroscience and cognitive science. Both brains and CNNs have, at various points, been described as black-box mechanisms \cite{castelvecchi2016can, shwartz2017opening, kietzmann2018deep}. But do the spectacular advances in deep learning shed light on the perceptual and cognitive processes of biological vision? Does similar performance imply similar mechanism or algorithm? Do different CNNs indeed make different errors?\footnote{\cite{geirhos2020on} found surprising similarities for self-supervised vs.\ supervised CNNs using error consistency.} We believe that fine-grained analysis techniques like error consistency may serve an important purpose in this debate.

\textbf{Molecular psychophysics.} Analysing errors for every single input is inspired by the idea of ``molecular psychophysics'' by David Green \cite{Green_1964}. He argued that the goal of psychophysics should be to predict human responses to individual stimuli (trials) and not only aggregated responses (accuracy), let alone only averages across many individuals, as is common in much of the behavioural sciences. Green also predicted that once models of perceptual processes became more advanced, accuracy would cease to be a good criterion to assess and compare them rigorously (see p.~394 in  \cite{Green_1964}).

\textbf{Related work.} 
Using error consistency we can analyse human and CNN error patterns in a way that has, we believe, not been done before. We obtain \emph{novel findings} but we do not consider error consistency to be an entirely \emph{novel method} by itself. Instead, it builds on, extends and adapts existing methods and ideas developed in three different fields: molecular psychophysics (as described above) as well as causal inference and the social sciences (as described below). Our goal is the systematic analysis of human and CNN error patterns at the trial-by-trial level. Many previous analyses have focused on the aggregated level instead: In machine learning, performance is predominantly measured by accuracy and existing metrics to analyse errors such as comparisons between confusion matrices \cite{ghodrati2014feedforward, rajalingham2015comparison, kheradpisheh2016deep, kheradpisheh2016humans, geirhos2017comparing} or scores based on KL divergence \cite{ma2020a} pool over single trials, thereby losing crucial information---they are not ``molecular'' but only ``molar'' in Green's terminology \cite{Green_1964}. \cite{kubilius2016deep, rajalingham2018large} went an important step further by comparing errors at an image-by-image level, but consistency was only computed \emph{after} aggregating across participants, and \cite{kubilius2016deep} use a metric that automatically leads to higher consistency when comparing two systems with higher accuracy (without discounting for consistency due to chance). Closely related to our analysis is \cite{mania2019model}, who investigated similarity between models in the context of overfitting. In the context of causal inference, \cite{Meding2019} performed a trial-by-trial analysis, plotting expected vs.\ observed behaviour (a starting point for our analysis). In social sciences, psychology and medicine, comparisons between participants are common, e.g. for problems like ``How do people differ when answering a questionnaire?''. In that context, so-called inter-rater agreement is measured by Cohen's kappa \cite{Cohen1960}. Here we repurpose and extend Cohen's kappa ($\kappa$) for the analysis of classification errors by humans and machines, and provide confidence intervals and analytical bounds (limiting possible consistency).

\textbf{Terminology.} A \emph{decision maker} is any (living or artificial) entity that implements a decision rule. A \emph{decision rule} is a function that defines a mapping from input to output (see \cite{geirhos2020shortcut} for a taxonomy of decision rules). Note that the same decision rule can result from different strategies. We use the term \emph{strategy} synonymously with the term \emph{algorithm}. For instance, \texttt{Quicksort(X)} and \texttt{Mergesort(X)} use a different algorithm (strategy), but they implement the same decision rule: the output will always be the same. \texttt{Permute(X)}, on the other hand, will (usually) lead to a different output. Hence, similar output (or similar errors, i.e., high error consistency) is a necessary, but not a sufficient condition for similar strategies.

\subsection{Intuition}
\label{subsec:intuition} Before going through the mathematical details in Section~\ref{section:methods}, let us consider a simple example of a psychophysical experiment where human observers and CNNs classified objects from 160 images (line drawing / edge-like stimuli in this case). There are three steps in order to analyse error consistency (visualised in Figure~\ref{fig:concept}). We can start by analysing how many of the decisions (either correct or incorrect) to individual trials are identical (\emph{observed error overlap}). This number only becomes meaningful when plotted against the \emph{error overlap expected by chance} (Figure~\ref{subfig:concept_a}): for instance, two observers with high accuracies will necessarily agree on many trials by chance alone. However, this visualisation may be hard to interpret since higher values do not simply correspond to higher consistency (instead, above-chance consistency is measured by distance from the diagonal). In a second step, we can therefore normalise the data (Figure~\ref{subfig:concept_b}) by dividing each datapoint's distance to the diagonal by the total distance between the diagonal and ceiling (1.0). Now, we can directly compare the error consistency between decision makers: if error consistency is measured by $\kappa$, then $\kappa=0$ means chance-level consistency (independent processing strategies), $\kappa>0$ indicates consistency beyond chance (similar strategies) and $\kappa<0$ inconsistency beyond chance (inverse strategies). Lastly, we can analyse the relationship between error consistency ($\kappa$) and an arbitrary other variable, for instance in order to determine whether better ImageNet accuracy leads to higher consistency between a CNN and human observers (Figure~\ref{subfig:concept_c}), which is not the case here.

\section{Methods}
\label{section:methods}
When comparing two decision makers the most obvious comparison is accuracy. Our goal is to go beyond accuracy per se by assessing the consistency of the responses with respect to individual stimuli. As a prerequisite, all decision makers need to evaluate the exact same stimuli. The order of presentation is irrelevant as long as the responses can be sorted w.r.t. stimuli afterwards.\footnote{For human observers the order of presentation can make a (typically small) difference as human observers exhibit serial dependencies and other non-stationarities \cite{Green_1964,Frund_2014}. Participants, e.g., may make more errors or lapses towards the end of an experiment due to fatigue \cite{wichmann_psychometric_2001a}), and it is thus recommended to randomly shuffle presentation order for each participant to avoid such a ``trivial'' consistency of errors. Luckily, non-stationarities are usually only problematic if the signal levels are low, i.e.\ near chance performance.} In the following, we show how error consistency can be computed and which bounds and confidence intervals apply for the observed error overlap (\ref{subsec:methods_c_obs}) and for $\kappa$ (\ref{subsec:methods_kappa}). Experimental methods are described in~\ref{meth:experimental_methods} and code is available from \url{https://github.com/wichmann-lab/error-consistency}.

\subsection{Observed vs.\ expected error overlap}
\label{subsec:methods_c_obs}
If two observers $i$ and $j$ (be they algorithms, humans or animals) respond to the same $n$ trials, we can investigate by how much their decisions overlap. For this purpose, we only analyse whether the decisions were correct/incorrect (irrespective of the number of choices). The observed error overlap $c_{obs}$ is defined as 
$c_{obs_{i,j}} = \frac{e_{i,j}}{n}$ where $e_{i,j}$ is the number of equal responses (either both correct or both incorrect). In order to find out whether this observed overlap is beyond what can be expected by chance, we can compare observers $i$ and $j$ to a theoretical model: independent binomial observers (binomial: making either a correct or an incorrect decision; independent: only random consistency). In this case, we can expect only overlap due to chance $c_{exp_{i,j}}$:
\begin{align}
    c_{exp_{i,j}} = \textcolor{correctcolor}{p_i p_j} + \textcolor{wrongcolor}{(1-p_i) (1-p_j)}. \label{eq:cexp}
\end{align}

This is the sum of the probabilities that two observers $i$ and $j$ with accuracies $p_i$ and $p_j$ give the same \textcolor{correctcolor}{correct} and \textcolor{wrongcolor}{incorrect} response by chance.\footnote{Note that $c_{exp}>0.5\iff p_1, p_2 > 0.5 \lor p_1, p_2 < 0.5$, see also Figure~\ref{fig:DensityCobs} in the appendix.}

\textbf{Confidence intervals.}
Unfortunately, the confidence interval of \(c_{obs_{i,j}}\) in the scatter-plot of Figure~\ref{subfig:concept_a} is not trivial to obtain. \cite{Meding2019} used a standard binomial confidence interval. This is, however, only a very rough estimate of the true confidence interval since the position on the x-axis ($c_{exp}$) itself is also estimated from the data and thus influenced by variation. We sample data for the null hypothesis of independent observers and calculate the corresponding 95\% percentiles (cf.~Figure~\ref{fig:CI_and_bounds}). This process is described in Section~\ref{app:CI_kappa} in the appendix.

\textbf{Bounds.}
Confidence intervals allow to investigate hypotheses. In addition, theoretical bounds might help to assess the degree of the observed consistency not being due to chance: a data point close or at the bound has maximum distance to the diagonal for a given value of \(c_{exp}\). For this end we have calculated bounds of \(c_{obs}\) as an additional diagnostic tool. The influence of these bounds on the confidence intervals is visualised in Figure~\ref{fig:CI_and_bounds}.

Ideally, we also want to express the bounds of $c_{obs}$ directly as a function of $c_{exp}$. The analytical derivation of the bounds below can be found in the Appendix (\ref{app:derivation_bounds_cobs}) and are visualised in Figure~\ref{fig:CI_and_bounds}.
\begin{align}
0 &\leq c_{obs_{i,j}} \leq 1 - \sqrt{1 -  2 c_{exp_{i,j} }} ~~~~  &\text{if}~~ c_{exp_{i,j}}  \leq 0.5, \\
\sqrt{2 c_{exp_{i,j}}-1} &\leq c_{obs_{i,j}} \leq 1 ~~~~ &\text{if}~ ~c_{exp_{i,j}}  \geq 0.5.
\end{align}

\begin{figure}[t]
    \centering
    \begin{subfigure}{0.75\textwidth}
        \centering
 \includegraphics[width=1\linewidth]{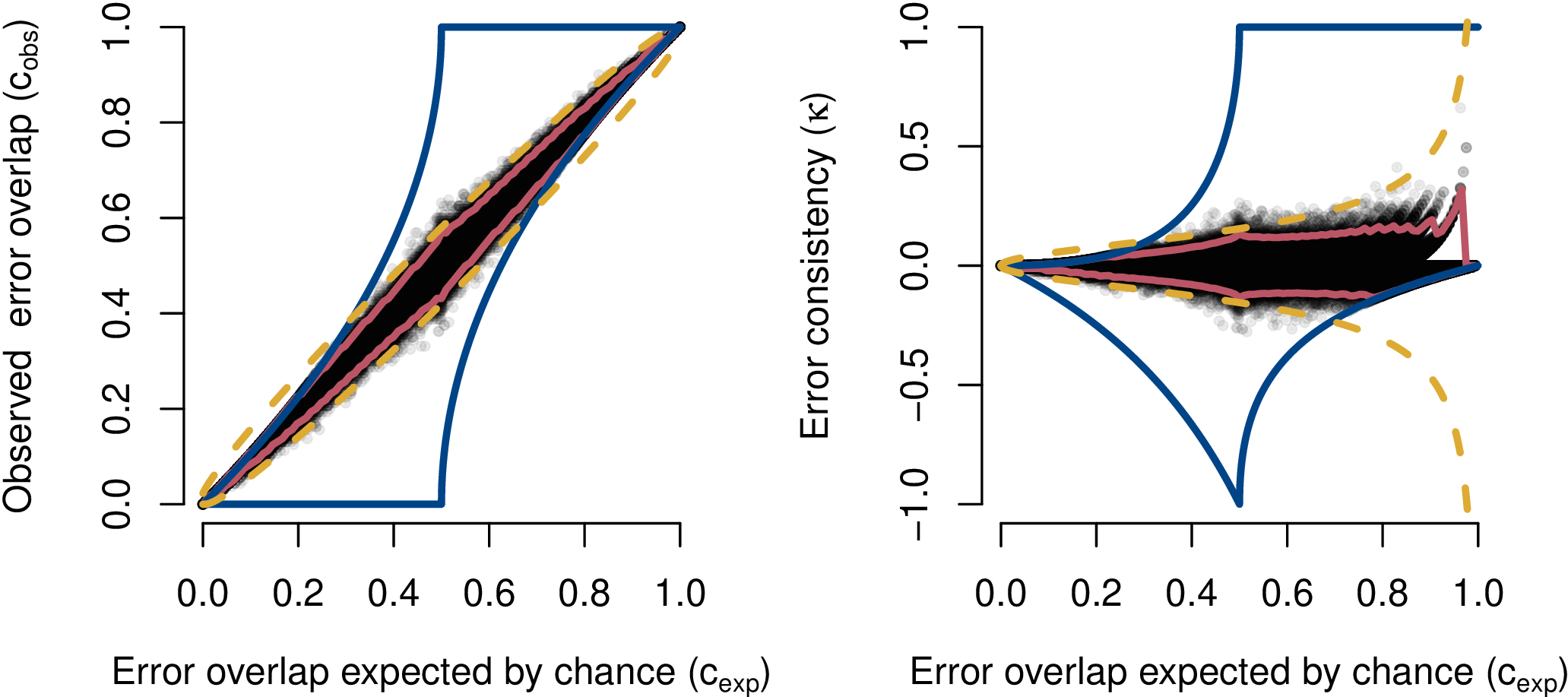}
    \end{subfigure}
    \begin{subfigure}{0.23\textwidth}
        \centering
        \includegraphics[width=1\linewidth]{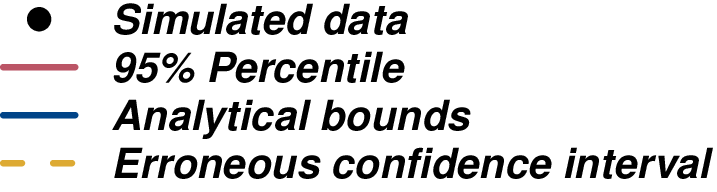}
    \end{subfigure}\hfill
   \caption{Simulated data of \(c_{exp},c_{obs}\) and $\kappa$ for 160 trials under the assumption of independent decision makers. Analytical bounds and 95\% percentile derived from the simulation of 100,000 experiments do not align with the often reported erroneous confidence interval.}
\label{fig:CI_and_bounds}
\end{figure}

\subsection{Error consistency measured by Cohen's kappa}
\label{subsec:methods_kappa}
$c_{obs_{i,j}}$ described above quantifies the observed error overlap between observers $i$ and $j$. In order to obtain a single behavioural score for error consistency, that is, one disentangled from accuracy\footnote{In fact, error consistency is an accuracy corrected metric, see \ref{sec::accvskappa} in appendix} , we need to discount for error overlap by chance $c_{exp_{i,j}}$. This is solved by Cohen's $\kappa$ \cite{Cohen1960} with which we measure error consistency:
\begin{align}
    \kappa_{i,j} = \frac{c_{obs_{i,j}}-c_{exp_{i,j}}}{1-c_{exp_{i,j}}}. \label{eq:kappa}
\end{align}
We do not include a comparison of $\kappa$ to the (Pearson) correlation coefficient since it has been shown that correlation is not a suitable measure of agreement \cite{Hunt1986,watson2010method}.

\textbf{Confidence intervals.}
Confidence intervals of the average $\kappa$ of groups, such as the average error consistency of humans vs.\ humans in Figure~\ref{subfig:concept_c}, are based on the empirical standard error of the mean and a normal distribution assumption of the average error consistency (a numerical simulation of binomial observers confirmed that this assumption is valid here). Analogous to the observed consistency we use a sampling approach to obtain confidence intervals of $\kappa$ given $c_{exp}$, see~\ref{app:CI_kappa} for details. This is necessary since the original confidence approximation interval derived by Cohen \cite{Cohen1960} (yellow dashes for error consistency in Figure~\ref{fig:CI_and_bounds}) were later shown to be erroneous \cite{Fleiss1969,Hudson1987}.\footnote{This erroneous confidence interval is still used in many publications, including very influential ones \cite{Mchugh2012, Bland2015}.} While a corrected approximate version for individual kappas does exist \cite{Fleiss1969,fleiss2003}, there is to our knowledge no analytical or approximate confidence interval for $\kappa$ given $c_{exp}$, and hence our sampling approach.

\textbf{Bounds.}
The following bounds show the limits of $\kappa$ given a specific value of \(c_{exp}\), please see Section~\ref{app:derivation_bounds_cobs} for the derivation and Figure~\ref{fig:CI_and_bounds} for visualisation\footnote{Bounds of kappa depending on $c_{obs}$ instead of $c_{exp}$ can be found in \cite{Umesh1989}.}:

\begin{align}
   \frac{-c_{exp_{i,j}}}{1-c_{exp_{i,j}}} &\leq \kappa_{i,j} \leq \frac{1-\sqrt{1 - 2 c_{exp_{i,j}}} - c_{exp_{i,j}}}{1- c_{exp_{i,j}}} ~~~~  & \text{if } c_{exp_{i,j}}  \leq 0.5, \\
\frac{\sqrt{2 c_{exp_{i,j}} -1} - c_{exp_{i,j}}}{1- c_{exp_{i,j}}} &\leq \kappa_{i,j} \leq 1 ~~~~ & \text{if } c_{exp_{i,j}}  \geq 0.5.
\end{align}

\subsection{Experimental methods}
\label{meth:experimental_methods}

\textbf{Stimuli: motivation.} Exemplary stimuli are visualised in Figure~\ref{fig:results_accuracy_generalisation}. We tested both ``vanilla'' images (plain unmodified colour images from ImageNet \cite{russakovsky2015imagenet}) and three different types of out-of-distribution (o.o.d.) images. The motivation for using o.o.d.\ images is the following: Significant progress in neuroscience---e.g., discovering receptive fields of simple and complex cells---was made using ``unnatural'' bar-like stimuli. In deep learning, adversarial examples and texture bias were discovered by testing models on (unnatural) images different than the training data. Hence, we can learn a lot about the inner workings of a system by probing it with appropriate ``artificial'' stimuli \cite{rust2005praise, martinez2019praise}; \cite{geirhos2020shortcut} even argues that o.o.d.\ testing is a necessity for drawing reliable inferences about a model's strategy. Standard ImageNet images (where human and pre-trained CNN accuracies are both very high and similar, $.960\pm .036\%$) are included as a baseline condition.

\textbf{Stimuli: method details.}
\cite{geirhos2019imagenettrained} tested N=10 human observers in their cue conflict, edge and silhouette experiments. Starting from normal images with a white background, different image manipulations were applied. For \emph{cue conflict} images, the texture of a different image was transferred to this image using neural style transfer \cite{Gatys2016}, creating a texture-shape cue conflict with a total of 1280 trials per observer and network. For \emph{edge} stimuli, a standard edge detector was applied to the original images to obtain line-drawing-like stimuli (160 trials per observer). \emph{Silhouette} stimuli were created by filling the outline of an object with black colour, leaving just the silhouette (160 trials per observer).\footnote{For parametrically distorted images (Appendix, Figure~\ref{fig:app_noise_generalisation}) we used the stimuli from \cite{geirhos2018generalisation}.} Lastly, \emph{ImageNet} stimuli were standard coloured ImageNet images; we used the behavioural data (N=2 observers) and stimuli from \cite{geirhos2018generalisation} for this experiment.

\textbf{Paradigm.}
In order to compare the error consistency of two perceptual systems (e.g. CNNs and humans), those two systems a) need to be evaluated on the exact same stimuli and b) need to be in a regime with neither perfect accuracy nor chance-level performance. We found the publicly available stimuli and data from \cite{geirhos2019imagenettrained} to be an ideal test case. \cite{geirhos2019imagenettrained} compared object recognition abilities of humans and algorithms in a carefully designed psychophysical experiment. After a 200 ms presentation of a $224\times224$ pixels image, observers had 16 categories to choose from (e.g. \texttt{car}, \texttt{dog}, \texttt{chair}). For ImageNet-trained networks, categorisation responses for 1,000 fine-grained classes were mapped to those 16 classes using the WordNet hierarchy \cite{Miller1995}. In order to obtain the probability of a broad category (e.g.\ \texttt{dog}), response probabilities of all corresponding fine-grained categories (e.g.\ all ImageNet dog breeds) were averaged using the arithmetic mean.\footnote{This aggregation is optimal. A derivation is included in the appendix of the arXiv version \texttt{v3} of \cite{geirhos2018generalisation}.}

\textbf{Convolutional Neural Networks.}
Human responses were compared against classification decisions of all available CNN models from the PyTorch model zoo (for \texttt{torchvision} version 0.2.2) and against a recurrent model, CORnet-S \cite{kubilius2019brain}. All CNNs were trained on ImageNet. Details here:~\ref{app:method_details_CNNs}. Additionally, we analysed the relationship between model shape bias (induced by training on Stylized-ImageNet) and error consistency with human observers:~\ref{app:consistency_shape_biased_models}.

\section{Results}
\label{section:results}
If two perceptual systems or decision makers implement the same strategy they can be expected to systematically make errors on the same stimuli. In the following, we show how \emph{error consistency} can be used within visual object recognition to compare algorithms with humans (Section~\ref{subsec:algorithms_vs_humans}) and algorithms with algorithms (Section~\ref{subsec:algorithms_vs_algorithms}).

\subsection{Comparing algorithms with humans: investigating whether better ImageNet models show higher error consistency with human behavioural data}
\label{subsec:algorithms_vs_humans}

In deep learning, there is a strong linear relationship between ImageNet accuracy and transfer learning performance \cite{kornblith2019better}; in computational neuroscience, better categorisation accuracy improves the prediction of neural firing patterns \cite{yamins2014performance}. But do better performing ImageNet models also make more human-like errors?

\textbf{Error consistency vs.\ model performance.}
In Figure~\ref{fig:results_accuracy_generalisation}, we analyse the error consistency between human observers and sixteen standard ImageNet-trained CNNs. We find that humans to humans show a fair degree of consistency w.r.t. individual stimuli. That is, their agreement on which cats or chairs or cars are easy/hard to categorise is well beyond chance. Interestingly, CNN-to-CNN consistency is even higher than human-to-human consistency in all three experiments. This occurs despite the fact that human accuracies are higher than CNN accuracies across experiments: for instance in the silhouette experiment, the average human accuracy is 0.75 whereas the average CNN accuracy is 0.54 (see Table~\ref{app:table_accuracies}, supplementary information). However, the consistency between CNNs and humans is close to zero for two experiments (cue conflict stimuli and line drawings); a linear model fit indicates no improvement with better ImageNet validation accuracy: $F(1, 158) = 0.086, p=.769, R^2=0.001$ for cue conflict and $F(1, 158) = 0.478, p=.491, R^2=0.003$ for line drawing stimuli. For silhouettes, there is a significant \emph{positive} relationship between ImageNet accuracy and error consistency with $F(1, 158) = 53.530, p=1.21\cdot 10^{-11}, R^2=0.253$; for ImageNet images, on the other hand, there is a significant \emph{negative} relationship between top-5 accuracy and error consistency with $F(1, 30) = 8.162, p=.008, R^2=0.214$.

\newcommand{\figwidthI}{0.5\textwidth} %
\newcommand{\figwidthII}{0.28\textwidth} %
\newcommand{\captionspaceGeneralisation}{-1.3\baselineskip} %
\newcommand{\captionspaceGeneralisationII}{-0.1\baselineskip} %
\begin{figure}[hp]
    \vspace{-0.3cm}
    \begin{subfigure}{\figwidthI}
        \centering
        \includegraphics[width=\linewidth]{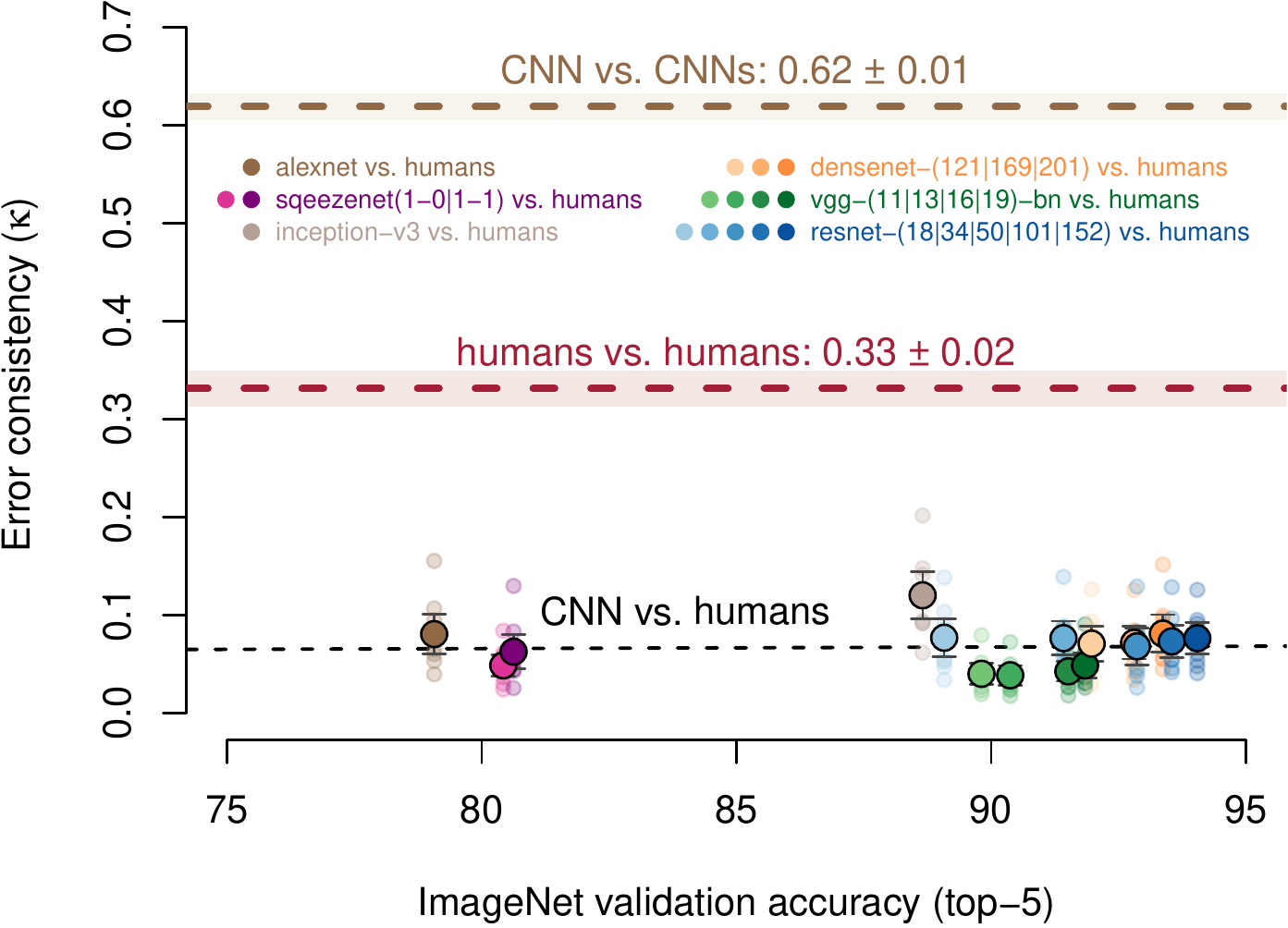}
        \vspace{\captionspaceGeneralisationII}
    \end{subfigure}\hfill
    \begin{subfigure}{\figwidthII}
        \centering
        \textbf{`cue conflict' stimuli}
        \raisebox{40pt}{
        \includegraphics[width=\linewidth]{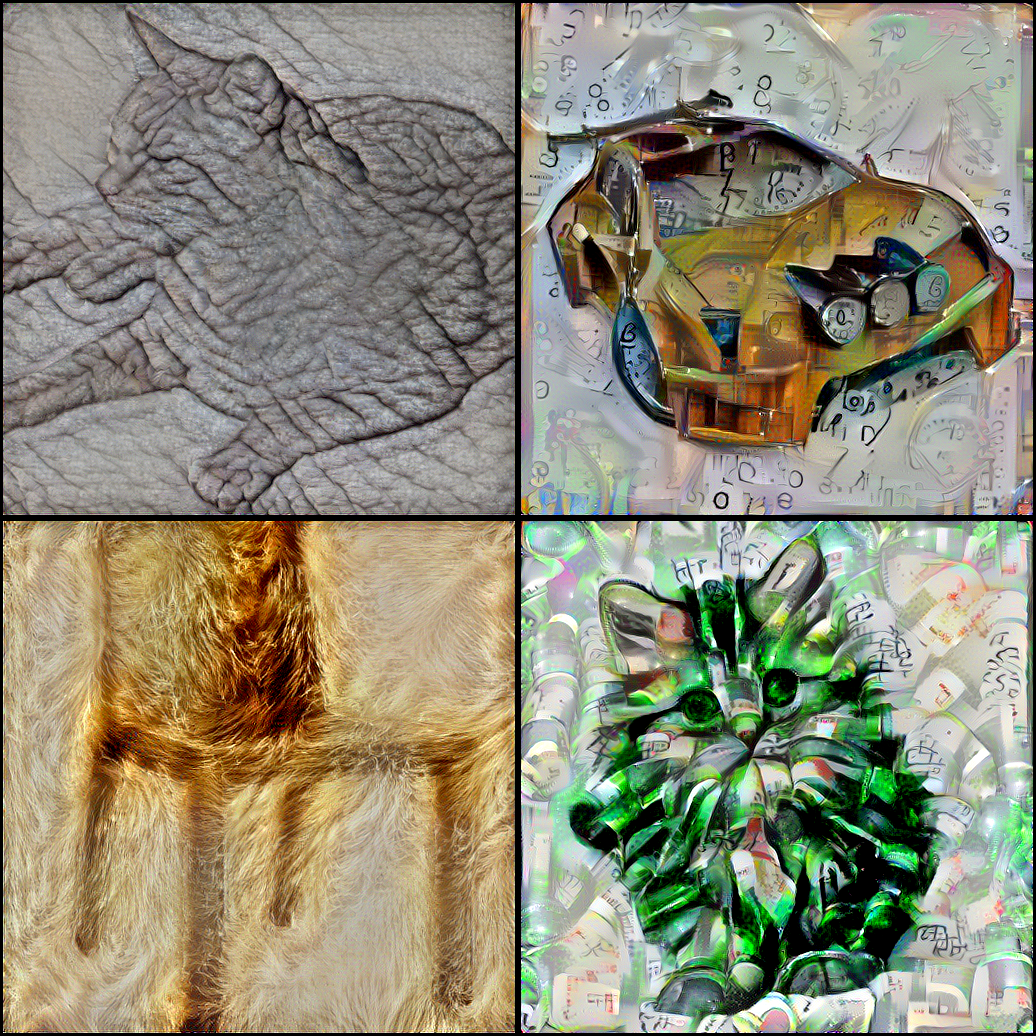}}
    \end{subfigure}
    \vspace{-0.6cm}

    \begin{subfigure}{\figwidthI}
        \centering
        \includegraphics[width=\linewidth]{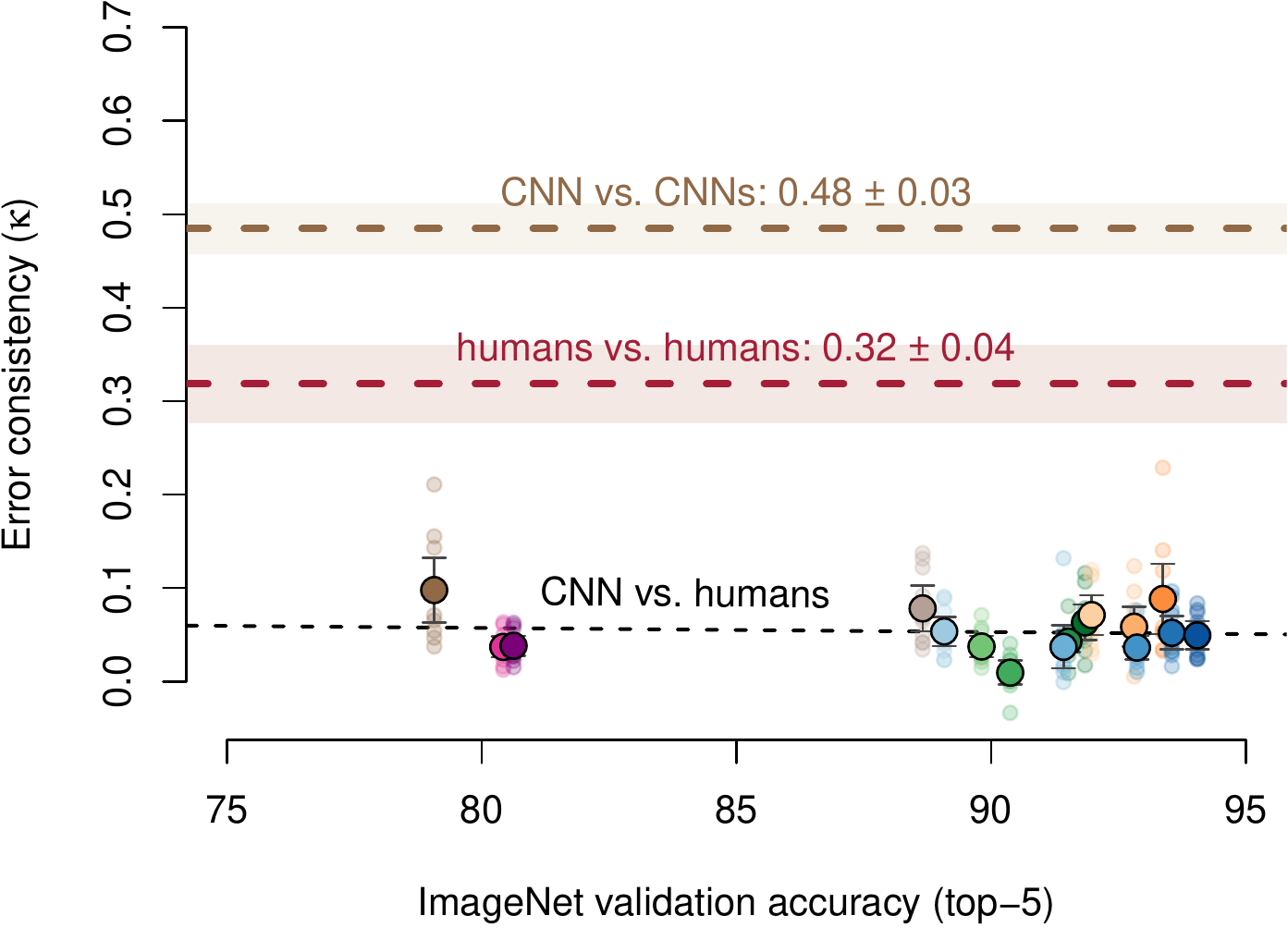}
        \vspace{\captionspaceGeneralisationII}
    \end{subfigure}\hfill
    \begin{subfigure}{\figwidthII}
        \centering
        \textbf{`edge' stimuli}
        \raisebox{40pt}{\includegraphics[width=\linewidth]{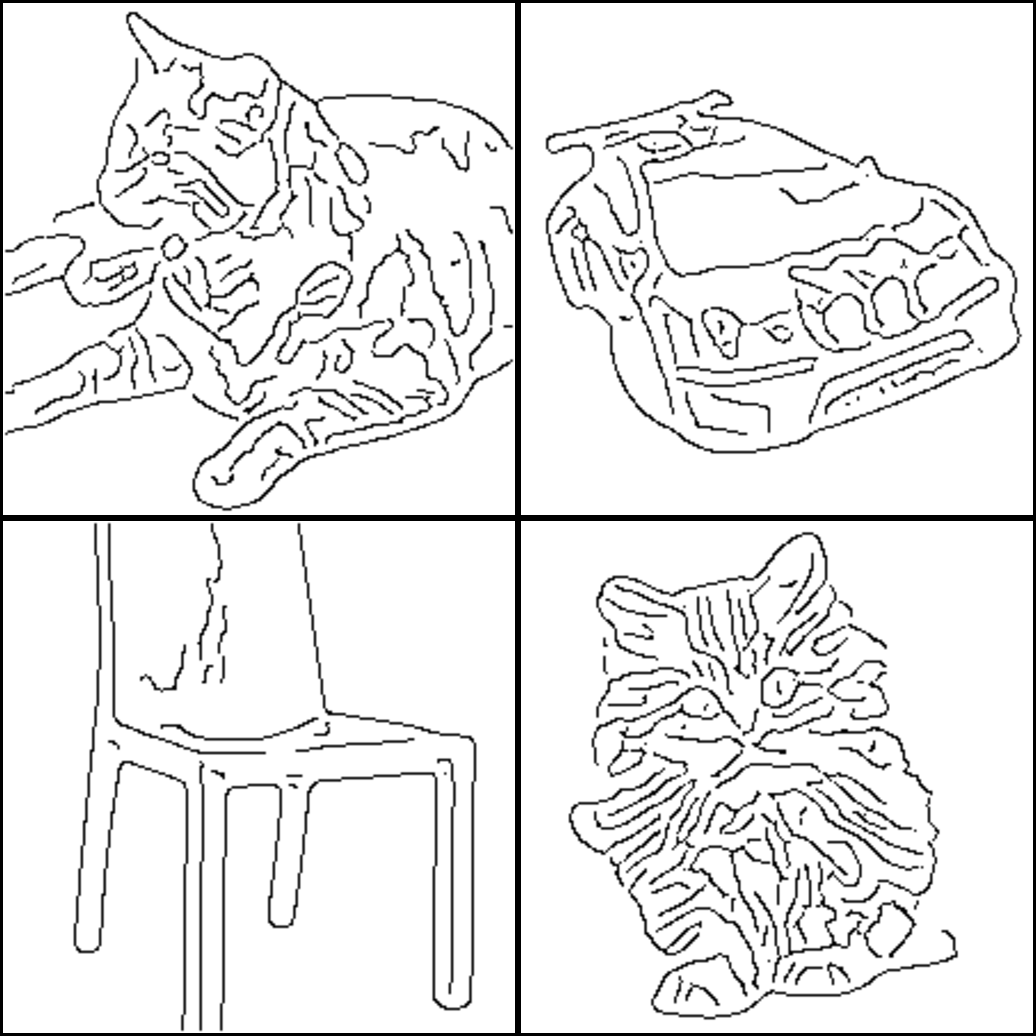}}
    \end{subfigure}
    \vspace{-0.1cm}

    \begin{subfigure}{\figwidthI}
        \centering
        \includegraphics[width=\linewidth]{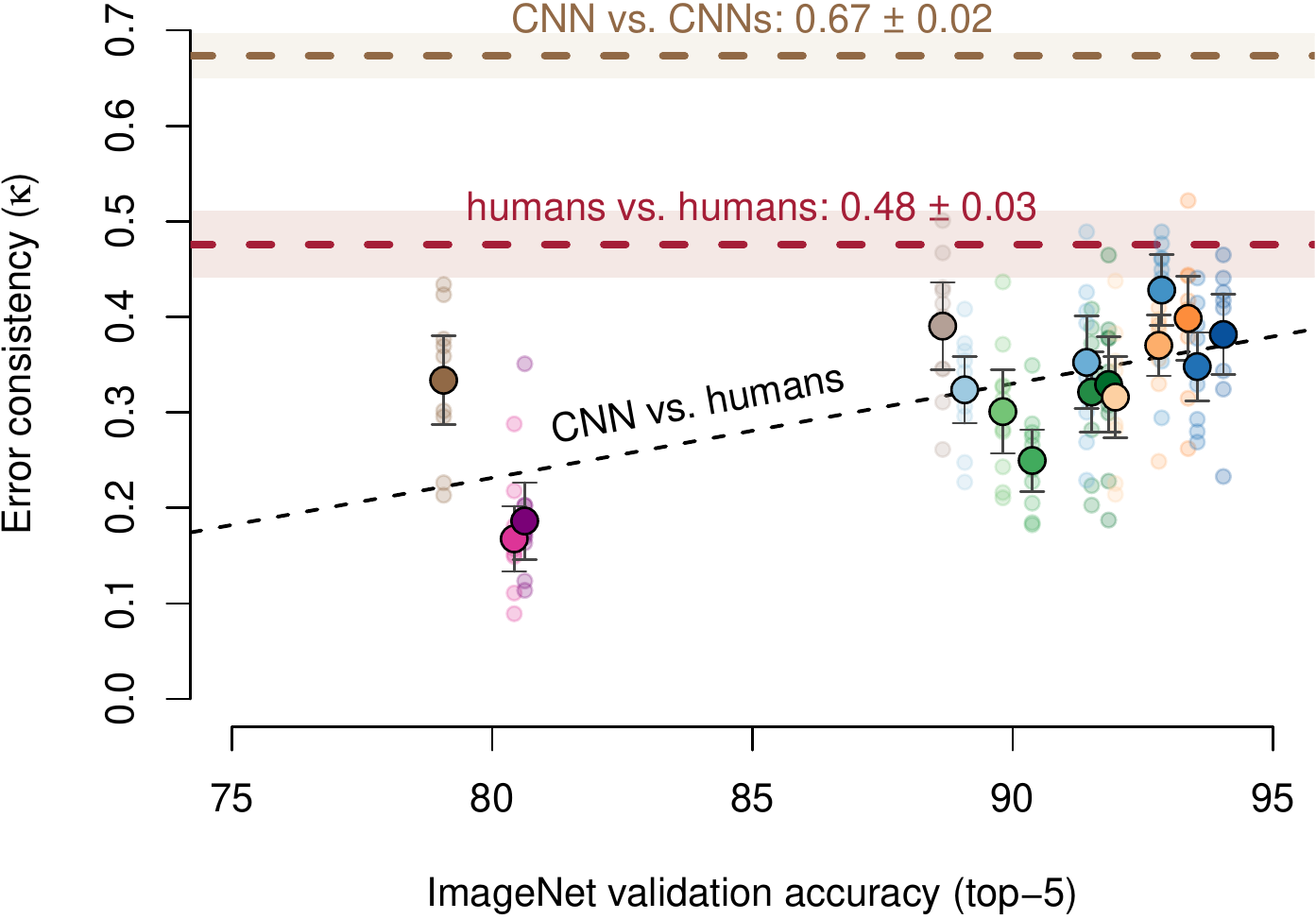}
        \vspace{\captionspaceGeneralisationII}
    \end{subfigure}\hfill
    \begin{subfigure}{\figwidthII}
        \centering
        \textbf{`silhouette' stimuli}
        \raisebox{35pt}{\includegraphics[width=\linewidth]{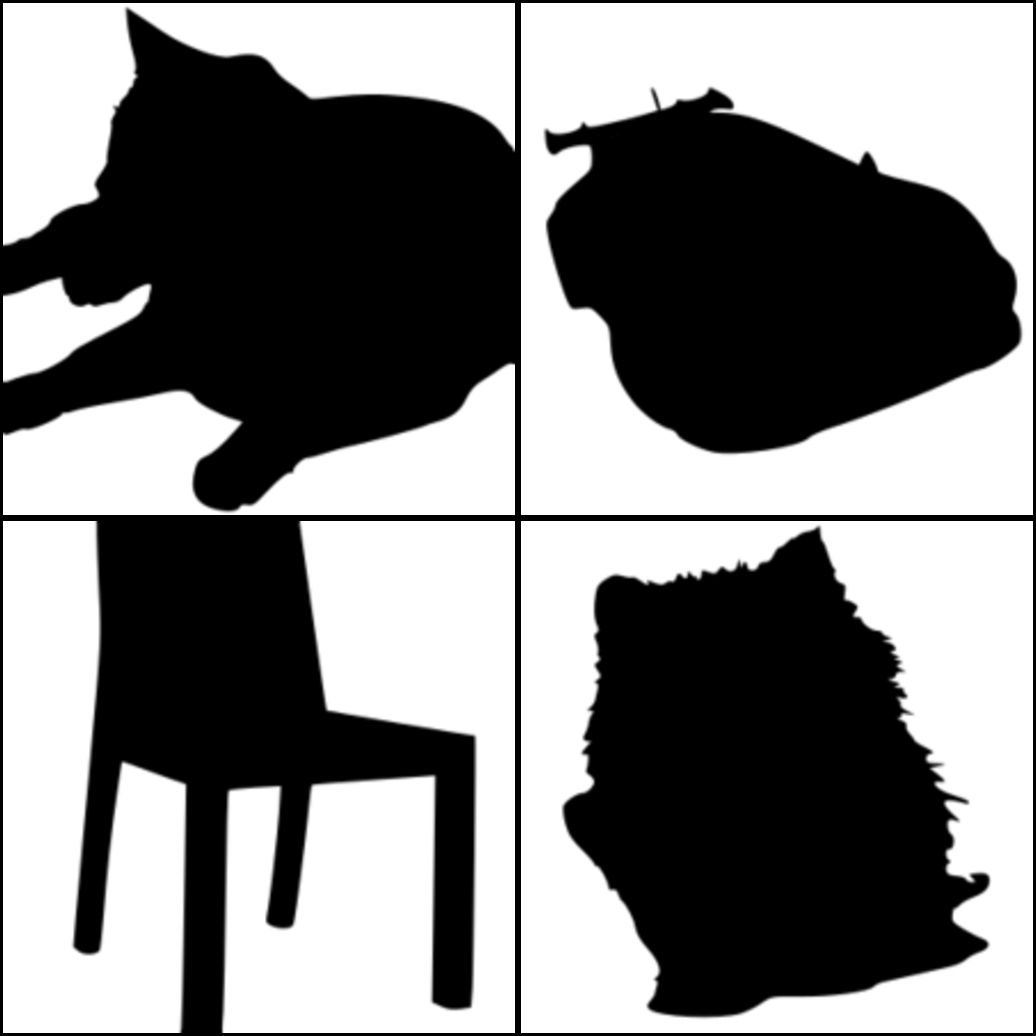}}
    \end{subfigure}
    \vspace{-0.3cm}

    \begin{subfigure}{\figwidthI}
        \centering
        \includegraphics[width=\linewidth]{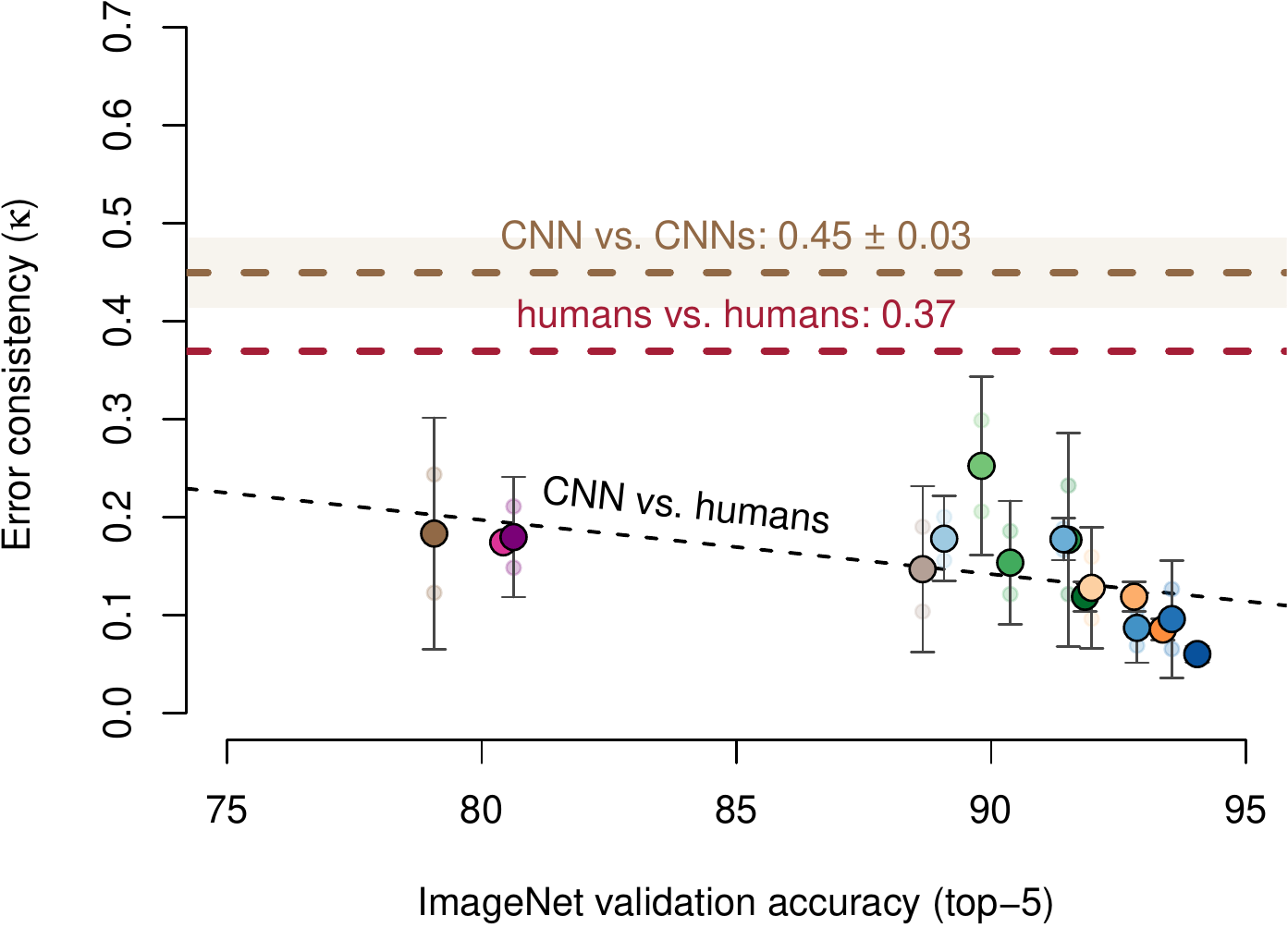}
        \vspace{\captionspaceGeneralisationII}
    \end{subfigure}\hfill
    \begin{subfigure}{\figwidthII}
        \centering
        \textbf{`ImageNet' stimuli}
        \raisebox{35pt}{\includegraphics[width=\linewidth]{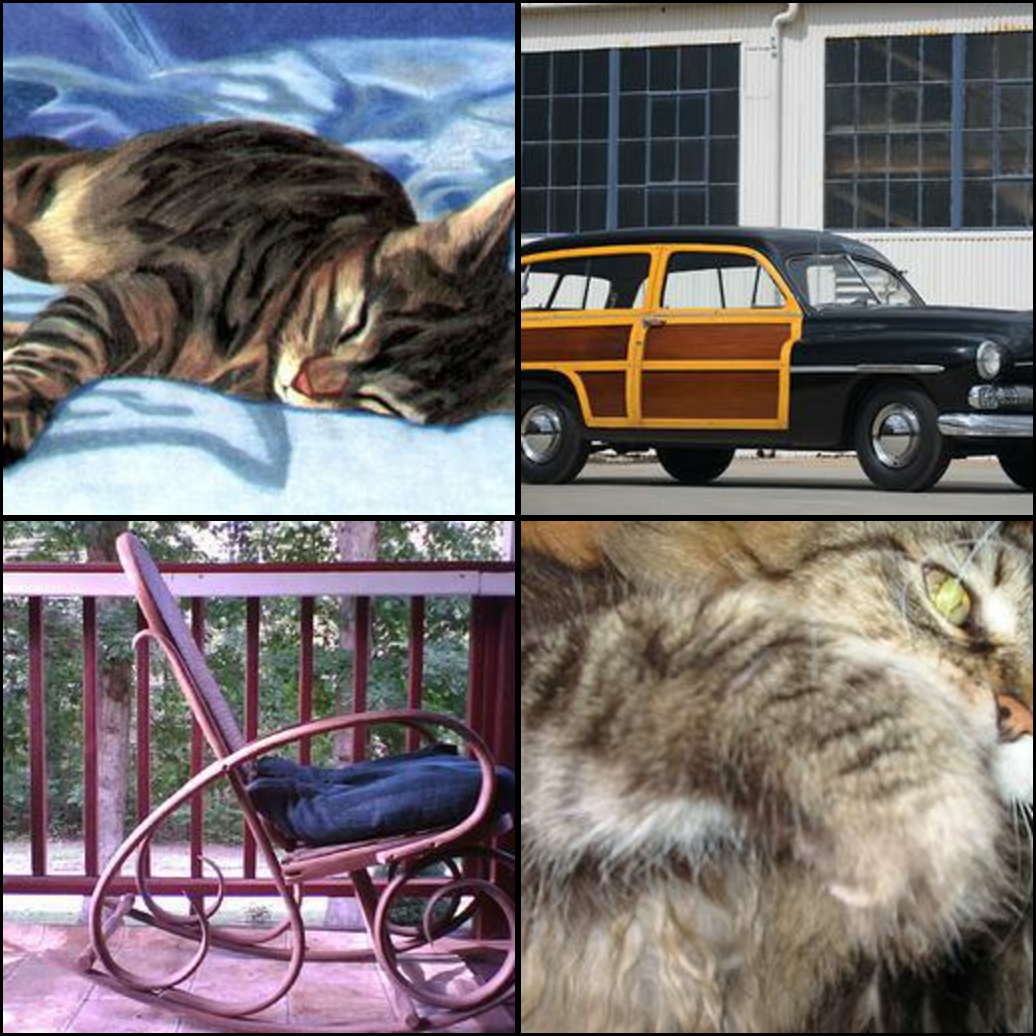}}
    \end{subfigure}
    \vspace{-0.1cm}
    \caption{Do better ImageNet models make more human-like errors? Error consistency vs.\ top-5 ImageNet validation accuracy for four experiments: cue-conflict, edges, silhouettes and standard ImageNet images (exemplary stimuli are visualised on the right). Model colours as in Figure~\ref{subfig:edges_all_pytorch_models}; similar colours indicate same model family. Dashed black lines plot a linear model fit. Whiskers and colored tube show 95\% confidence intervals around the mean. Small transparent circles indicate error consistency between a CNN and an individual human observer; mean consistency is shown as a larger saturated circle.}
    \vspace{-0.5cm} %
    \label{fig:results_accuracy_generalisation}
\end{figure}

We conclude that there is a substantial algorithmic difference between human observers and the investigated sixteen CNNs: humans and CNNs are very likely implementing different strategies. This difference is narrowing down for silhouette stimuli, whereas it is as big as ever for cue conflict, line drawing and ImageNet stimuli: AlexNet from 2012 is just as error-consistent as recent models. Our results are in stark contrast to the observation that better ImageNet models appear to be better models of the primate visual cortex, even if they better predict neural activity \cite{yamins2014performance}.

\textbf{Error consistency vs.\ model architecture.}
We were surprised to see that the consistency between different CNNs is even higher than the consistency between different human observers. In Figure~\ref{subfig:edges_all_pytorch_models}, we investigate the degree to which this CNN-CNN consistency is influenced by similarities in model architecture. When distinguishing between models from the same architecture family (e.g., all ResNet models) and models from a different model family (e.g., ResNet vs. VGG) we observe that even though models from the same family score higher on average, model-to-model consistency is generally very high.\footnote{Results for two other experiments are plotted in the appendix, Figure~\ref{fig:app_all_pytorch_models}.} In line with these results, \cite{mania2019model} also reported extremely high similarity between different models on the ImageNet test set. This might shed some light on the finding that many trained and fitted CNNs predict neural data similarly well, largely irrespective of architecture \cite{storrs2020diverse}. Interestingly, the highest observed error consistency ($\kappa=0.793$) occurs for DenseNet-121 vs.\ ResNet-18: two models from a different model family with different depth (121~vs.~18 layers) and different connectivity. High error consistency between different CNNs suggests that using CNNs as an ensemble may currently be less effective than desirable, since ensembles benefit from independent (rather than consistent) models. It remains an open question why even multiple instances of a single model (trained with a different random seed) \emph{internally} often differ substantially \cite{akbarinia2019paradox, mehrer2020individual}, yet in spite of large architectural differences across models and model families, all CNNs that we investigated seem to be implementing fairly similar strategies.

\subsection{Comparing algorithms with algorithms: the ``current best model of the primate ventral visual stream'' behaves like a vanilla ResNet-50 according to error consistency analysis}
\label{subsec:algorithms_vs_algorithms}

\begin{figure}[t]
    \begin{subfigure}{0.49\textwidth}
        \centering
        \includegraphics[width=\linewidth]{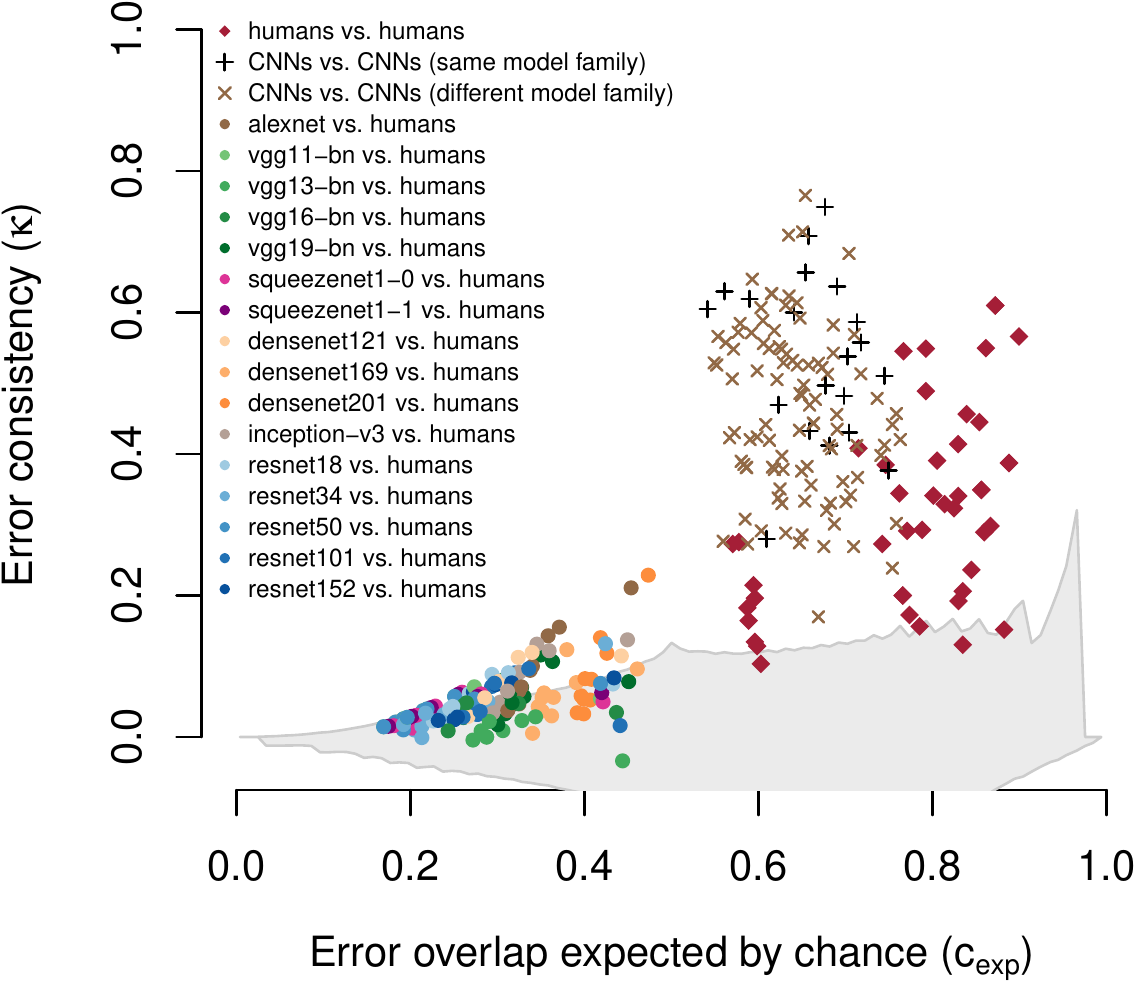}
        \caption{PyTorch models}
        \label{subfig:edges_all_pytorch_models}
    \end{subfigure}\hfill
    \begin{subfigure}{0.49\textwidth}
        \centering
        \includegraphics[width=\linewidth]{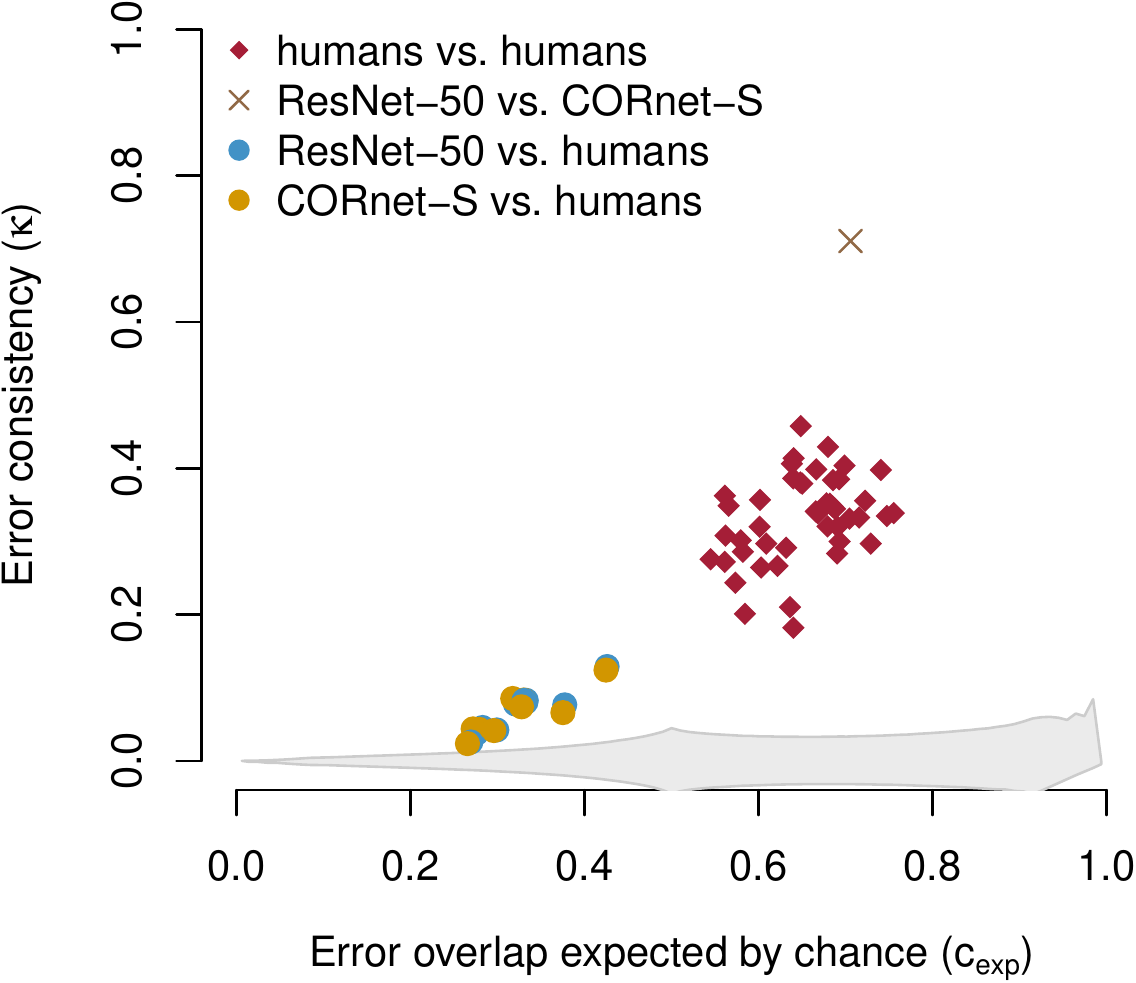}
        \caption{CORnet-S vs. ResNet-50}
        \label{subfig:cue_conflict_cornet_S}
    \end{subfigure}\hfill
    \caption{\textbf{(a)} How is error consistency influenced by model architecture? PyTorch models tested on edge stimuli (160 trials per observer). \textbf{(b)} Recurrent CORnet-S behaves just like a standard feedforward ResNet-50 on cue conflict stimuli (1280 trials).  Shaded areas indicate a simulated 95\% percentile for consistency by chance.}
    \label{fig:cornet}
    \vspace{-0.5cm}
\end{figure}

In order to understand how object recognition is achieved in brains, a necessary---but not sufficient---pre-requisite are quantitative metrics to track improvements and models that improve on those metrics. \citep{kubilius2019brain} went an important step in both directions by proposing \texttt{Brain-Score}, a benchmark where models can be ranked according to a number of metrics, for instance how well their activations predict how biological neurons fire when primates see the same images as an ImageNet-trained CNN. Using this benchmark, the authors tested hundreds of architectures to develop CORnet-S, a brain-inspired recurrent neural network. CORnet-S is able to capture recurrent dynamics (so-called object solution times) of monkey behaviour and achieves previously unmatched performance on \texttt{Brain-Score} while retaining good ImageNet performance (73.1\% top-1). These results, in the author's words, ``establish CORnet-S, a compact, recurrent ANN, as the current best model of the primate ventral visual stream'' performing ``brain-like object recognition'' \cite[p. 1]{kubilius2019brain}. Building such a model is an exciting undertaking and, as perhaps indicated by the highly competitive selection as an ``Oral'' contribution to NeurIPS 2019, an endeavour that sparked considerable excitement at the intersection of the neuroscience and machine learning communities. But how much is behavioural consistency improved in comparison to a baseline model (ResNet-50)? This is exactly the type of question that can be answered with the help of our error consistency analysis.

Figure~\ref{fig:cornet} shows that CORnet-S shares only slightly above-chance error consistency with most human observers---even the highest CORnet-S-to-human error consistency is lower than the lowest human-to-human error consistency. However, there is no improvement whatsoever over a ResNet-50 baseline: Cohen's $\kappa$ for CNN-human consistency is very low for both models (.068: ResNet-50; .066: CORnet-S) compared to .331 for human-human consistency. Perhaps worse still, AlexNet from 2012 has higher error consistency than CORnet-S (.080). CNN-CNN consistency between CORnet-S and ResNet-50 is exceptionally high (.711), many datapoints even overlap exactly---a pattern confirmed by additional experiments in the appendix (Figures~\ref{fig:app_cornet_shape_bias}, \ref{fig:app_cornet_edge_silhouettes} and \ref{fig:app_noise_generalisation}), where we also perform a more detailed comparison to all six \texttt{Brain-Score} metrics (Figures~\ref{fig:app_brainscore_cue_conflict}, \ref{fig:app_brainscore_edges}, \ref{fig:app_brainscore_silhouettes} and \ref{fig:app_brainscore_ImageNet} showing, if at all, only a weak relationship between error consistency and \texttt{Brain-Score} metrics). This indicates that CORnet-S is likely implementing a very different strategy than the human brain: in our analysis, CORnet-S has more behavioural similarities with a standard feedforward ResNet-50 than with human object recognition.\footnote{Interestingly, CORnet-S and ResNet-50 also score fairly similarly on a few metrics of \texttt{Brain-Score}.} This provides evidence that recurrent computations---often argued to be one of the key missing ingredients in standard CNNs towards a better account of biological vision \cite{kriegeskorte2015deep, spoerer2017recurrent, kubilius2019brain, kietzmann2019recurrence, serre2019deep, van2020going}---do not necessarily lead to different behaviour compared to a purely feedforward CNN. It is still an open question to determine the conditions under which recurrence provides advantages over feedforward networks. Recent evidence seems to indicate that recurrence may be especially useful for difficult images \cite{linsley2018learning, tang2018recurrent, kar2019evidence}.

Overall, the observed discrepancy between the leading score of CORnet-S on \texttt{Brain-Score} and its similarity to a standard ResNet-50 according to error consistency analysis points to the decisive importance of metrics: CORnet-S was mainly built for neural predictivity and while it scores very well on a number of other benchmarks, such as capturing object solution times and even a previously reported behavioural error analysis \cite{rajalingham2018large}, it performs poorly on the behavioural metric reported here, \emph{trial-by-trial error consistency}. New metrics to scrutinise models will hopefully lead to an improved generation of models, which in turn might inspire ever-more challenging analyses. An ideal model of biological object recognition would score well on multiple metrics (both neural and behavioural data, an important idea behind \texttt{Brain-Score}), including on metrics that the model was not directly optimised for.
\vspace{-0.25cm}
\section{Conclusion}
\vspace{-0.25cm}
Error consistency is a quantitative analysis for comparing strategies/methods of black-box decision makers---be they brains or algorithms. Accuracy alone is insufficient for distinguishing between strategies: two decision makers may achieve similar accuracy with very different strategies. In contrast to aggregated metrics (averaging across trials/stimuli and observers/networks), error consistency measures behavioural errors on a fine-grained level following the idea of ``molecular psychophysics'' \cite{Green_1964}. Using error consistency we find:
\begin{itemize}[parsep=0.0cm, itemsep=0.2cm,topsep=0pt,leftmargin=0.4cm]
    \item Irrespective of architecture, CNNs are remarkably consistent with one another
    \item The consistency between humans and CNNs, however, is little beyond what can be expected by chance alone, indicating that CNNs still employ very different perceptual mechanisms and ``brain-like machine learning'' may be still but a distant dream (cf.~\cite{lake_building_2017})
    \item Recurrent CORnet-S, termed the ``current best model of the primate ventral visual stream'', fails to capture essential characteristics of human behavioural data and instead behaves effectively like a standard feedforward ResNet-50 in our analysis.
\end{itemize}

Taken together, error consistency analysis suggests that the strategies used by human and machine vision are still very different---but we envision that error consistency will be a useful analysis in the quest to understand complex systems, be they CNNs or the human mind and brain.

\section*{Broader Impact}

\emph{Error consistency} is a statistical analysis for measuring whether two or more decision makers make similar errors. Like any statistical analysis, it can be used for better or worse. For instance, as a very simple example, calculating the \emph{mean} of a number of observations can be used to quantify a world-wide temperature increase caused by human carbon emissions \cite{hagedorn2019a, hagedorn2019b} (positive impact). However, calculating the mean could just as well be utilised by authoritarian governments to obtain an aggregated credit score of ``social''---i.e., conformist---behaviour (negative impact) \cite{creemers2018china}. Concerning error consistency, we could envisage the following broader impact.

\paragraph{Potential positive impact.}
Quantifying differences between decision making strategies can contribute to a better understanding of algorithmic decisions. This improves model interpretability, which is a scientific goal by itself but also closely linked to societal requirements like accountability of algorithmic decision making and the ``right to explanation'' in the European Union \cite{selbst2017meaningful}. Furthermore, calculating the error consistency between humans and CNNs can be used for fact-checking overly hyped ``human-like AI'' statements, e.g. by startups. We argue that human-level accuracy does not imply human-like decision making, which might contribute to increased rigour in model evaluation.

\paragraph{Potential negative impact.}
While not intended to cause any harm, quantifying differences between individuals can be used to identify group-conform and outlier behaviour. Furthermore, measuring error consistency between machines and humans might be used to quantify progress towards building machines that mimic human decision making on certain tasks. While this might sound exciting to a scientist, it very likely sounds a lot more frightening from the perspective of someone losing their job because a machine would then be capable of doing the same work more cheaply. Depending on the complexity of the task, this may not be a problem in the near future but, given current trends in the use of machine learning for automation, perhaps in the distant future.

\subsubsection*{Acknowledgments \& funding disclosure}
Funding was provided, in part, by the Deutsche Forschungsgemeinschaft (DFG, German Research Foundation) -- project number 276693517 -- SFB 1233, TP 4 Causal inference strategies in human vision (K.M. and F.A.W.). The authors thank the International Max Planck Research School for Intelligent Systems (IMPRS-IS) for supporting R.G; and the German Research Foundation through the Cluster of Excellence ``Machine Learning---New Perspectives for Science'', EXC 2064/1, project number 390727645, for supporting F.A.W. The authors declare no competing interests. 

We would like to thank Silke Gramer and Leila Masri for administrative and Uli Wannek for technical support; and David-Elias Künstle, Bernhard Lang, Maximus Mutschler as well as Uli Wannek for helpful comments. We thank \citet{kubilius2019brain} for making their implementation of CORnet-S publicly available. Furthermore, we would like to thank Jonas Kubilius and Martin Schrimpf for feedback and many valuable suggestions.

\subsubsection*{Author contributions}
Based on ideas from \cite{schonfelder2013} and \cite{Meding2019}, R.G. first applied trial-by-trial analysis ideas to CNNs. Thereafter, all three authors jointly initiated the project. R.G. and K.M. jointly led the project. K.M. derived the bounds, performed and visualised the simulations and acquired the Brain-Score data (with input from R.G.). The CNN experiments were performed, analysed and visualised by R.G. (with input from K.M.). F.A.W. provided guidance, feedback, and pointed out the link to molecular psychophysics. All three authors planned and structured the manuscript. R.G. and K.M. wrote the paper with active input from F.A.W.

\small
\bibliographystyle{unsrtnat}
\bibliography{references}

\begin{thebibliography}{69}
\providecommand{\natexlab}[1]{#1}
\providecommand{\url}[1]{\texttt{#1}}
\expandafter\ifx\csname urlstyle\endcsname\relax
  \providecommand{\doi}[1]{doi: #1}\else
  \providecommand{\doi}{doi: \begingroup \urlstyle{rm}\Url}\fi

\bibitem[Lillicrap and Kording(2019)]{lillicrap2019does}
T.~P. Lillicrap and K.~P. Kording.
\newblock What does it mean to understand a neural network?
\newblock \emph{arXiv preprint arXiv:1907.06374}, 2019.

\bibitem[Szegedy et~al.(2013)Szegedy, Zaremba, Sutskever, Bruna, Erhan,
  Goodfellow, and Fergus]{szegedy2013intriguing}
C.~Szegedy, W.~Zaremba, I.~Sutskever, J.~Bruna, D.~Erhan, I.~Goodfellow, and
  R.~Fergus.
\newblock Intriguing properties of neural networks.
\newblock \emph{arXiv preprint arXiv:1312.6199}, 2013.

\bibitem[Ilyas et~al.(2019)Ilyas, Santurkar, Tsipras, Engstrom, Tran, and
  Madry]{ilyas2019adversarial}
A.~Ilyas, S.~Santurkar, D.~Tsipras, L.~Engstrom, B.~Tran, and A.~Madry.
\newblock Adversarial examples are not bugs, they are features.
\newblock In \emph{Advances in Neural Information Processing Systems}, pages
  125--136, 2019.

\bibitem[Geirhos et~al.(2020{\natexlab{a}})Geirhos, Jacobsen, Michaelis, Zemel,
  Brendel, Bethge, and Wichmann]{geirhos2020shortcut}
R.~Geirhos, J.-H. Jacobsen, C.~Michaelis, R.~Zemel, W.~Brendel, M.~Bethge, and
  F.~A. Wichmann.
\newblock Shortcut learning in deep neural networks.
\newblock \emph{{Nature Machine Intelligence}}, in press, 2020{\natexlab{a}}.

\bibitem[Mausfeld(2003)]{Mausfeld_2003a}
R.~Mausfeld.
\newblock No {Psychology} {In} - {No} {Psychology} {Out}.
\newblock \emph{Psychologische Rundschau}, 54\penalty0 (3):\penalty0 185--191,
  2003.

\bibitem[Zeiler and Fergus(2014)]{zeiler2014visualizing}
M.~D. Zeiler and R.~Fergus.
\newblock Visualizing and understanding convolutional networks.
\newblock In \emph{European conference on computer vision}, pages 818--833.
  Springer, 2014.

\bibitem[Zintgraf et~al.(2017)Zintgraf, Cohen, Adel, and
  Welling]{zintgraf2017visualizing}
L.~M. Zintgraf, T.~S. Cohen, T.~Adel, and M.~Welling.
\newblock Visualizing deep neural network decisions: Prediction difference
  analysis.
\newblock \emph{arXiv preprint arXiv:1702.04595}, 2017.

\bibitem[Olah et~al.(2020)Olah, Cammarata, Schubert, Goh, Petrov, and
  Carter]{olah2020zoom}
C.~Olah, N.~Cammarata, L.~Schubert, G.~Goh, M.~Petrov, and S.~Carter.
\newblock Zoom in: An introduction to circuits.
\newblock \emph{Distill}, 5\penalty0 (3):\penalty0 e00024--001, 2020.

\bibitem[Nie et~al.(2018)Nie, Zhang, and Patel]{nie2018theoretical}
W.~Nie, Y.~Zhang, and A.~Patel.
\newblock A theoretical explanation for perplexing behaviors of
  backpropagation-based visualizations.
\newblock \emph{arXiv preprint arXiv:1805.07039}, 2018.

\bibitem[Adebayo et~al.(2018)Adebayo, Gilmer, Muelly, Goodfellow, Hardt, and
  Kim]{adebayo2018sanity}
J.~Adebayo, J.~Gilmer, M.~Muelly, I.~Goodfellow, M.~Hardt, and B.~Kim.
\newblock Sanity checks for saliency maps.
\newblock In \emph{Advances in Neural Information Processing Systems}, pages
  9505--9515, 2018.

\bibitem[Geirhos et~al.(2019)Geirhos, Rubisch, Michaelis, Bethge, Wichmann, and
  Brendel]{geirhos2019imagenettrained}
R.~Geirhos, P.~Rubisch, C.~Michaelis, M.~Bethge, Felix~A. Wichmann, and
  W.~Brendel.
\newblock {ImageNet}-trained {CNN}s are biased towards texture; increasing
  shape bias improves accuracy and robustness.
\newblock In \emph{{International Conference on Learning Representations}},
  2019.

\bibitem[Marr(1982)]{marr1982vision}
D.~Marr.
\newblock \emph{Vision: A computational investigation into the human
  representation and processing of visual information}.
\newblock W.H.Freeman \& Co Ltd, San Francisco, 1982.

\bibitem[Reichenbach(1956)]{reichenbach1956direction}
Hans Reichenbach.
\newblock \emph{The direction of time}.
\newblock Univ of California Press, 1956.

\bibitem[Castelvecchi(2016)]{castelvecchi2016can}
D.~Castelvecchi.
\newblock Can we open the black box of {AI}?
\newblock \emph{Nature News}, 538:\penalty0 20--23, 2016.

\bibitem[Shwartz-Ziv and Tishby(2017)]{shwartz2017opening}
R.~Shwartz-Ziv and N.~Tishby.
\newblock Opening the black box of deep neural networks via information.
\newblock \emph{arXiv preprint arXiv:1703.00810}, 2017.

\bibitem[Kietzmann et~al.(2018)Kietzmann, McClure, and
  Kriegeskorte]{kietzmann2018deep}
T.~C. Kietzmann, P.~McClure, and N.~Kriegeskorte.
\newblock Deep neural networks in computational neuroscience.
\newblock \emph{BioRxiv}, 2018.

\bibitem[Geirhos et~al.(2020{\natexlab{b}})Geirhos, Narayanappa, Mitzkus,
  Bethge, Wichmann, and Brendel]{geirhos2020on}
Robert Geirhos, Kantharaju Narayanappa, Benjamin Mitzkus, Matthias Bethge,
  Felix~A. Wichmann, and Wieland Brendel.
\newblock On the surprising similarities between supervised and self-supervised
  models.
\newblock \emph{arXiv preprint arXiv:2010.08377}, 2020{\natexlab{b}}.

\bibitem[Green(1964)]{Green_1964}
D.~M. Green.
\newblock Consistency of auditory detection judgments.
\newblock \emph{Psychological Review}, 71\penalty0 (5):\penalty0 392--407,
  1964.

\bibitem[Ghodrati et~al.(2014)Ghodrati, Farzmahdi, Rajaei, Ebrahimpour, and
  Khaligh-Razavi]{ghodrati2014feedforward}
M.~Ghodrati, A.~Farzmahdi, K.~Rajaei, R.~Ebrahimpour, and S.-M. Khaligh-Razavi.
\newblock Feedforward object-vision models only tolerate small image variations
  compared to human.
\newblock \emph{Frontiers in Computational Neuroscience}, 8:\penalty0 74, 2014.

\bibitem[Rajalingham et~al.(2015)Rajalingham, Schmidt, and
  DiCarlo]{rajalingham2015comparison}
R.~Rajalingham, K.~Schmidt, and J.~J. DiCarlo.
\newblock Comparison of object recognition behavior in human and monkey.
\newblock \emph{Journal of Neuroscience}, 35\penalty0 (35):\penalty0
  12127--12136, 2015.

\bibitem[Kheradpisheh et~al.(2016{\natexlab{a}})Kheradpisheh, Ghodrati,
  Ganjtabesh, and Masquelier]{kheradpisheh2016deep}
S.~R. Kheradpisheh, M.~Ghodrati, M.~Ganjtabesh, and T.~Masquelier.
\newblock Deep networks can resemble human feed-forward vision in invariant
  object recognition.
\newblock \emph{Scientific Reports}, 6:\penalty0 32672, 2016{\natexlab{a}}.

\bibitem[Kheradpisheh et~al.(2016{\natexlab{b}})Kheradpisheh, Ghodrati,
  Ganjtabesh, and Masquelier]{kheradpisheh2016humans}
S.~R. Kheradpisheh, M.~Ghodrati, M.~Ganjtabesh, and T.~Masquelier.
\newblock Humans and deep networks largely agree on which kinds of variation
  make object recognition harder.
\newblock \emph{Frontiers in Computational Neuroscience}, 10:\penalty0 92,
  2016{\natexlab{b}}.

\bibitem[Geirhos et~al.(2017)Geirhos, Janssen, Sch{\"u}tt, Rauber, Bethge, and
  Wichmann]{geirhos2017comparing}
R.~Geirhos, D.~H.J. Janssen, H.~H. Sch{\"u}tt, J.~Rauber, M.~Bethge, and F.~A
  Wichmann.
\newblock Comparing deep neural networks against humans: object recognition
  when the signal gets weaker.
\newblock \emph{arXiv preprint arXiv:1706.06969}, 2017.

\bibitem[Ma and Peters(2020)]{ma2020a}
W.~J. Ma and B.~Peters.
\newblock A neural network walks into a lab: towards using deep nets as models
  for human behavior.
\newblock \emph{arXiv preprint arXiv:2005.02181}, 2020.

\bibitem[Kubilius et~al.(2016)Kubilius, Bracci, and Op~de
  Beeck]{kubilius2016deep}
J.~Kubilius, S.~Bracci, and H.~P. Op~de Beeck.
\newblock Deep neural networks as a computational model for human shape
  sensitivity.
\newblock \emph{PLoS Computational Biology}, 12\penalty0 (4):\penalty0
  e1004896, 2016.

\bibitem[Rajalingham et~al.(2018)Rajalingham, Issa, Bashivan, Kar, Schmidt, and
  DiCarlo]{rajalingham2018large}
R.~Rajalingham, E.~B. Issa, P.~Bashivan, K.~Kar, K.~Schmidt, and J.~J. DiCarlo.
\newblock Large-scale, high-resolution comparison of the core visual object
  recognition behavior of humans, monkeys, and state-of-the-art deep artificial
  neural networks.
\newblock \emph{Journal of Neuroscience}, 38\penalty0 (33):\penalty0
  7255--7269, 2018.

\bibitem[Mania et~al.(2019)Mania, Miller, Schmidt, Hardt, and
  Recht]{mania2019model}
Horia Mania, John Miller, Ludwig Schmidt, Moritz Hardt, and Benjamin Recht.
\newblock Model similarity mitigates test set overuse.
\newblock In \emph{{Advances in Neural Information Processing Systems}}, pages
  9993--10002, 2019.

\bibitem[Meding et~al.(2019)Meding, Janzing, Sch{\"o}lkopf, and
  Wichmann]{Meding2019}
K.~Meding, D.~Janzing, B.~Sch{\"o}lkopf, and F.~A. Wichmann.
\newblock Perceiving the arrow of time in autoregressive motion.
\newblock In \emph{Advances in Neural Information Processing Systems}, pages
  2303--2314, 2019.

\bibitem[Cohen(1960)]{Cohen1960}
J.~Cohen.
\newblock A coefficient of agreement for nominal scales.
\newblock \emph{Educational and psychological measurement}, 20\penalty0
  (1):\penalty0 37--46, 1960.

\bibitem[Fründ et~al.(2014)Fründ, Wichmann, and Macke]{Frund_2014}
I.~Fründ, F.~A. Wichmann, and J.~H. Macke.
\newblock Quantifying the effect of intertrial dependence on perceptual
  decisions.
\newblock \emph{Journal of Vision}, 14\penalty0 (7), 2014.

\bibitem[Wichmann and Hill(2001)]{wichmann_psychometric_2001a}
F.~A. Wichmann and N.~J. Hill.
\newblock The psychometric function: {I}. {Fitting}, sampling, and goodness of
  fit.
\newblock \emph{Perception \& Psychophysics}, 63\penalty0 (8):\penalty0
  1293--1313, 2001.

\bibitem[Hunt(1986)]{Hunt1986}
R.J. Hunt.
\newblock Percent agreement, pearson's correlation, and kappa as measures of
  inter-examiner reliability.
\newblock \emph{Journal of Dental Research}, 65\penalty0 (2):\penalty0
  128--130, 1986.

\bibitem[Watson and Petrie(2010)]{watson2010method}
PF~Watson and A~Petrie.
\newblock Method agreement analysis: a review of correct methodology.
\newblock \emph{Theriogenology}, 73\penalty0 (9):\penalty0 1167--1179, 2010.

\bibitem[Fleiss et~al.(1969)Fleiss, Cohen, and Everitt]{Fleiss1969}
J.~L. Fleiss, J.~Cohen, and B.~S. Everitt.
\newblock Large sample standard errors of kappa and weighted kappa.
\newblock \emph{Psychological Bulletin}, 72\penalty0 (5):\penalty0 323--327,
  1969.

\bibitem[Hudson and Ramm(1987)]{Hudson1987}
W.~D. Hudson and C.~W. Ramm.
\newblock Correct formulation of the kappa coefficient of agreement.
\newblock \emph{Photogrammetric engineering and remote sensing}, 53\penalty0
  (4):\penalty0 421--422, 1987.

\bibitem[McHugh(2012)]{Mchugh2012}
M.~L. McHugh.
\newblock Interrater reliability: the kappa statistic.
\newblock \emph{Biochemia medica: Biochemia medica}, 22\penalty0 (3):\penalty0
  276--282, 2012.

\bibitem[Bland(2015)]{Bland2015}
M.~Bland.
\newblock \emph{An introduction to medical statistics}.
\newblock Oxford University Press (UK), 2015.

\bibitem[Fleiss et~al.(2003)Fleiss, Levin, and Paik]{fleiss2003}
J.~L. Fleiss, B.~Levin, and M.~C. Paik.
\newblock \emph{Statistical methods for rates and proportions}.
\newblock J. Wiley, Hoboken, N.J, 3rd ed edition, 2003.
\newblock ISBN 978-0-471-52629-2.

\bibitem[Umesh et~al.(1989)Umesh, Peterson, and Sauber]{Umesh1989}
U.~N. Umesh, R.~A. Peterson, and M.~H. Sauber.
\newblock Interjudge agreement and the maximum value of kappa.
\newblock \emph{Educational and Psychological Measurement}, 49\penalty0
  (4):\penalty0 835--850, 1989.

\bibitem[Russakovsky et~al.(2015)Russakovsky, Deng, Su, Krause, Satheesh, Ma,
  Huang, Karpathy, Khosla, Bernstein, et~al.]{russakovsky2015imagenet}
Olga Russakovsky, Jia Deng, Hao Su, Jonathan Krause, Sanjeev Satheesh, Sean Ma,
  Zhiheng Huang, Andrej Karpathy, Aditya Khosla, Michael Bernstein, et~al.
\newblock {ImageNet} large scale visual recognition challenge.
\newblock \emph{{International Journal of Computer Vision}}, 115\penalty0
  (3):\penalty0 211--252, 2015.

\bibitem[Rust and Movshon(2005)]{rust2005praise}
Nicole~C Rust and J~Anthony Movshon.
\newblock In praise of artifice.
\newblock \emph{{Nature Neuroscience}}, 8\penalty0 (12):\penalty0 1647--1650,
  2005.

\bibitem[Martinez-Garcia et~al.(2019)Martinez-Garcia, Bertalm{\'\i}o, and
  Malo]{martinez2019praise}
Marina Martinez-Garcia, Marcelo Bertalm{\'\i}o, and Jes{\'u}s Malo.
\newblock In praise of artifice reloaded: caution with natural image databases
  in modeling vision.
\newblock \emph{{Frontiers in Neuroscience}}, 13:\penalty0 8, 2019.

\bibitem[Gatys et~al.(2016)Gatys, Ecker, and Bethge]{Gatys2016}
L.~A. Gatys, A.~S. Ecker, and M.~Bethge.
\newblock Image style transfer using convolutional neural networks.
\newblock In \emph{{Proceedings of the IEEE Conference on Computer Vision and
  Pattern Recognition}}, pages 2414--2423, 2016.

\bibitem[Geirhos et~al.(2018)Geirhos, Temme, Rauber, Sch{\"u}tt, Bethge, and
  Wichmann]{geirhos2018generalisation}
R.~Geirhos, C.~R.~M. Temme, J.~Rauber, H.~H. Sch{\"u}tt, M.~Bethge, and F.~A.
  Wichmann.
\newblock Generalisation in humans and deep neural networks.
\newblock In \emph{{Advances in Neural Information Processing Systems}}, pages
  7538--7550, 2018.

\bibitem[Miller(1995)]{Miller1995}
G.~A. Miller.
\newblock {WordNet}: a lexical database for {E}nglish.
\newblock \emph{Communications of the ACM}, 38\penalty0 (11):\penalty0 39--41,
  1995.

\bibitem[Kubilius et~al.(2019)Kubilius, Schrimpf, Kar, Rajalingham, Hong,
  Majaj, Issa, Bashivan, Prescott-Roy, Schmidt, et~al.]{kubilius2019brain}
J.~Kubilius, M.~Schrimpf, K.~Kar, R.~Rajalingham, H.~Hong, N.~Majaj, E.~Issa,
  P.~Bashivan, J.~Prescott-Roy, K.~Schmidt, et~al.
\newblock Brain-like object recognition with high-performing shallow recurrent
  {ANNs}.
\newblock In \emph{{Advances in Neural Information Processing Systems}}, pages
  12785--12796, 2019.

\bibitem[Kornblith et~al.(2019)Kornblith, Shlens, and Le]{kornblith2019better}
S.~Kornblith, J.~Shlens, and Q.~V. Le.
\newblock Do better imagenet models transfer better?
\newblock In \emph{Proceedings of the IEEE conference on computer vision and
  pattern recognition}, pages 2661--2671, 2019.

\bibitem[Yamins et~al.(2014)Yamins, Hong, Cadieu, Solomon, Seibert, and
  DiCarlo]{yamins2014performance}
D.~L.~K. Yamins, H.~Hong, C.~F. Cadieu, E.~A. Solomon, D.~Seibert, and J.~J.
  DiCarlo.
\newblock Performance-optimized hierarchical models predict neural responses in
  higher visual cortex.
\newblock \emph{Proceedings of the National Academy of Sciences}, 111\penalty0
  (23):\penalty0 8619--8624, 2014.

\bibitem[Storrs et~al.(2020)Storrs, Kietzmann, Walther, Mehrer, and
  Kriegeskorte]{storrs2020diverse}
Katherine~R Storrs, Tim~C Kietzmann, Alexander Walther, Johannes Mehrer, and
  Nikolaus Kriegeskorte.
\newblock Diverse deep neural networks all predict human it well, after
  training and fitting.
\newblock \emph{bioRxiv}, 2020.

\bibitem[Akbarinia and Gegenfurtner(2019)]{akbarinia2019paradox}
Arash Akbarinia and Karl~R Gegenfurtner.
\newblock Paradox in deep neural networks: Similar yet different while
  different yet similar.
\newblock \emph{arXiv preprint arXiv:1903.04772}, 2019.

\bibitem[Mehrer et~al.(2020)Mehrer, Spoerer, Kriegeskorte, and
  Kietzmann]{mehrer2020individual}
Johannes Mehrer, Courtney~J Spoerer, Nikolaus Kriegeskorte, and Tim~C
  Kietzmann.
\newblock Individual differences among deep neural network models.
\newblock \emph{bioRxiv}, 2020.

\bibitem[Kriegeskorte(2015)]{kriegeskorte2015deep}
N.~Kriegeskorte.
\newblock Deep neural networks: a new framework for modeling biological vision
  and brain information processing.
\newblock \emph{Annual review of vision science}, 1:\penalty0 417--446, 2015.

\bibitem[Spoerer et~al.(2017)Spoerer, McClure, and
  Kriegeskorte]{spoerer2017recurrent}
C.~J. Spoerer, P.~McClure, and N.~Kriegeskorte.
\newblock Recurrent convolutional neural networks: a better model of biological
  object recognition.
\newblock \emph{Frontiers in psychology}, 8:\penalty0 1551, 2017.

\bibitem[Kietzmann et~al.(2019)Kietzmann, Spoerer, S{\"o}rensen, Cichy, Hauk,
  and Kriegeskorte]{kietzmann2019recurrence}
T.~C. Kietzmann, C.~J. Spoerer, L.~K.~A. S{\"o}rensen, R.~M. Cichy, O.~Hauk,
  and N.~Kriegeskorte.
\newblock Recurrence is required to capture the representational dynamics of
  the human visual system.
\newblock \emph{Proceedings of the National Academy of Sciences}, 116\penalty0
  (43):\penalty0 21854--21863, 2019.

\bibitem[Serre(2019)]{serre2019deep}
T.~Serre.
\newblock Deep learning: the good, the bad, and the ugly.
\newblock \emph{Annual Review of Vision Science}, 5:\penalty0 399--426, 2019.

\bibitem[van Bergen and Kriegeskorte(2020)]{van2020going}
R.~S. van Bergen and N.~Kriegeskorte.
\newblock Going in circles is the way forward: the role of recurrence in visual
  inference.
\newblock \emph{arXiv preprint arXiv:2003.12128}, 2020.

\bibitem[Linsley et~al.(2018)Linsley, Kim, Veerabadran, Windolf, and
  Serre]{linsley2018learning}
D.~Linsley, J.~Kim, V.~Veerabadran, C.~Windolf, and T.~Serre.
\newblock Learning long-range spatial dependencies with horizontal gated
  recurrent units.
\newblock In \emph{{Advances in Neural Information Processing Systems}}, pages
  152--164, 2018.

\bibitem[Tang et~al.(2018)Tang, Schrimpf, Lotter, Moerman, Paredes, Caro,
  Hardesty, Cox, and Kreiman]{tang2018recurrent}
H.~Tang, M.~Schrimpf, W.~Lotter, C.~Moerman, A.~Paredes, J.~O. Caro,
  W.~Hardesty, D.~Cox, and G.~Kreiman.
\newblock Recurrent computations for visual pattern completion.
\newblock \emph{{Proceedings of the National Academy of Sciences}},
  115\penalty0 (35):\penalty0 8835--8840, 2018.

\bibitem[Kar et~al.(2019)Kar, Kubilius, Schmidt, Issa, and
  DiCarlo]{kar2019evidence}
K.~Kar, J.~Kubilius, K.~Schmidt, E.~B. Issa, and J.~J. DiCarlo.
\newblock Evidence that recurrent circuits are critical to the ventral
  stream’s execution of core object recognition behavior.
\newblock \emph{{Nature Neuroscience}}, 22\penalty0 (6):\penalty0 974--983,
  2019.

\bibitem[Lake et~al.(2017)Lake, Ullman, Tenenbaum, and
  Gershman]{lake_building_2017}
B.~M. Lake, T.~D. Ullman, J.~B. Tenenbaum, and S.~J. Gershman.
\newblock Building machines that learn and think like people.
\newblock \emph{Behavioral and Brain Sciences}, 40, 2017.

\bibitem[Hagedorn et~al.(2019{\natexlab{a}})Hagedorn, Kalmus, Mann, Vicca,
  Van~den Berge, van Ypersele, Bourg, Rotmans, Kaaronen, Rahmstorf,
  et~al.]{hagedorn2019a}
G.~Hagedorn, P.~Kalmus, M.~Mann, S.~Vicca, J.~Van~den Berge, J.-P. van
  Ypersele, D.~Bourg, J.~Rotmans, R.~Kaaronen, S.~Rahmstorf, et~al.
\newblock Concerns of young protesters are justified.
\newblock \emph{Science}, 364:\penalty0 139--140, 2019{\natexlab{a}}.

\bibitem[Hagedorn et~al.(2019{\natexlab{b}})Hagedorn, Loew, Seneviratne, Lucht,
  Beck, Hesse, Knutti, Quaschning, Schleimer, Mattauch, et~al.]{hagedorn2019b}
G.~Hagedorn, T.~Loew, S.~I. Seneviratne, W.~Lucht, M.-L. Beck, J.~Hesse,
  R.~Knutti, V.~Quaschning, J.-H. Schleimer, L.~Mattauch, et~al.
\newblock The concerns of the young protesters are justified: A statement by
  scientists for future concerning the protests for more climate protection.
\newblock \emph{GAIA-Ecological Perspectives for Science and Society},
  28\penalty0 (2):\penalty0 79--87, 2019{\natexlab{b}}.

\bibitem[Creemers(2018)]{creemers2018china}
R.~Creemers.
\newblock China's social credit system: an evolving practice of control.
\newblock \emph{Available at SSRN 3175792}, 2018.

\bibitem[Selbst and Powles(2017)]{selbst2017meaningful}
A.~D. Selbst and J.~Powles.
\newblock Meaningful information and the right to explanation.
\newblock \emph{International Data Privacy Law}, 7\penalty0 (4):\penalty0
  233--242, 2017.

\bibitem[Schönfelder and Wichmann(2013)]{schonfelder2013}
Vinzenz~H. Schönfelder and Felix~A. Wichmann.
\newblock Identification of stimulus cues in narrow-band tone-in-noise
  detection using sparse observer models.
\newblock \emph{The Journal of the Acoustical Society of America}, 134\penalty0
  (1):\penalty0 447--463, 2013.

\bibitem[Hyndman and Fan(1996)]{Hyndman1996}
R.~J. Hyndman and Y.~Fan.
\newblock Sample quantiles in statistical packages.
\newblock \emph{The American Statistician}, 50\penalty0 (4):\penalty0 361--365,
  1996.

\bibitem[Paszke et~al.(2019)Paszke, Gross, Massa, Lerer, Bradbury, Chanan,
  Killeen, Lin, Gimelshein, Antiga, Desmaison, Kopf, Yang, DeVito, Raison,
  Tejani, Chilamkurthy, Steiner, Fang, Bai, and Chintala]{pytorch}
A.~Paszke, S.~Gross, F.~Massa, A.~Lerer, J.~Bradbury, G.~Chanan, T.~Killeen,
  Z.~Lin, N.~Gimelshein, L.~Antiga, A.~Desmaison, A.~Kopf, E.~Yang, Z.~DeVito,
  M.~Raison, A.~Tejani, S.~Chilamkurthy, B.~Steiner, L.~Fang, J.~Bai, and
  S.~Chintala.
\newblock Pytorch: An imperative style, high-performance deep learning library.
\newblock In \emph{Advances in Neural Information Processing Systems 32}, pages
  8024--8035. Curran Associates, Inc., 2019.

\bibitem[Simonyan and Zisserman(2015)]{simonyan2015very}
K.~Simonyan and A.~Zisserman.
\newblock Very deep convolutional networks for large-scale image recognition.
\newblock arXiv preprint arXiv:1409.1556, 2015.

\bibitem[Hermann and Kornblith(2019)]{hermann2019exploring}
Katherine~L Hermann and Simon Kornblith.
\newblock Exploring the origins and prevalence of texture bias in convolutional
  neural networks.
\newblock \emph{arXiv preprint arXiv:1911.09071}, 2019.

\end{thebibliography}

\newpage
\appendix
\renewcommand{\thesection}{}
\renewcommand{\thesubsection}{S.\arabic{subsection}}
\counterwithin{figure}{section}
\section{\hspace{-0.4cm}Supplementary Material}
\renewcommand{\thesection}{SF}

Code and data to reproduce results and figures are available from \url{https://github.com/wichmann-lab/error-consistency}.

The supplementary material is structured as follows. We start with terminology in Section~\ref{app:sec:terminology}, afterwards we derive bounds of \(c_{obs}\) and kappa in Section~\ref{app:derivation_bounds_cobs} (limiting possible consistency), followed by a description of how we simulated the confidence intervals for \(c_{exp}\) and kappa under the null hypothesis of independent observers in Section~\ref{app:CI_kappa}. Finally, we provide method details for \texttt{Brain-Score} and the evaluated CNNs in Section~\ref{app:method_details_CNNs} and report accuracies across experiments in Table~\ref{app:table_accuracies}.

\begin{figure}
    \begin{subfigure}{0.49\textwidth}
        \centering
        \includegraphics[width=\linewidth]{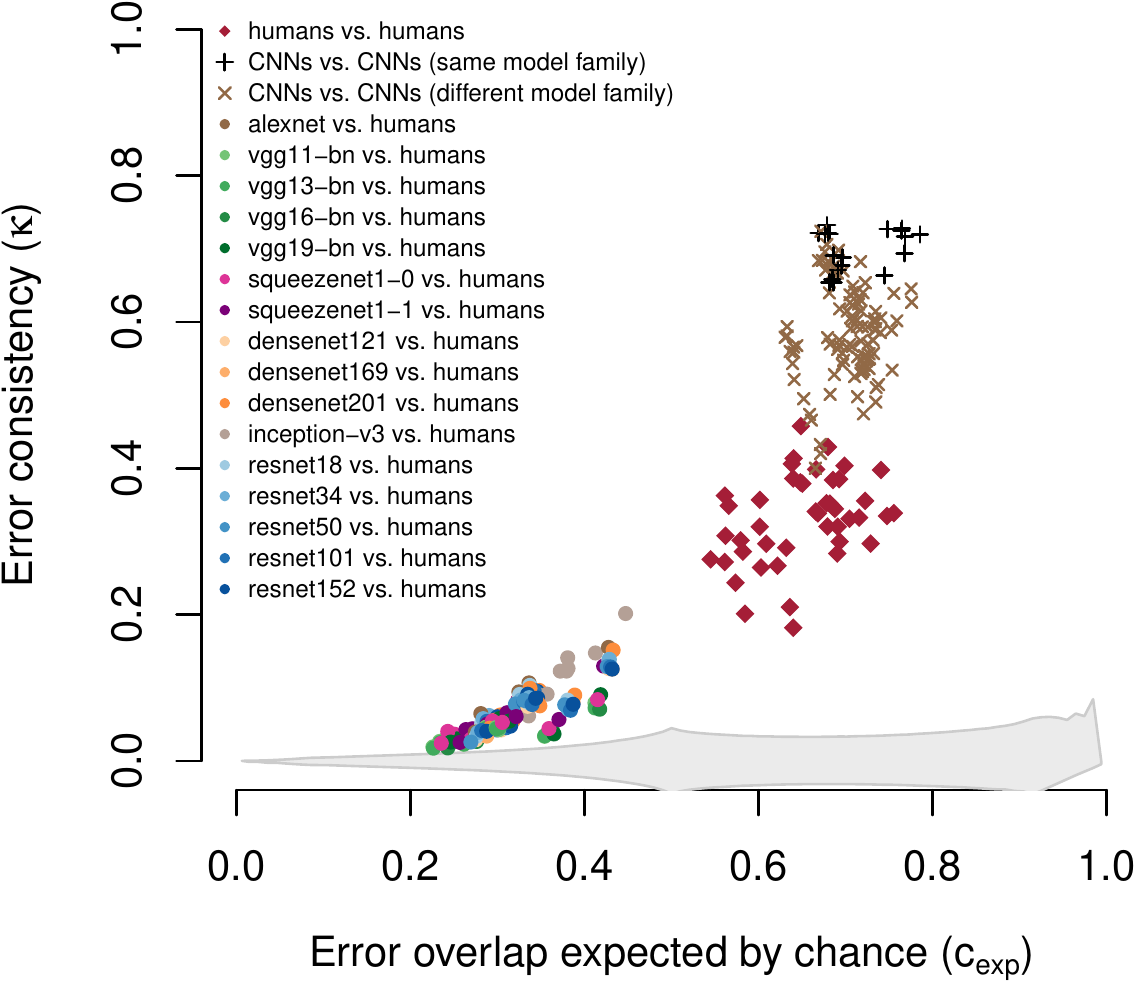}
        \caption{Cue conflict stimuli}
        \label{subfig:cue_conflict_all_pytorch_models}
    \end{subfigure}\hfill
    \begin{subfigure}{0.49\textwidth}
        \centering
        \includegraphics[width=\linewidth]{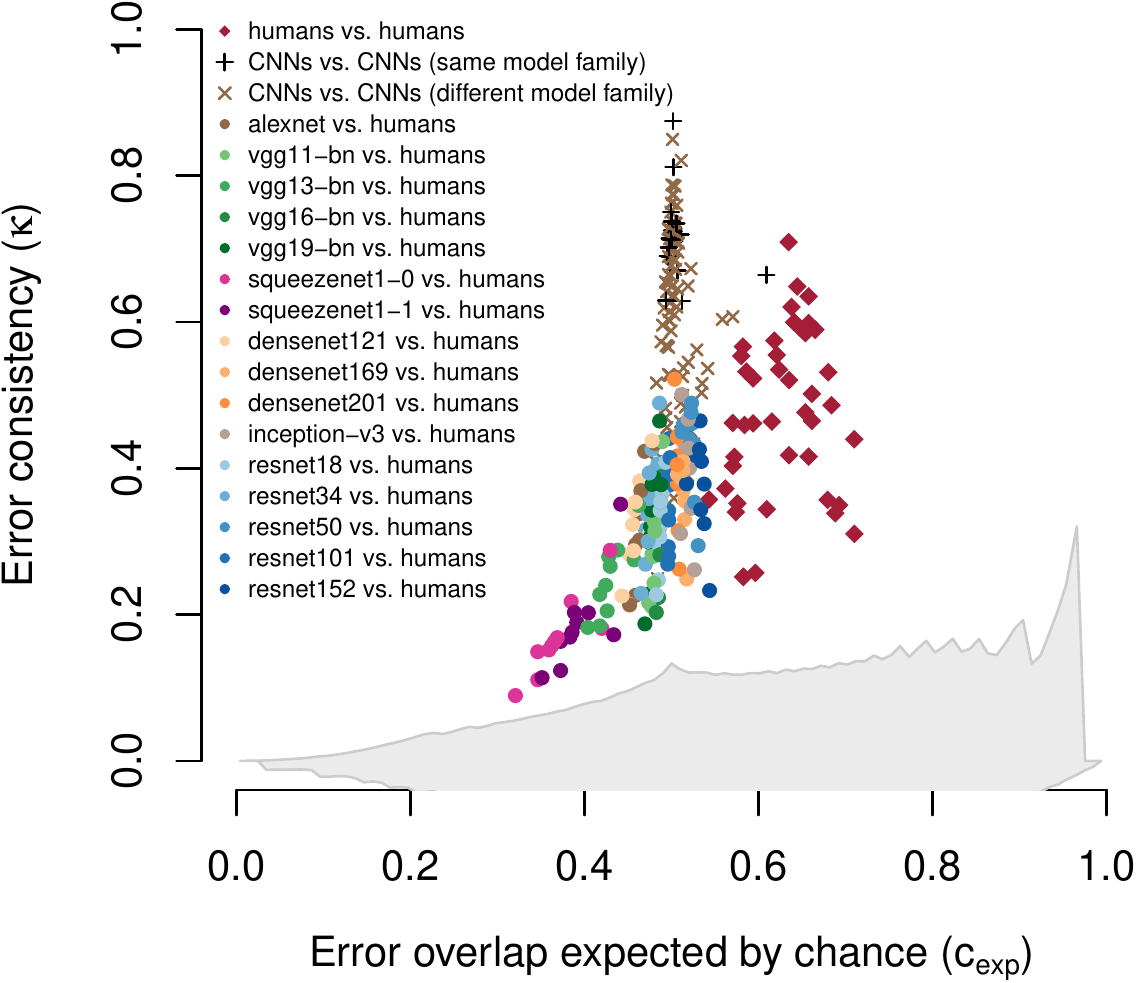}
        \caption{Edge stimuli}
        \label{app_subfig_silhouettes_cornet_S}
    \end{subfigure}\hfill
    \caption{Error consistencs vs. expected error overlap for all PyTorch models.}
    \label{fig:app_all_pytorch_models}
\end{figure}

\begin{figure}
    \centering
    \includegraphics[width=0.7\linewidth]{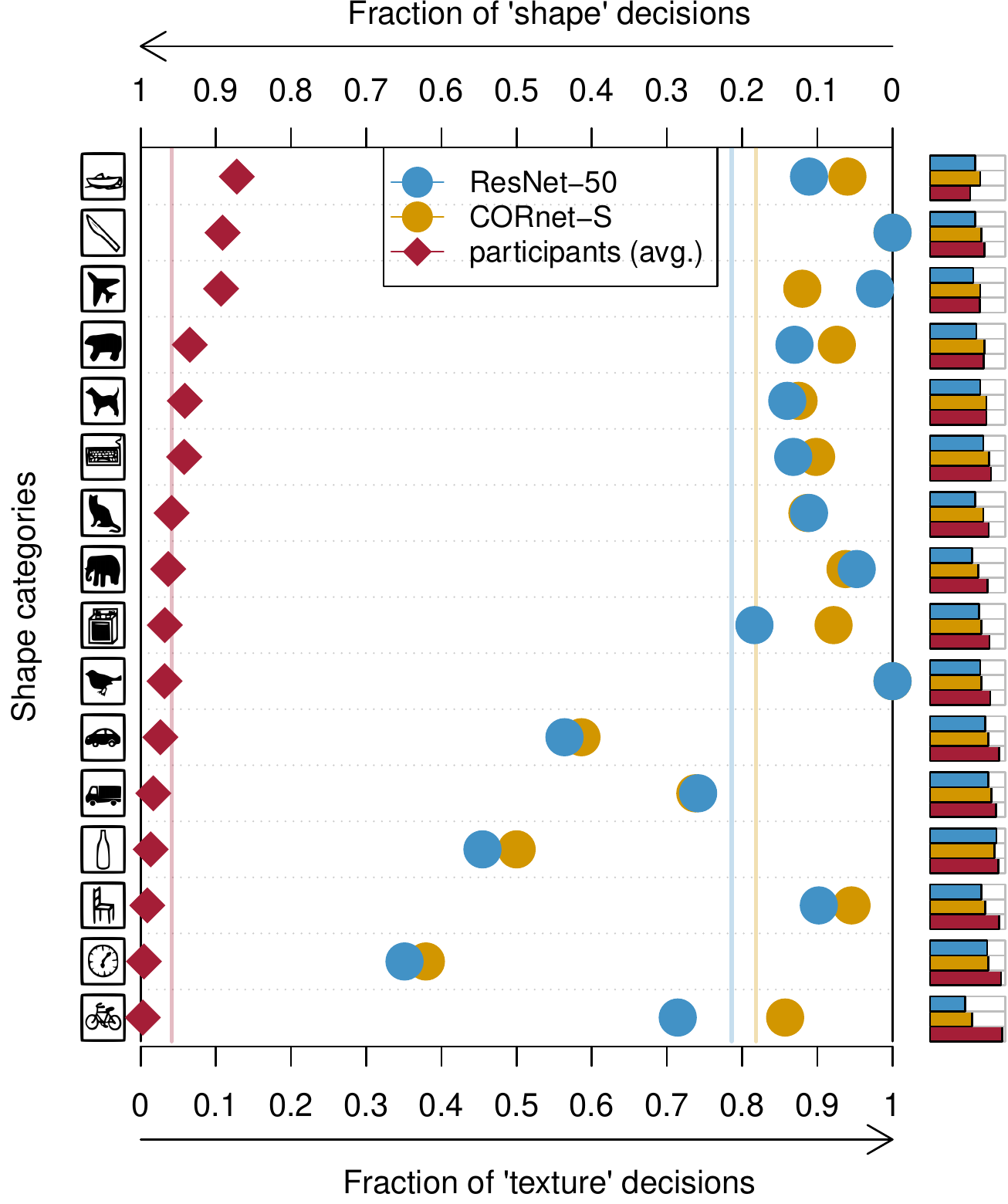}
    \caption{Shape bias of CORnet-S and ResNet-50 in comparison to human observers. Human observers categorise objects by shape rather than texture \cite{geirhos2019imagenettrained}, which differentiates them from standard ImageNet-trained CNNs like ResNet-50 (categorising predominantly by texture). In this experiment, CORnet-S again behaves similarly to ResNet-50 but does not show a human-like shape bias as would be expected for an accurate model of human object recognition. Small bar plots on the right indicate accuracy (answer corresponds to either correct texture category or correct shape category). This pattern was also observed by \citet{hermann2019exploring}, who performed a detailed investigation of the factors that influence model shape bias.}
    \label{fig:app_cornet_shape_bias}
\end{figure}

\begin{figure}
    \begin{subfigure}{0.49\textwidth}
        \centering
        \includegraphics[width=\linewidth]{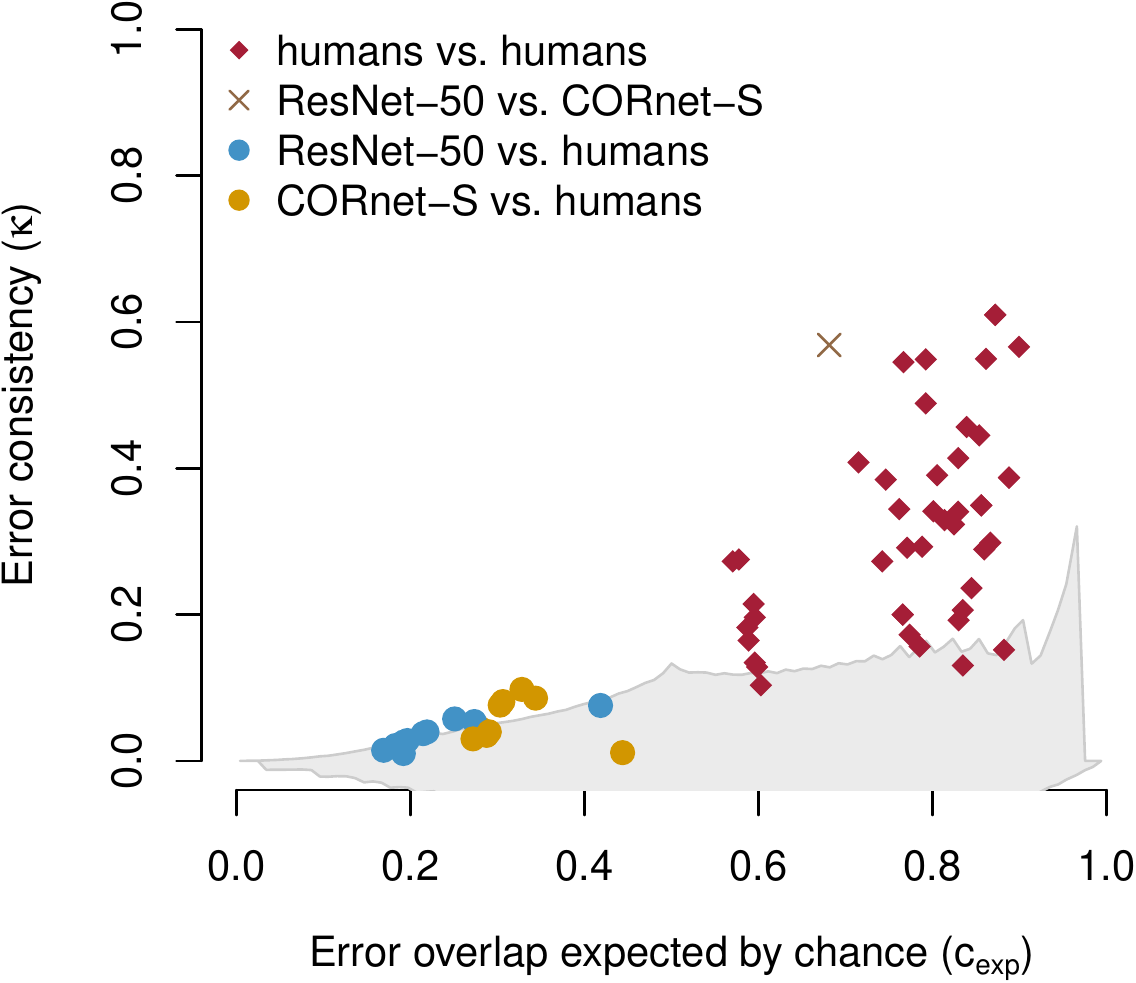}
        \caption{Edge stimuli}
    \end{subfigure}\hfill
    \begin{subfigure}{0.49\textwidth}
        \centering
        \includegraphics[width=\linewidth]{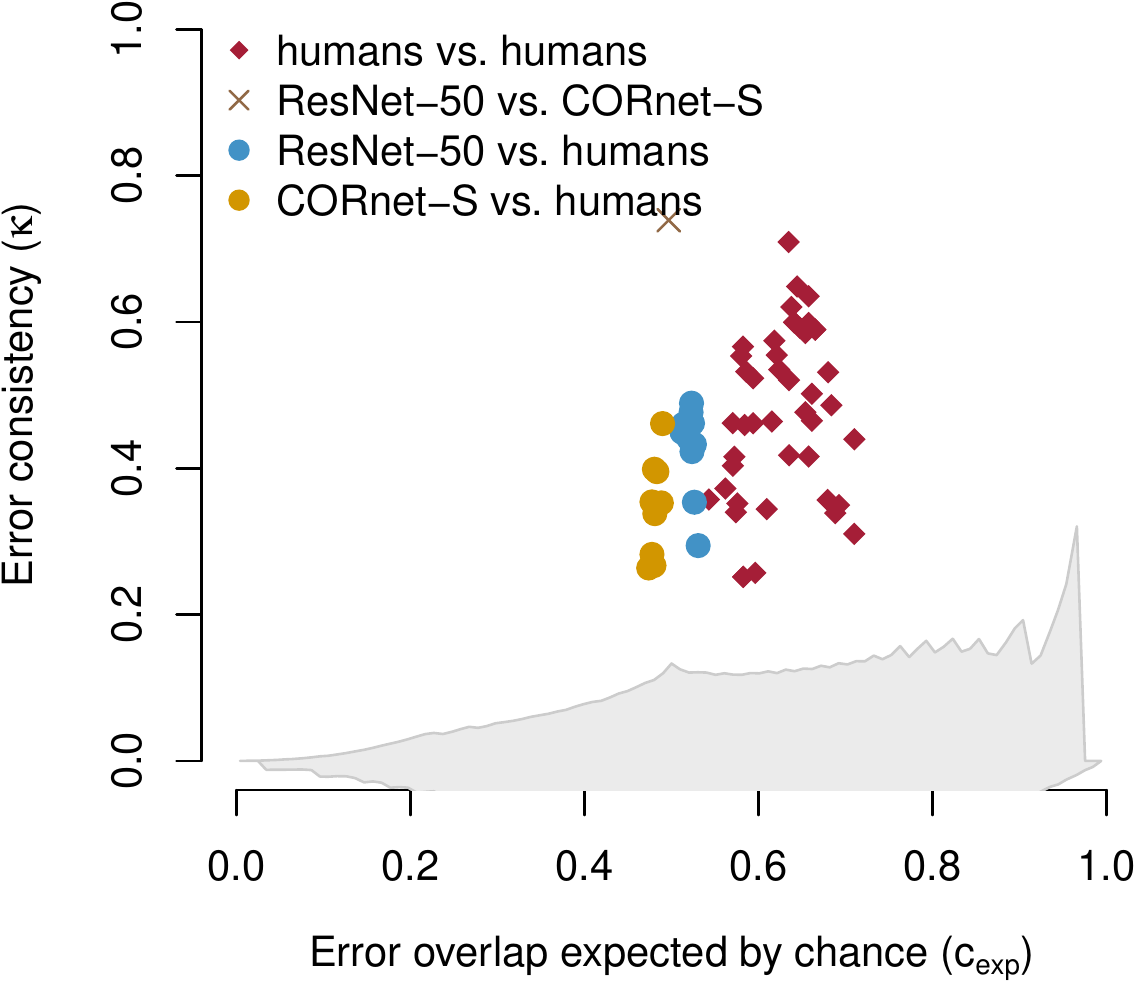}
        \caption{Silhouette stimuli}
    \end{subfigure}\hfill
    \caption{Error consistency of CORnet-S vs. ResNet-50 for edge and silhouette stimuli.}
    \label{fig:app_cornet_edge_silhouettes}
\end{figure}

In addition to method details, we provide extended experimental results in Figure~\ref{fig:app_all_pytorch_models} (error consistency of all PyTorch models for cue conflict and edge stimuli) as well as Figures~\ref{fig:app_cornet_shape_bias}, \ref{fig:app_cornet_edge_silhouettes}, \ref{fig:app_noise_generalisation}, \ref{fig:app_confusion} (detailed analyses of CORnet-S vs.\ ResNet-50). Figures~\ref{fig:app_brainscore_cue_conflict}, \ref{fig:app_brainscore_edges} and \ref{fig:app_brainscore_silhouettes} and \ref{fig:app_brainscore_ImageNet} (investigating the relationship between \texttt{Brain-Score} metrics and error consistency).

\begin{figure}[h!]
    \begin{subfigure}{0.9\textwidth}
        \centering
        \includegraphics[width=\linewidth]{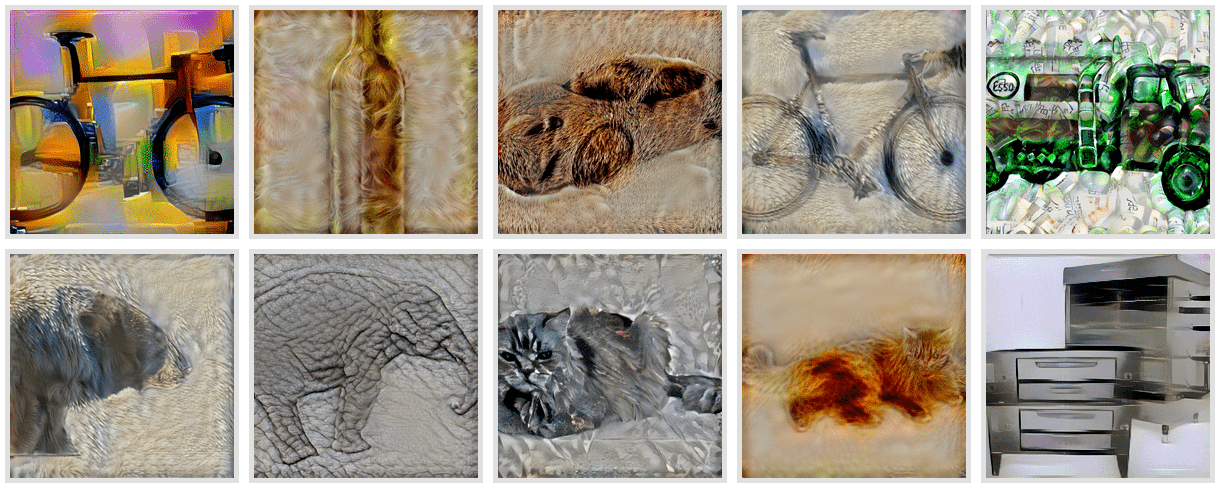}
    \caption{Cue conflict stimuli}
    \end{subfigure}\hfill
    
    \begin{subfigure}{0.9\textwidth}
        \centering
        \includegraphics[width=\linewidth]{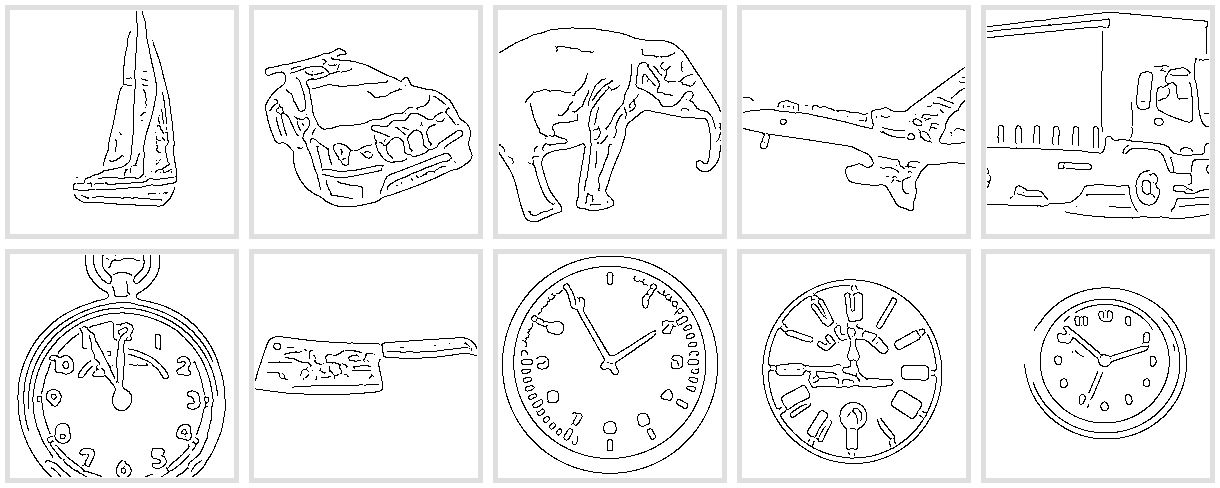}
    \caption{Edge stimuli}
    \end{subfigure}\hfill
    
    \begin{subfigure}{0.9\textwidth}
        \centering
        \includegraphics[width=\linewidth]{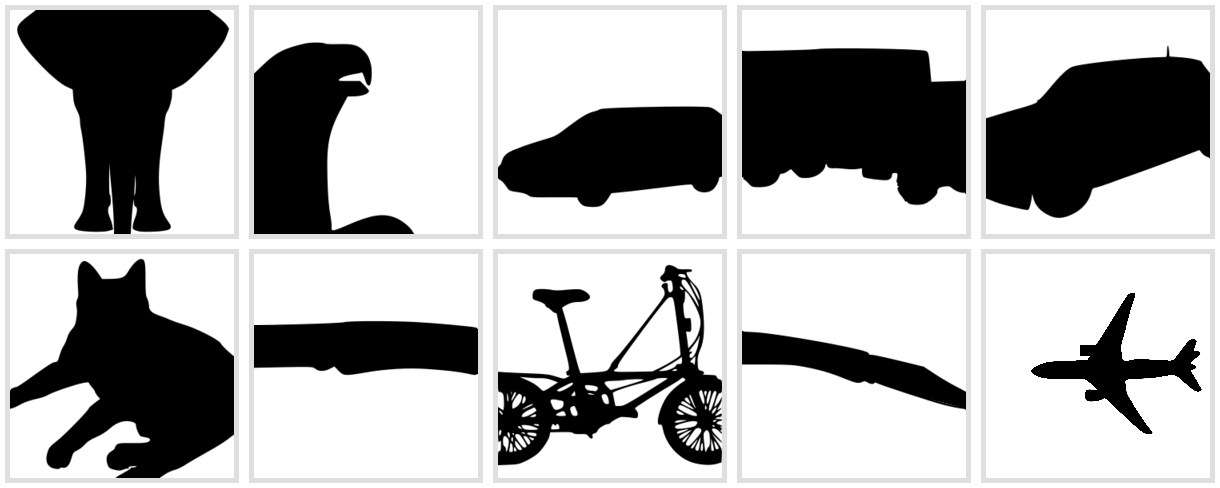}
    \caption{Silhouette stimuli}
    \end{subfigure}\hfill
    \caption{``Easy'' stimuli for humans and CNNs. For each experiment, the images in the top row were those that most humans correctly classified. In the bottom row: stimuli that most CNNs correctly classified. If there were more than five images where humans were very accurate on, we here selected those where CNNs were the least accurate, and vice versa. ImageNet stimuli are not visualised due to image permission reasons.}
    \label{fig:app_easy_stimuli}
\end{figure}

Furthermore, Figure~\ref{fig:app_easy_stimuli} visualises qualitative error differences by plotting which stimuli were particularly easy for humans and CNNs, respectively.

\subsection{Terminology: ``error consistency''}
\label{app:sec:terminology}
We would like to briefly clarify the name \textit{error consistency}. Our analysis helps to compare the consistency of two decision makers. Two decision makers necessarily show some degree of consistency due to chance agreement. Error consistency helps to examine whether the two decision makers show significantly more consistency than expected by chance by analysing behavioural error patterns.
However, this analysis takes into account not only the consistency of errors but also the consistency of correctly answered trials, hence `error consistency' may sound imprecise at first. Nonetheless, we believe that the term captures the most crucial aspect of this analysis: Humans and CNNs ---which are particularly well suited for our analysis---are often close to ceiling performance or at least have high accuracies. Thus trials where the decision makers agree do not provide much evidence for distinguishing between processing strategies. In contrast, the (few) errors of the decision makers are the most informative trials in this respect: Hence the name error consistency.

\subsection{Derivation of bounds for $c_{obs}$ and kappa given $c_{exp}$}
How much observed consistency can we expect at most for a given expected consistency? We assume two independent observers $i$ and $j$ with accuracies $p_i$ and $p_j$. For given $p_i,p_j$ only a certain range of $c_{obs}$ is possible: 
\begin{align}
c_{obs_{max}} = 1-|p_i-p_j| \text{~and~~}
c_{obs_{min}}  = |p_j + p_i -1|. \label{eq:obsminmax}
\end{align}

Ideally, we also want to express the bounds of $c_{obs}$ directly as a function of $c_{exp}$. We obtain the following bounds:

\begin{align}
0 &\leq c_{obs_{i,j}} \leq 1 - \sqrt{1 -  2 c_{exp_{i,j} }} ~~~~  &if~~ c_{exp_{i,j}}  < 0.5, \\
\sqrt{2 c_{exp_{i,j}}-1} &\leq c_{obs_{i,j}} \leq 1 ~~~~ &if~ ~c_{exp_{i,j}}  \geq 0.5.
\end{align}

These bounds are visualised in Figure~\ref{fig:CI_and_bounds}.

\label{app:derivation_bounds_cobs}
The derivation is as follows. We distinguish between two cases.
\paragraph{Case 1:} \(p_i \leq0.5~ \& ~p_j \leq0.5\) or \(p_i \geq0.5~ \&~ p_j \geq0.5 \Longleftrightarrow c_{exp_{i,j}} \geq 0.5\)

\begin{figure}[!h]
    \centering
    \includegraphics[width=0.9\linewidth]{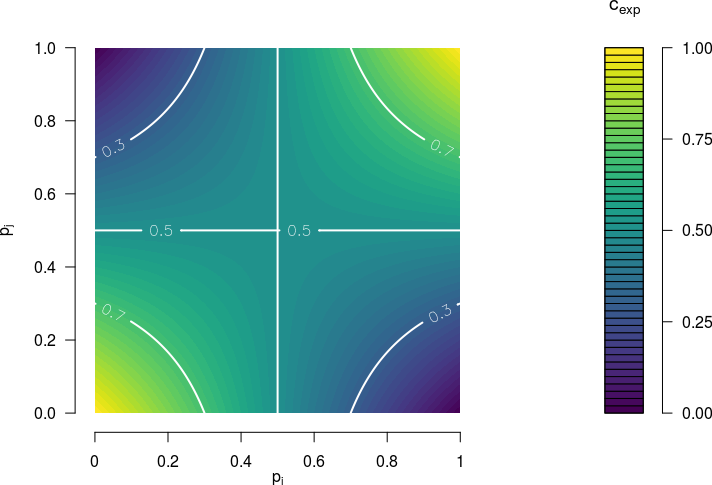}
    \caption{Values that \(c_{exp}\) can take depending on \(p_i\) and \(p_j\) for two independent observers.}
    \label{fig:DensityCobs}
\end{figure}

The expected consistency then lies in the interval of \([0.5,1]\), see Figure~\ref{fig:DensityCobs}.
First we calculate the upper bound \(b_{obs_{max}}\) given \(c_{exp_{i,j}}\). Please note that a specific \( c_{exp_{i,j}}\) can be obtained by multiple combinations of values for \(p_i\) and \(p_j\). For a given \( c_{exp_{i,j}}\) we choose \(p_j = p_i\). We can calculate the exact value of \(p_i\) in this case with eq. \eqref{eq:cexp}. However since \(p_j = p_i\) we get with eq. \eqref{eq:obsminmax} that \( b_{obs_{max}} = 1\). Thus we directly obtain from eq. \eqref{eq:obsminmax} that the upper bound of \( c_{obs_{i,j}}\)is always 1 for all \( c_{exp_{i,j}}\) in the interval [0.5, 1].\\

It is a bit more challenging to derive the lower bound \(b_{obs_{min}}\) given \( c_{exp_{i,j}}\). Using equation \eqref{eq:obsminmax} and \eqref{eq:cexp} we obtain
\begin{align}
b_{obs_{min}} &= p_i + \frac{c_{exp_{i,j}} + p_i -1}{2 p_i -1} -1. \label{eq:bobsI}
\end{align}
Setting $\frac{ \partial b_{obs_{min}}}{ \partial p_i} =  0$ to find the minimum results in\\
\begin{align}
p_{i_{min}} &=\frac{1}{2} \pm \sqrt{\frac{1}{4} - \frac{-2 c_{exp_{i,j}} +2}{4}}.\label{eq:pmin}
\end{align}
We only take the positive term in eq. \eqref{eq:pmin} since \(p_i > 0.5\) by definition. Checking the second order derivative confirms a minimum.
Finally using equation eq. \eqref{eq:pmin}  with eq. \eqref{eq:bobsI}  we calculate
\begin{align}
b_{obs_{min}} &= \sqrt{2 c_{exp_{i,j}} -1}\label{eq:bobsII}, \text{~thus}\\
\sqrt{ 2 c_{exp_{i,j}} -1} &\leq c_{obs_{i,j}} \leq 1.
\end{align} 

\paragraph{Case 2:} \(p_i >0.5~ \& ~p_j <0.5\) or \(p_i <0.5~ \&~ p_j >0.5 \Longleftrightarrow c_{exp_{i,j}} < 0.5\)

The expected consistency then lies in the interval of \([0,0.5[\), see Figure~\ref{fig:DensityCobs}. This case is point symmetric to the right part. Thus we obtain for the bounds of the left part
\begin{align}
b_{{obs_{max}}_2} & =  1 -b_{obs_{min}} (1-c_{exp_{i,j}}), \\
b_{{obs_{min}}_2} & = 0 \text{~ and finally} \\
0 \leq c_{obs_{i,j}} &\leq 1 - \sqrt{1 -  2 c_{exp_{i,j} }}.
\end{align}

\paragraph{Bounds for kappa}
If we plug in the bounds of \( c_{obs_{i,j}}\) into the equation of kappa, we obtain the following bounds for kappa:

\begin{align}
   \frac{-c_{exp_{i,j}}}{1-c_{exp_{i,j}}} &\leq \kappa_{i,j} \leq \frac{1-\sqrt{1 - 2 c_{exp_{i,j}}} - c_{exp_{i,j}}}{1- c_{exp_{i,j}}} ~~~~  & \text{if } c_{exp_{i,j}}  < 0.5, \\
\frac{\sqrt{2 c_{exp_{i,j}} -1} - c_{exp_{i,j}}}{1- c_{exp_{i,j}}} &\leq \kappa_{i,j} \leq 1 ~~~~ & \text{if } c_{exp_{i,j}}  \geq 0.5.
\end{align}

\subsection{Calculating 95\% percentiles of observed overlap and kappa for the null hypothesis of independent observers given an expected consistency}
\label{app:CI_kappa}

Here we describe the procedure to calculate 95\% percentiles of $\kappa$ and $c_{obs}$.

Our null hypothesis is that two decision makers are independent. Assuming independence, we can easily simulate these two observers. 
Based on \(p_i,p_j\) (the accuracies of decision makers $i$ and $j$) we sample $n$ trials and calculate \(c_{exp_{i,j}},c_{obs_{i,j}},\text{and} \kappa_{i,j}\) accordingly based on these simulated values.
This process is repeated systematically for different \(p_i\) and \(p_j\). For this purpose we sample a grid of 4200 x 4200 points in the range \([[0,1],[0,1]]\). For each individual combination of \(p_i\) and \(p_j\), the sampling is repeated five times, thus in total we simulate $4200 \times 4200 \times 5 = 88,200,000$ values.\footnote{The more values are simulated, the better: we chose the maximum number of samples feasible to simulate on our hardware within reasonable time.}

The grid is not divided equally. 66\% of \(p_i\) and \(p_j\) are located in the upper and lower 15\% of the domain. This is important because kappa diverges for large values of $c_{exp}$ (small and large values of \(p_i\) and \(p_j\)); thus a dense sampling is necessary there. 

Based on these simulated data we obtain 95\% percentiles for \(c_{obs}\) and \(\kappa\). We binned the data in 1\% steps and used the standard quantile-function of R (type 7, see \cite{Hyndman1996}). It is important to note that we have only a small number of trials (160 or 1280).\footnote{Percentiles for a different number of trials can also be computed with the code that we provide.} Therefore \(c_{obs}\) can take a maximum number of 161 or 1281 values respectively. The range of uniquely observed values is very small for a given \(c_{exp}\). This implies that the accuracy of our percentiles is limited for data points that are very close to the quantiles. However, this does not influence our findings.

\begin{figure}
    \begin{subfigure}{0.99\textwidth}
        \centering
            \includegraphics[width=\linewidth]{figures/DiagnosticPlotsKappa/DiagnosticPlotSimulateCobsKappa_160.png}
    \end{subfigure}\hfill
    \begin{subfigure}{0.99\textwidth}
        \centering
        \includegraphics[width=\linewidth]{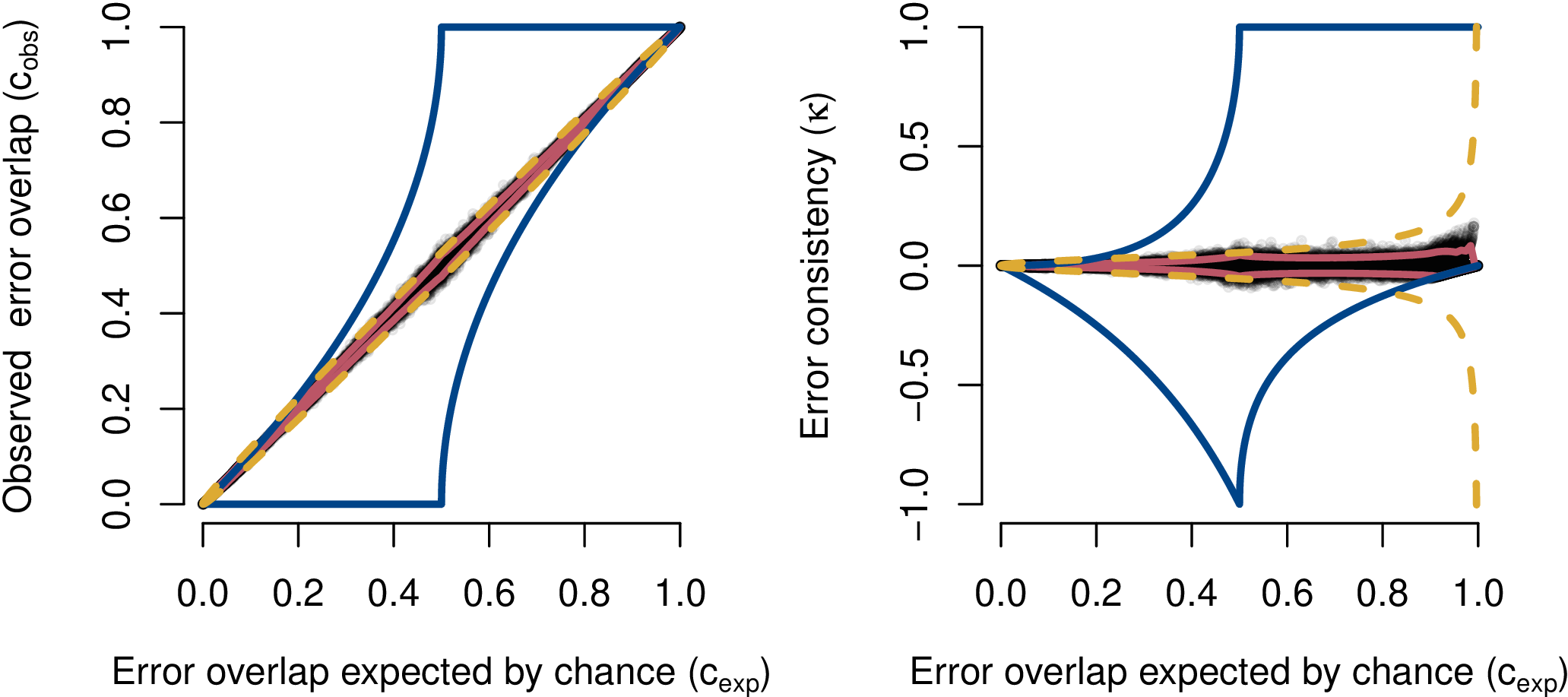}
    \end{subfigure}
    \begin{subfigure}{0.99\textwidth}
        \centering
        \includegraphics[width=0.7\linewidth]{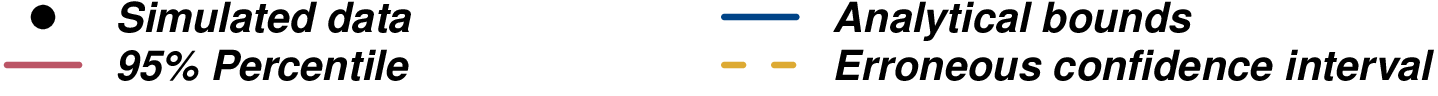}

    \end{subfigure}\hfill
    \caption{Simulated data of \(c_{exp},c_{obs}\) and $\kappa$ for 160 (top) and 1280 (bottom) trials per block. Black dots show 100.000 randomly drawn blocks from our simulation. Blue lines show analytical bounds. Red lines show the 95\% percentiles. Orange dashed lines show the wrong binomial confidence interval (left) and the erroneous confidence interval for $\kappa$ (right) reported in many papers.}
    \label{fig:KappaDiagnostic}
\end{figure}

Please note that the denominator of kappa gets very small for high values of \(c_{exp}\). Thus we see some instability of kappa towards high expected consistencies. Figure~\ref{fig:KappaDiagnostic} shows diagnostic plots for both cases.

\subsection{Disentangling of Error consistency and Accuracy}
\label{sec::accvskappa}
Our argument for the disentanglement between kappa and accuracy is as follows. For independent observer no correlation between accuracy and kappa is observed, e.g. In Figure \ref{fig:CI_and_bounds}b, $\kappa $ and $c_{exp}$ \footnote{Accuracy and $c_{exp}$ are linked as one can see in figure \ref{fig:DensityCobs}}  are not correlated (r=-0.00015, \textit{p} > 0.05). As expressed by the bounds in Figure~\ref{fig:CI_and_bounds}, $\kappa$ is limited by accuracy. If two observers have an accuracy for 90\%, only certain levels of (dis-)agreement are possible. Error consistency (measured by $\kappa$) aims to correct for accuracy and thus in our experiments different kinds of correlations between error consistency and overall accuracy occur. We observe zero correlation in (Figures \ref{fig:results_accuracy_generalisation}a, \ref{fig:results_accuracy_generalisation}b) and positive correlation in Figure \ref{fig:results_accuracy_generalisation}c. In Figure \ref{fig:results_accuracy_generalisation}d we observe a negative correlation between accuracy and error consistency. We conclude that there is \textit{no} correlation between consistency ($\kappa$) and accuracy for independent observers whilst for dependent (consistent) observers correlations are possible. Kappa corrects for accuracy but is not independent from it.

\subsection{Method details for Brain-Score and CNNs}
\label{app:method_details_CNNs}
Human responses were compared against classification decisions of all available CNN models from the PyTorch model zoo (for \texttt{torchvision} version 0.2.2) \cite{pytorch}, namely \texttt{alexnet, vgg11-bn, vgg13-bn, vgg16-bn, vgg19-bn, squeezenet1-0, squeezenet1-1, densenet121, densenet169, densenet201, inception-v3, resnet18, resnet34, resnet50, resnet101, resnet152}. For the VGG model family \cite{simonyan2015very}, we used the implementation with batch norm. CORnet-S, an additional recurrent model \cite{kubilius2019brain} analysed in Section~\ref{subsec:algorithms_vs_algorithms}, was obtained from the author's github implementation.\footnote{\url{https://github.com/dicarlolab/CORnet}} The comparison to \texttt{Brain-Score} in Figures~\ref{fig:app_brainscore_cue_conflict}, \ref{fig:app_brainscore_edges}, \ref{fig:app_brainscore_silhouettes} and \ref{fig:app_brainscore_ImageNet} uses \texttt{Brain-Score} values obtained from the \href{http://www.brain-score.org/}{Brain-Score website}(date of download: April 17, 2020) and error consistency values obtained by us. Note that the model implementations differ slightly: we consistently used PyTorch models whereas \texttt{Brain-Score} tested models from a few different frameworks (the full list can be seen \href{https://github.com/brain-score/candidate_models/blob/745f9d1bc747e936cbdef1e7fc599b16cf9dc677/candidate_models/base_models/__init__.py#L350}{here}). Namely, \texttt{squeezenet1-0, squeezenet1-1, resnet18, resnet-34} are identical (PyTorch); the VGG models use Keras instead (without batch norm) and so do the \texttt{Brain-Score} DenseNet models; \texttt{inception\_v3, resnet50\_v1, resnet101\_v1, resnet152\_v1} are TFSlim models. Since model implementations usually differ slightly across frameworks, a small variation in the results can be expected depending on the chosen model and framework.

\subsection{Error consistency of shape-biased models}
\label{app:consistency_shape_biased_models}
We analyzed three CNNs with different degrees of stylized training data from \cite{geirhos2019imagenettrained}. Model shape bias predicts human-CNN error consistency for cue conflict stimuli, indicating that networks basing their decisions on object shape (rather than texture) make more human-like errors:

\begin{tabular}{ r | r | r | r | r}
  \hline			
  model shape bias (\%) & 20.5 & 21.4 & 34.7 & 81.4 \\
  human-CNN consistency ($\kappa$) & .066 & .068 & .098 & .195 \\
  \hline  
\end{tabular}

\begin{table}[ht]
\centering
\begin{tabular}{rlccc}
  \hline
 & observer / model & cue conflict & edge & silhouette \\ 
  \hline
  1 & subject-01 & 0.69 & 0.89 & 0.80 \\ 
  2 & subject-02 & 0.76 & 0.94 & 0.66 \\ 
  3 & subject-03 & 0.84 & 0.93 & 0.80 \\ 
  4 & subject-04 & 0.62 & 0.84 & 0.78 \\ 
  5 & subject-05 & 0.85 & 0.89 & 0.77 \\ 
  6 & subject-06 & 0.82 & 0.93 & 0.72 \\ 
  7 & subject-07 & 0.76 & 0.81 & 0.76 \\ 
  8 & subject-08 & 0.78 & 0.96 & 0.64 \\ 
  9 & subject-09 & 0.86 & 0.61 & 0.76 \\ 
  10 & subject-10 & 0.77 & 0.92 & 0.85 \\ 
  \hline
  11 & alexnet & 0.19 & 0.29 & 0.43 \\ 
  12 & vgg11-bn & 0.12 & 0.14 & 0.46 \\ 
  13 & vgg13-bn & 0.12 & 0.25 & 0.36 \\ 
  14 & vgg16-bn & 0.14 & 0.22 & 0.47 \\ 
  15 & vgg19-bn & 0.15 & 0.28 & 0.46 \\ 
  16 & squeezenet1-0 & 0.14 & 0.15 & 0.24 \\ 
  17 & squeezenet1-1 & 0.17 & 0.14 & 0.29 \\ 
  18 & densenet121 & 0.19 & 0.24 & 0.42 \\ 
  19 & densenet169 & 0.21 & 0.33 & 0.53 \\ 
  20 & densenet201 & 0.21 & 0.38 & 0.51 \\ 
  21 & inception-v3 & 0.27 & 0.28 & 0.54 \\ 
  22 & resnet18 & 0.19 & 0.20 & 0.47 \\ 
  23 & resnet34 & 0.19 & 0.16 & 0.45 \\ 
  24 & resnet50 & 0.18 & 0.14 & 0.54 \\ 
  25 & resnet101 & 0.20 & 0.24 & 0.49 \\ 
  26 & resnet152 & 0.21 & 0.21 & 0.56 \\ 
  \hline
  27 & cornet-s & 0.18 & 0.25 & 0.46 \\ 
  \hline
\end{tabular}
\caption{Accuracies for human observers and CNNs for all three experiments. In the cue conflict experiment case, an answer is counted as correct in this table if this answer corresponds to the correct shape category (other choices are possible).}
\label{app:table_accuracies}
\end{table}

\newcommand{\figwidthV}{0.255\textwidth} %
\newcommand{\figwidthVI}{0.23\textwidth} %
\newcommand{\captionspaceGeneralisationIII}{-1.3\baselineskip} %
\newcommand{\captionspaceGeneralisationIV}{0.5\baselineskip} %
\begin{figure}
    \begin{subfigure}{\figwidthV}
        \centering
        \includegraphics[width=\linewidth]{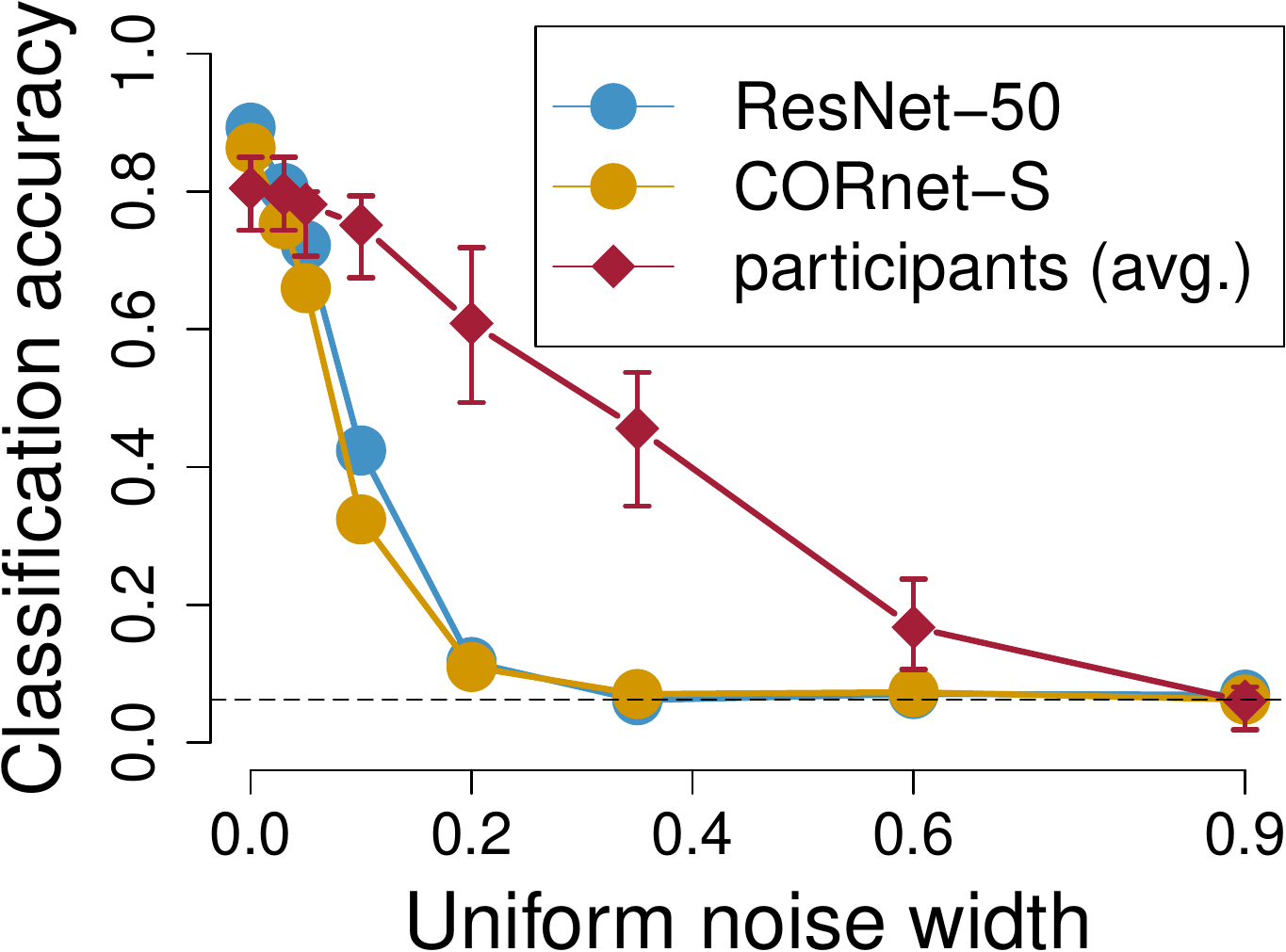}
        \vspace{\captionspaceGeneralisationIII}
        \caption{Uniform noise}
        \vspace{\captionspaceGeneralisationIV}
    \end{subfigure}\hfill
    \begin{subfigure}{\figwidthVI}
        \centering
        \includegraphics[width=\linewidth]{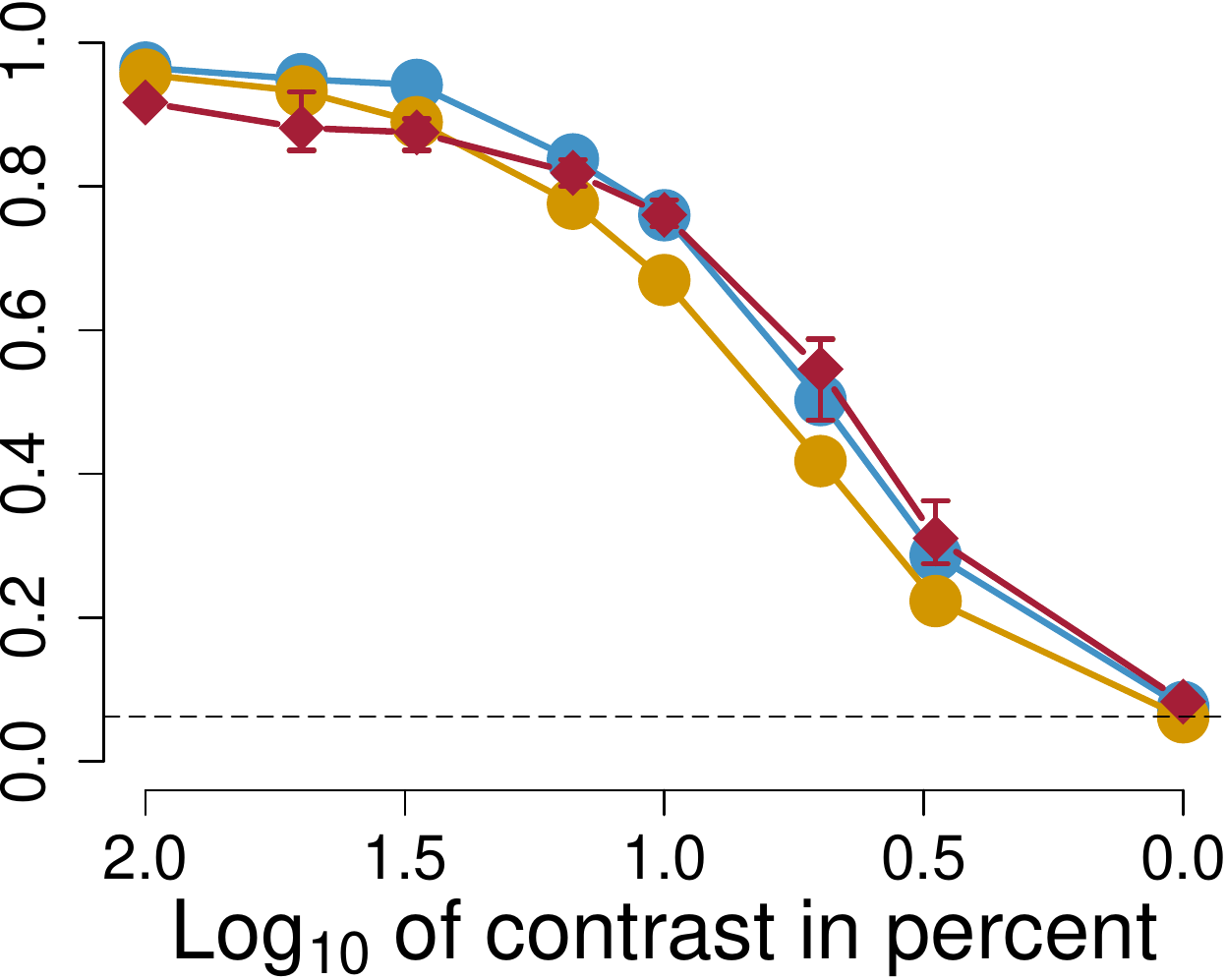}
        \vspace{\captionspaceGeneralisationIII}
        \caption{Contrast}
        \vspace{\captionspaceGeneralisationIV}
    \end{subfigure}\hfill
    \begin{subfigure}{\figwidthVI}
        \centering
        \includegraphics[width=\linewidth]{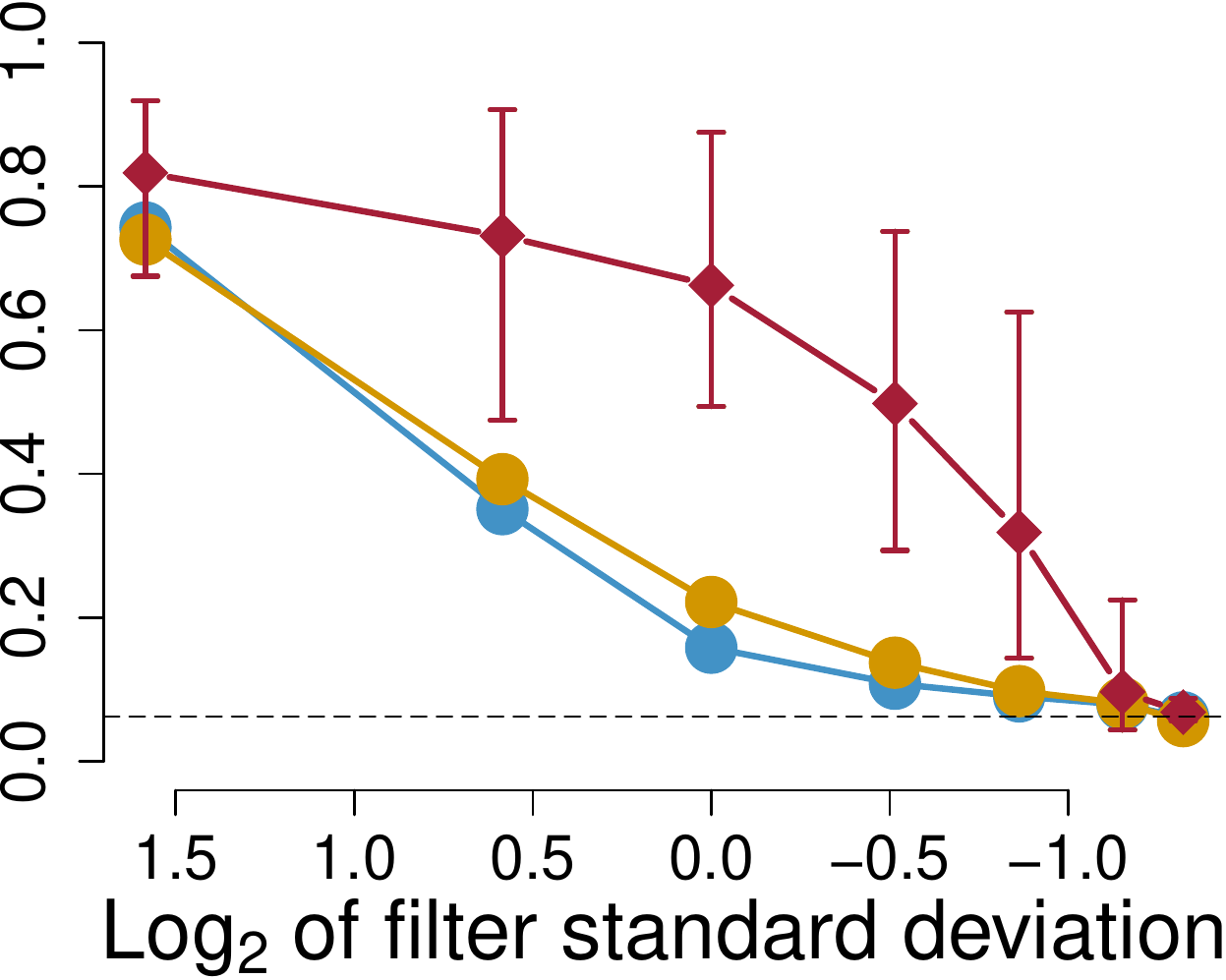}
        \vspace{\captionspaceGeneralisationIII}
        \caption{High-pass}
        \vspace{\captionspaceGeneralisationIV}
    \end{subfigure}\hfill
    \begin{subfigure}{\figwidthVI}
        \centering
        \includegraphics[width=\linewidth]{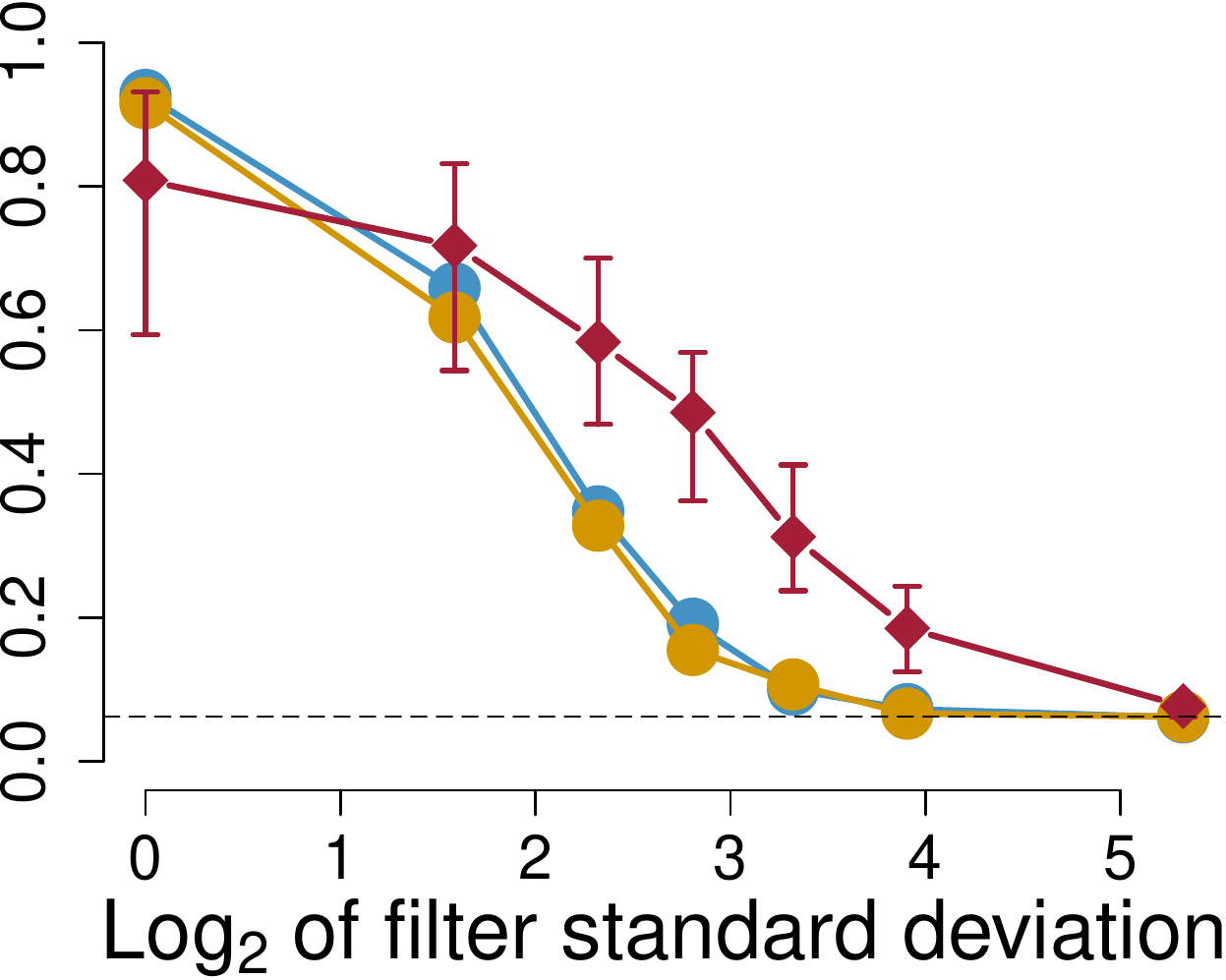}
        \vspace{\captionspaceGeneralisationIII}
        \caption{Low-pass}
        \vspace{\captionspaceGeneralisationIV}
    \end{subfigure}\hfill

    \begin{subfigure}{\figwidthV}
        \centering
        \includegraphics[width=\linewidth]{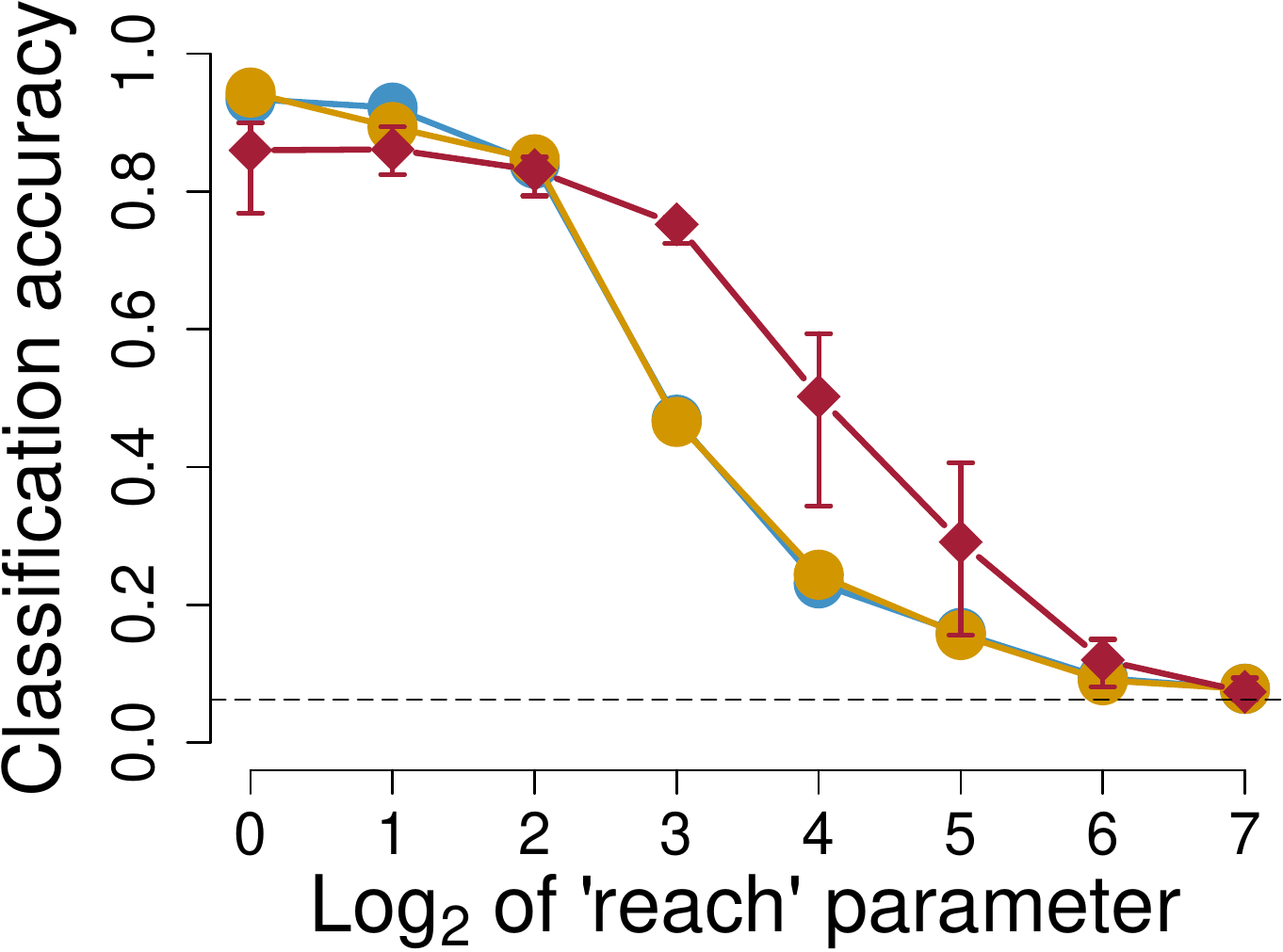}
        \vspace{\captionspaceGeneralisationIII}
        \caption{Eidolon I}
    \end{subfigure}\hfill
    \begin{subfigure}{\figwidthVI}
        \centering
        \includegraphics[width=\linewidth]{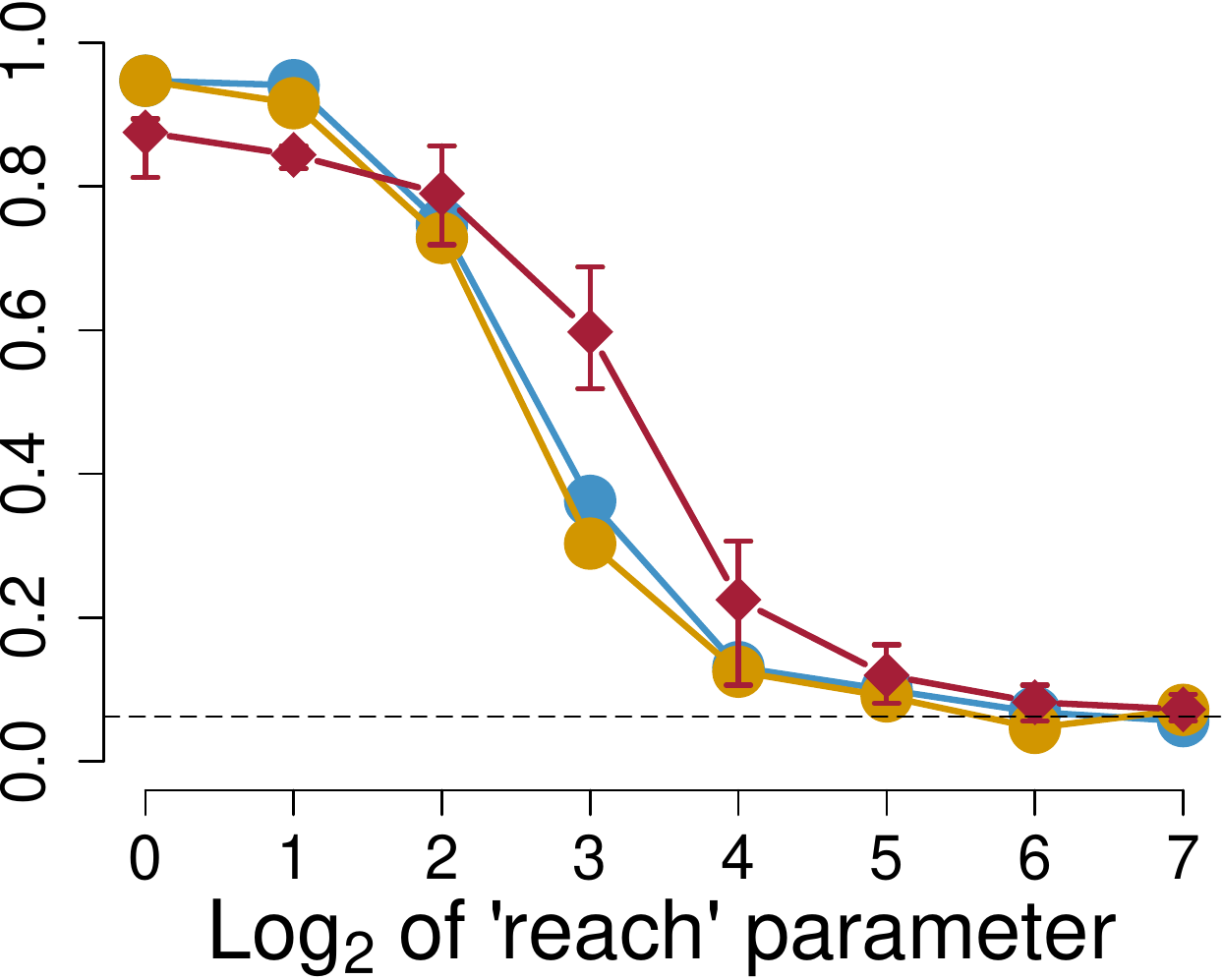}
        \vspace{\captionspaceGeneralisationIII}
        \caption{Eidolon II}
    \end{subfigure}\hfill
    \begin{subfigure}{\figwidthVI}
        \centering
        \includegraphics[width=\linewidth]{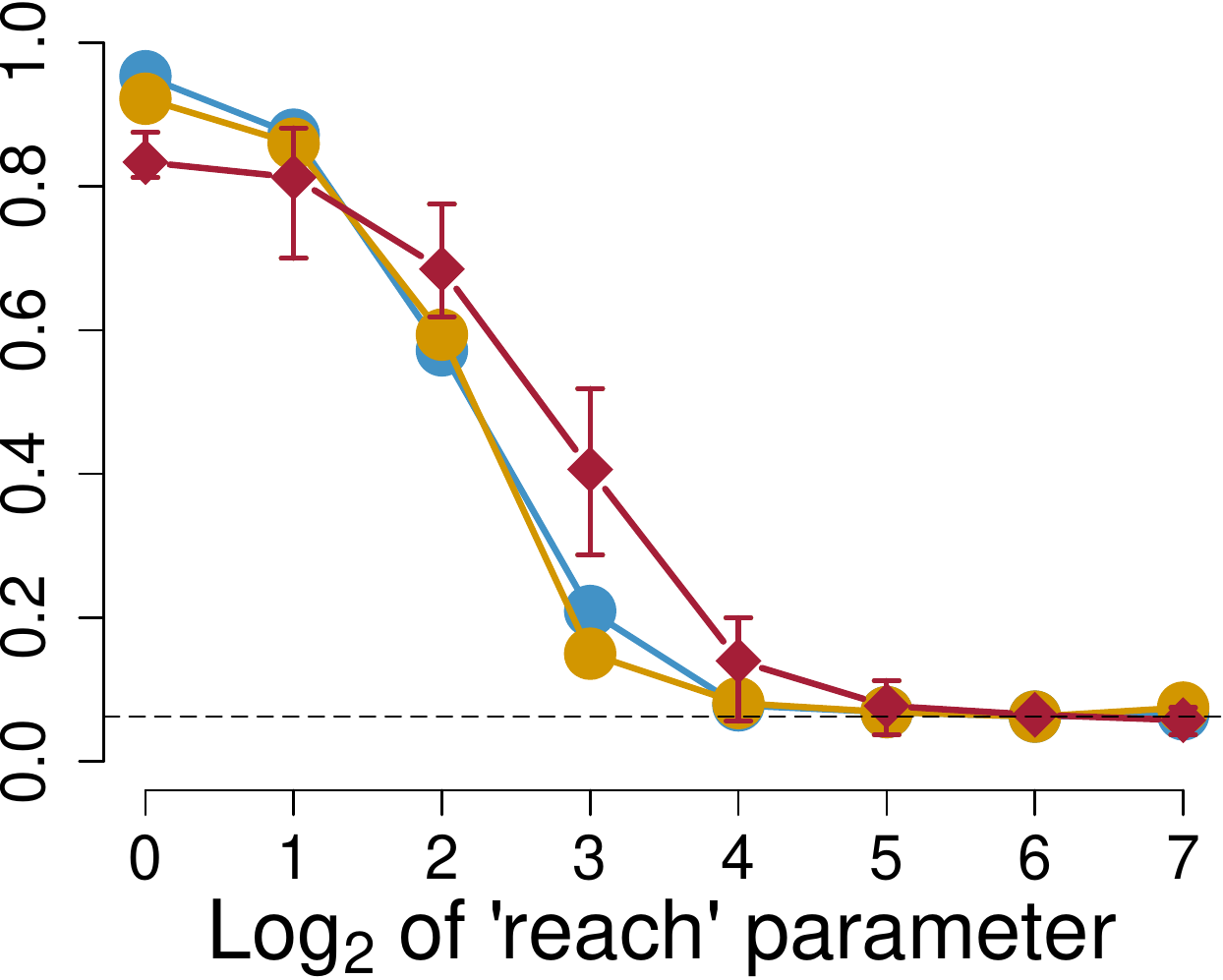}
        \vspace{\captionspaceGeneralisationIII}
        \caption{Eidolon III}
    \end{subfigure}\hfill
    \begin{subfigure}{\figwidthVI}
        \centering
        \includegraphics[width=\linewidth]{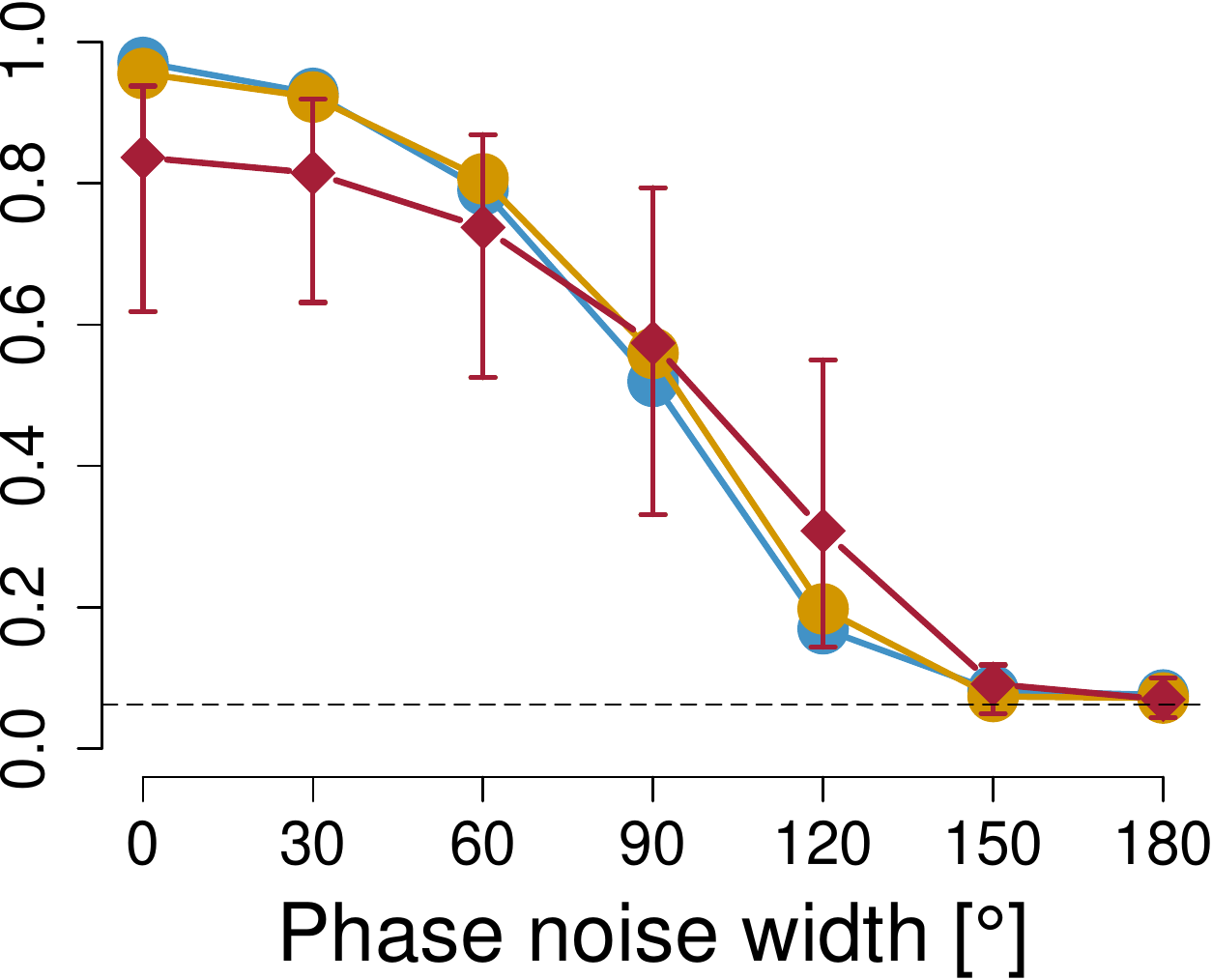}
        \vspace{\captionspaceGeneralisationIII}
        \caption{Phase noise}
    \end{subfigure}\hfill
    \caption{Classification accuracy on parametrically distorted images for ResNet-50, CORnet-S and human observers. Again, CORnet-S behaves like a ResNet-50 rather than like human observers.}
    \label{fig:app_noise_generalisation}
\end{figure}

\newcommand{\figwidthVII}{0.32\textwidth}
\begin{figure}
    \begin{subfigure}{\figwidthVII}
        \centering
        \caption*{Human average}
        \includegraphics[width=\linewidth]{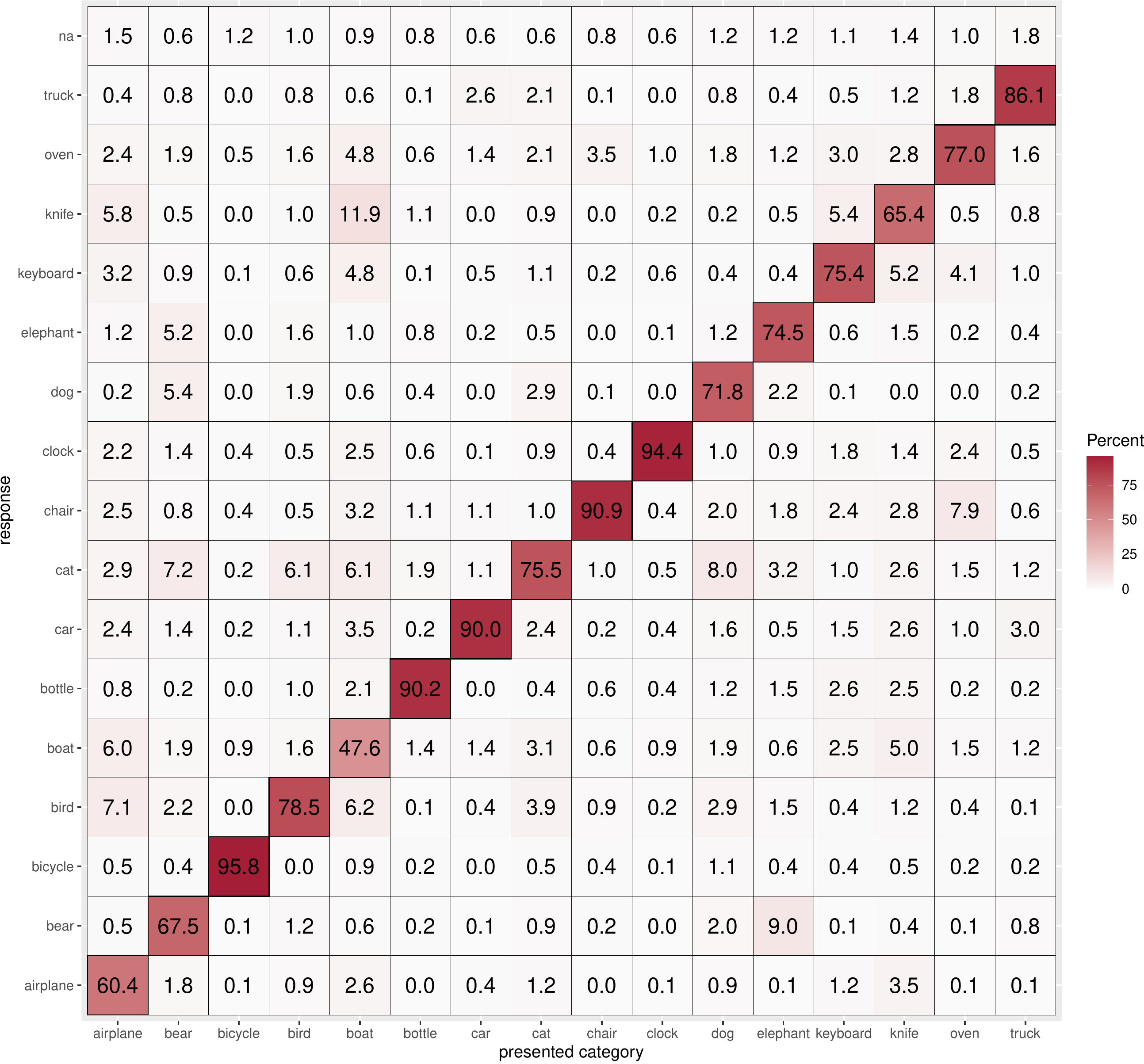}
    \end{subfigure}\hfill
    \begin{subfigure}{\figwidthVII}
        \centering
        \caption*{ResNet-50}
        \includegraphics[width=\linewidth]{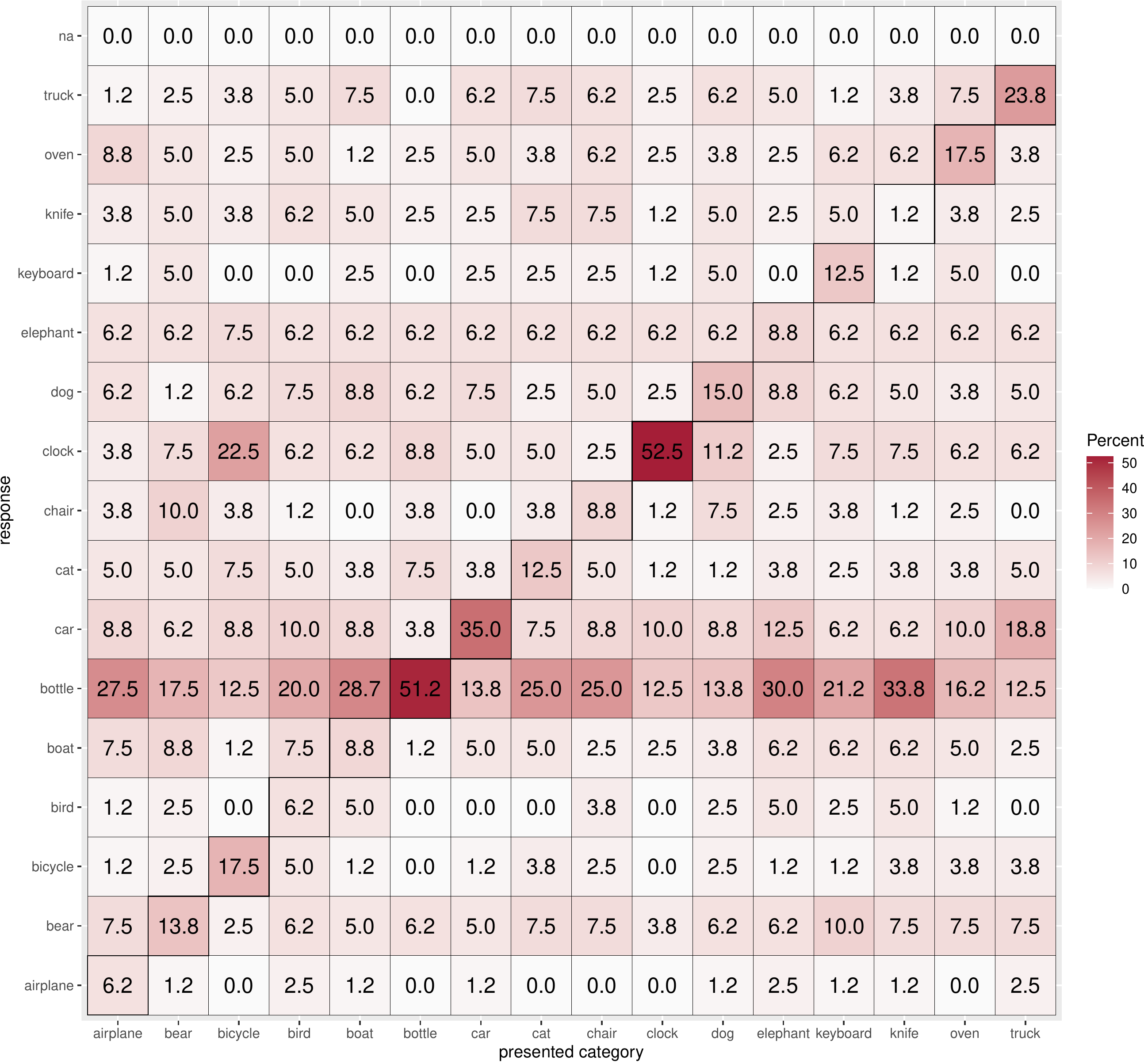}
    \end{subfigure}\hfill
    \begin{subfigure}{\figwidthVII}
        \centering
        \caption*{CORnet-S}
        \includegraphics[width=\linewidth]{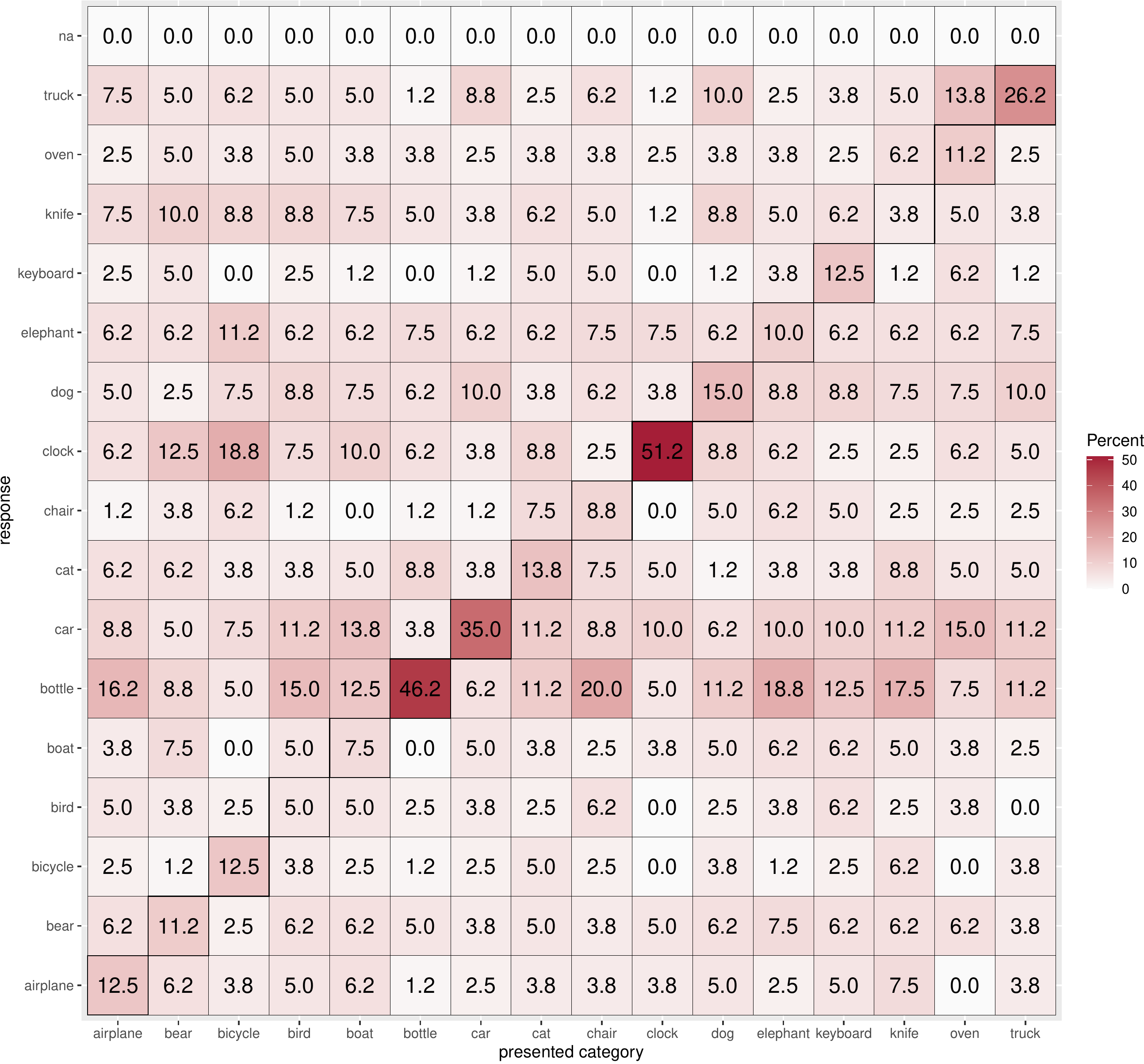}
    \end{subfigure}\hfill

    \begin{subfigure}{\figwidthVII}
        \centering
        \includegraphics[width=\linewidth]{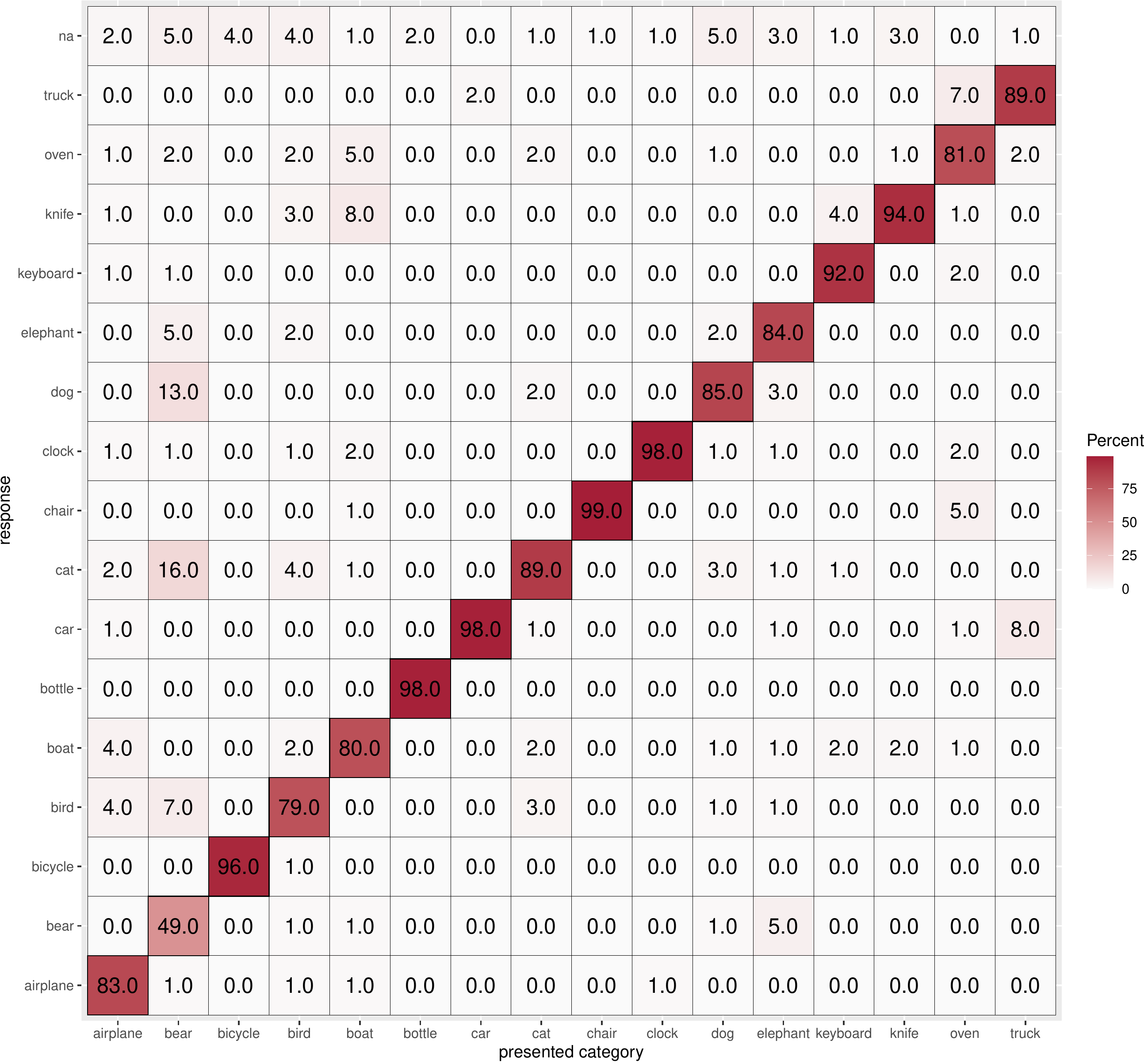}
    \end{subfigure}\hfill
    \begin{subfigure}{\figwidthVII}
        \centering
        \includegraphics[width=\linewidth]{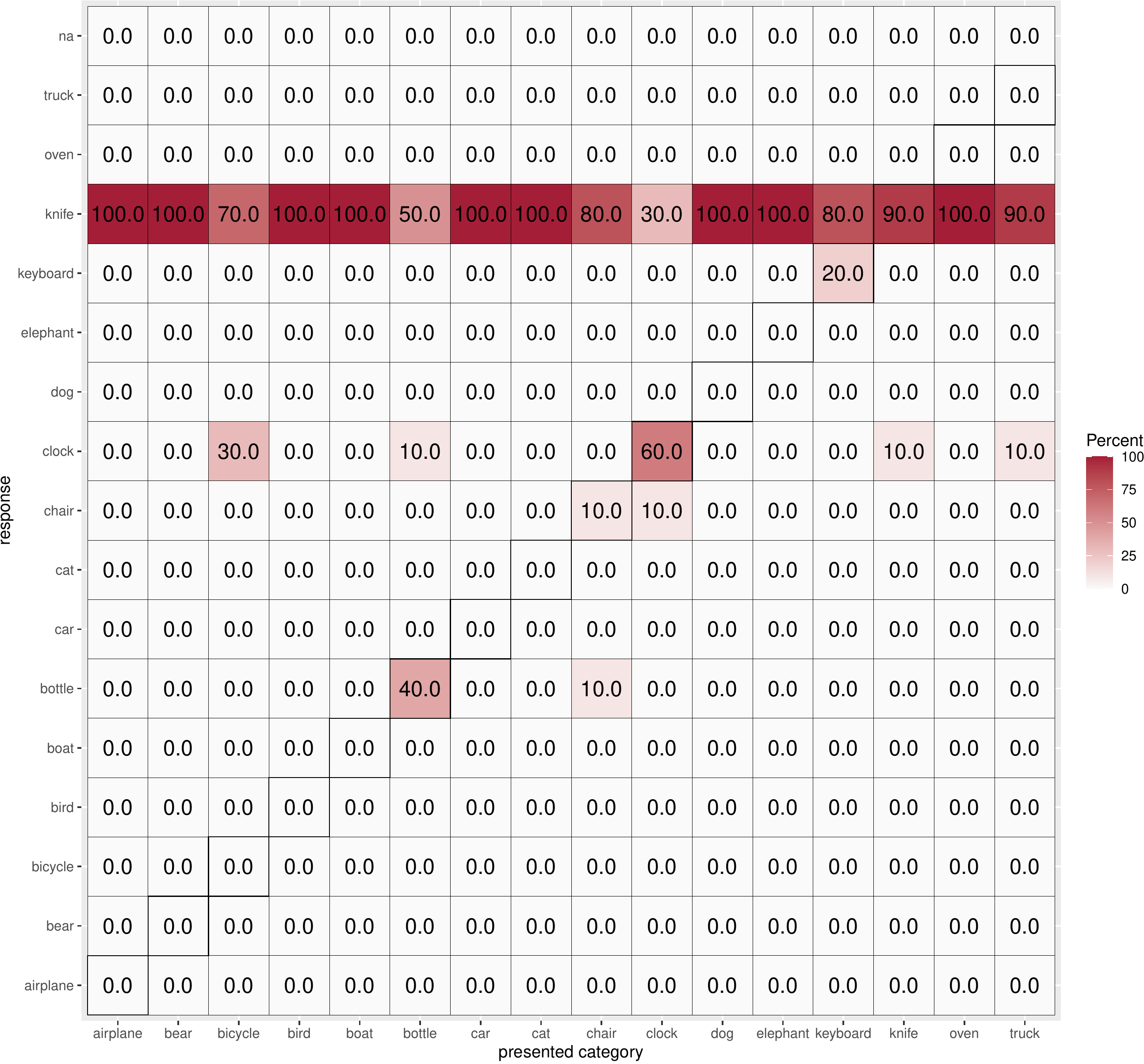}
    \end{subfigure}\hfill
    \begin{subfigure}{\figwidthVII}
        \centering
        \includegraphics[width=\linewidth]{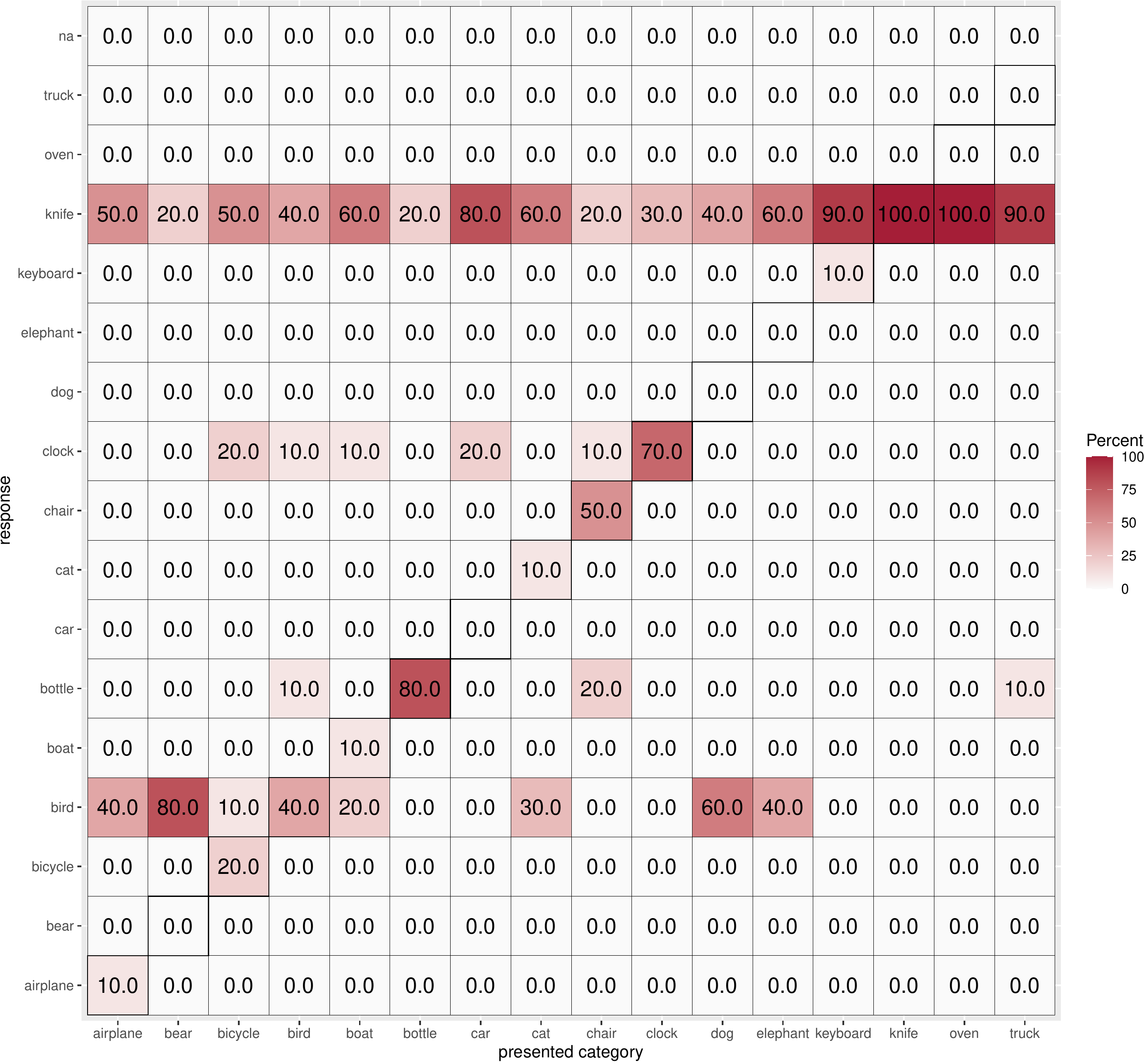}
    \end{subfigure}\hfill

    \begin{subfigure}{\figwidthVII}
        \centering
        \includegraphics[width=\linewidth]{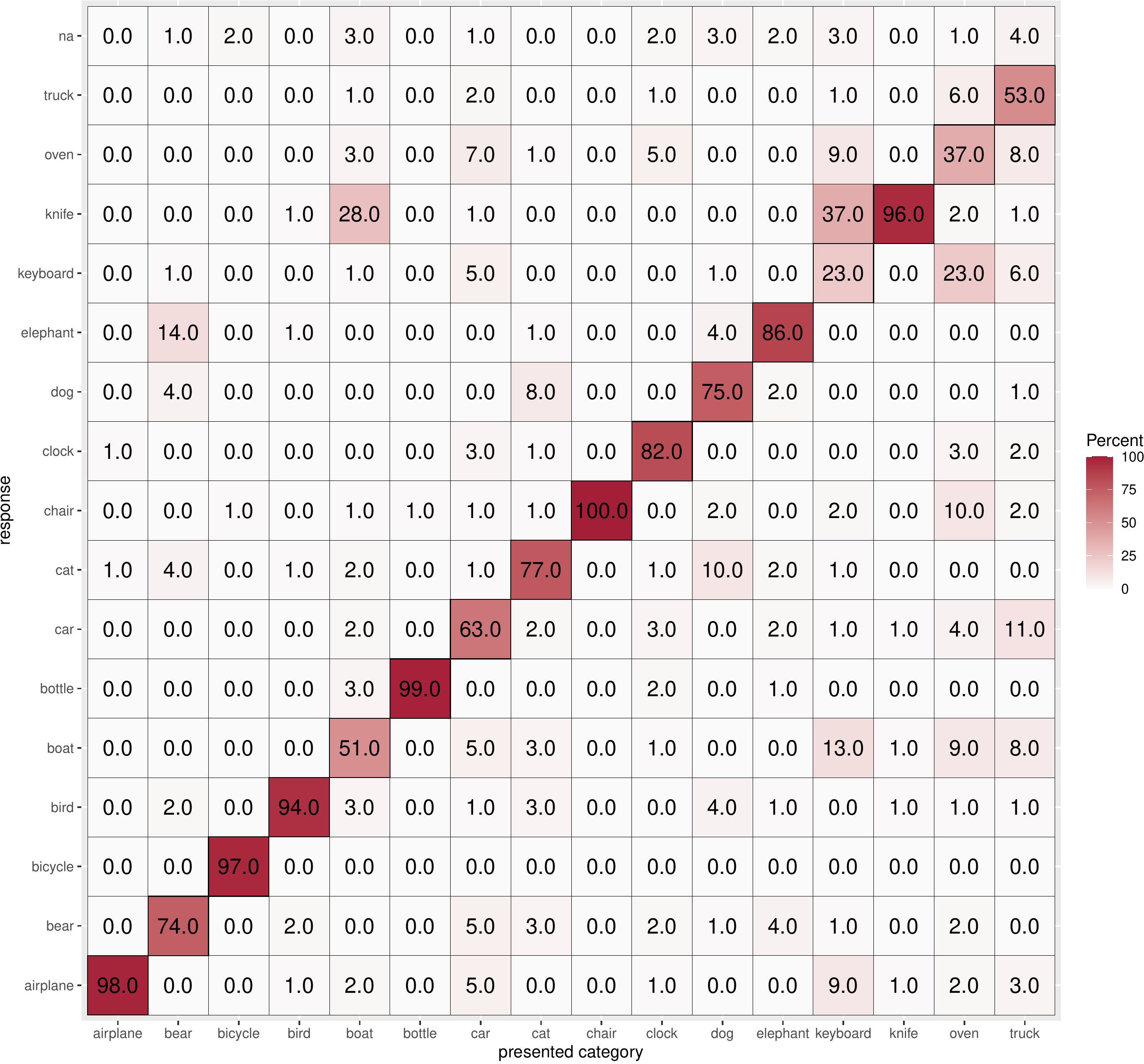}
    \end{subfigure}\hfill
    \begin{subfigure}{\figwidthVII}
        \centering
        \includegraphics[width=\linewidth]{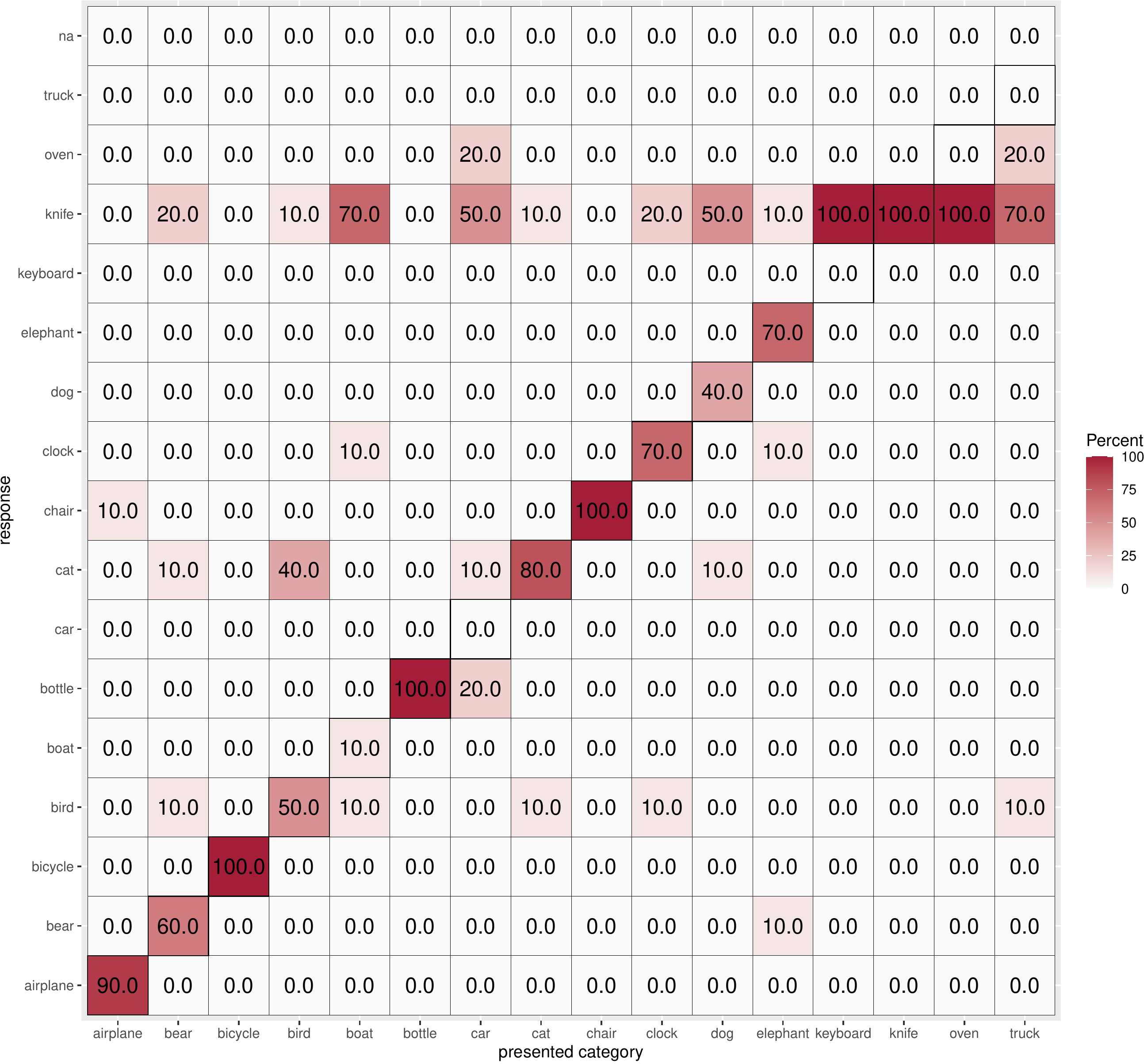}
    \end{subfigure}\hfill
    \begin{subfigure}{\figwidthVII}
        \centering
        \includegraphics[width=\linewidth]{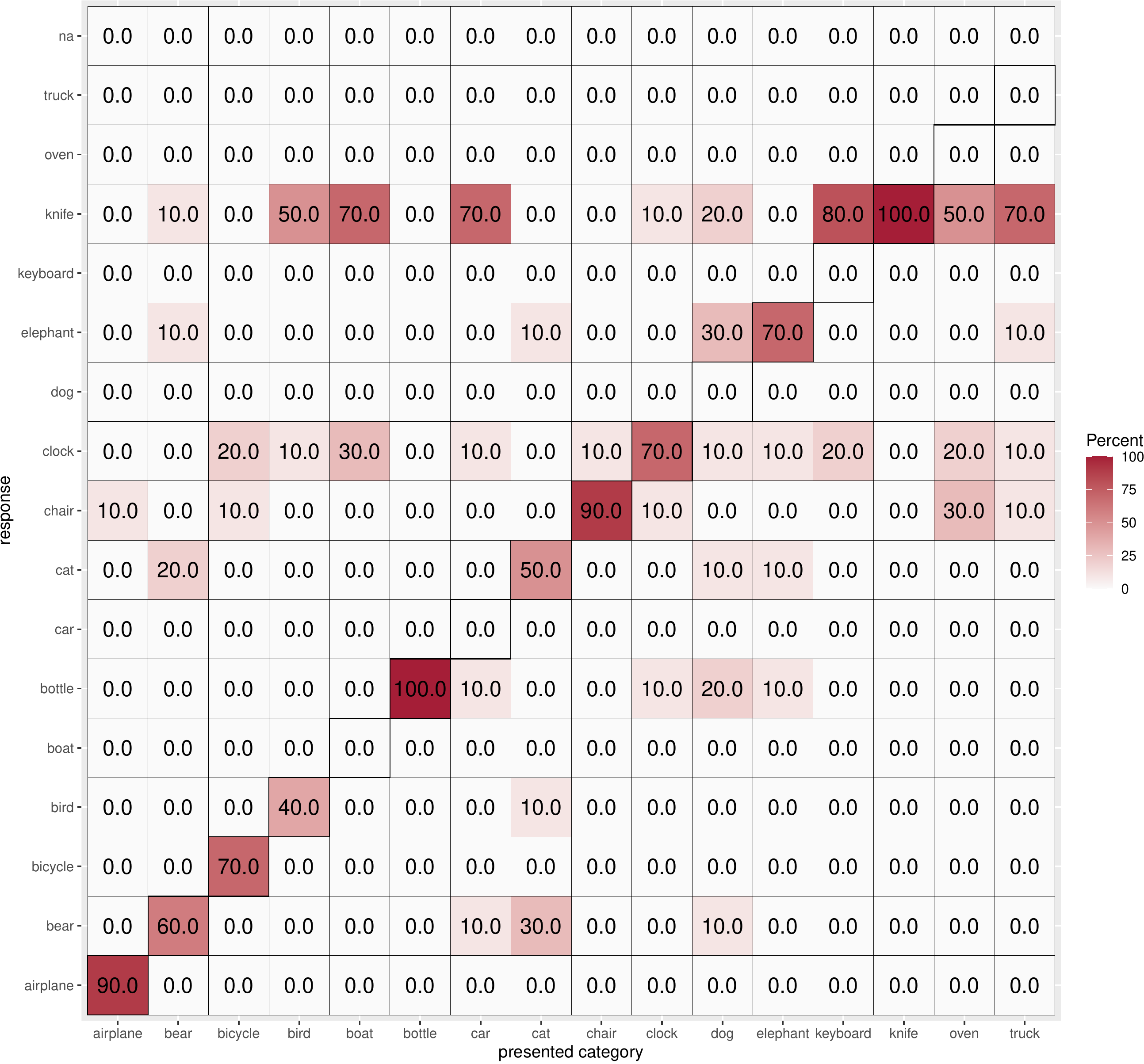}
    \end{subfigure}\hfill

    \begin{subfigure}{\figwidthVII}
        \centering
        \includegraphics[width=\linewidth]{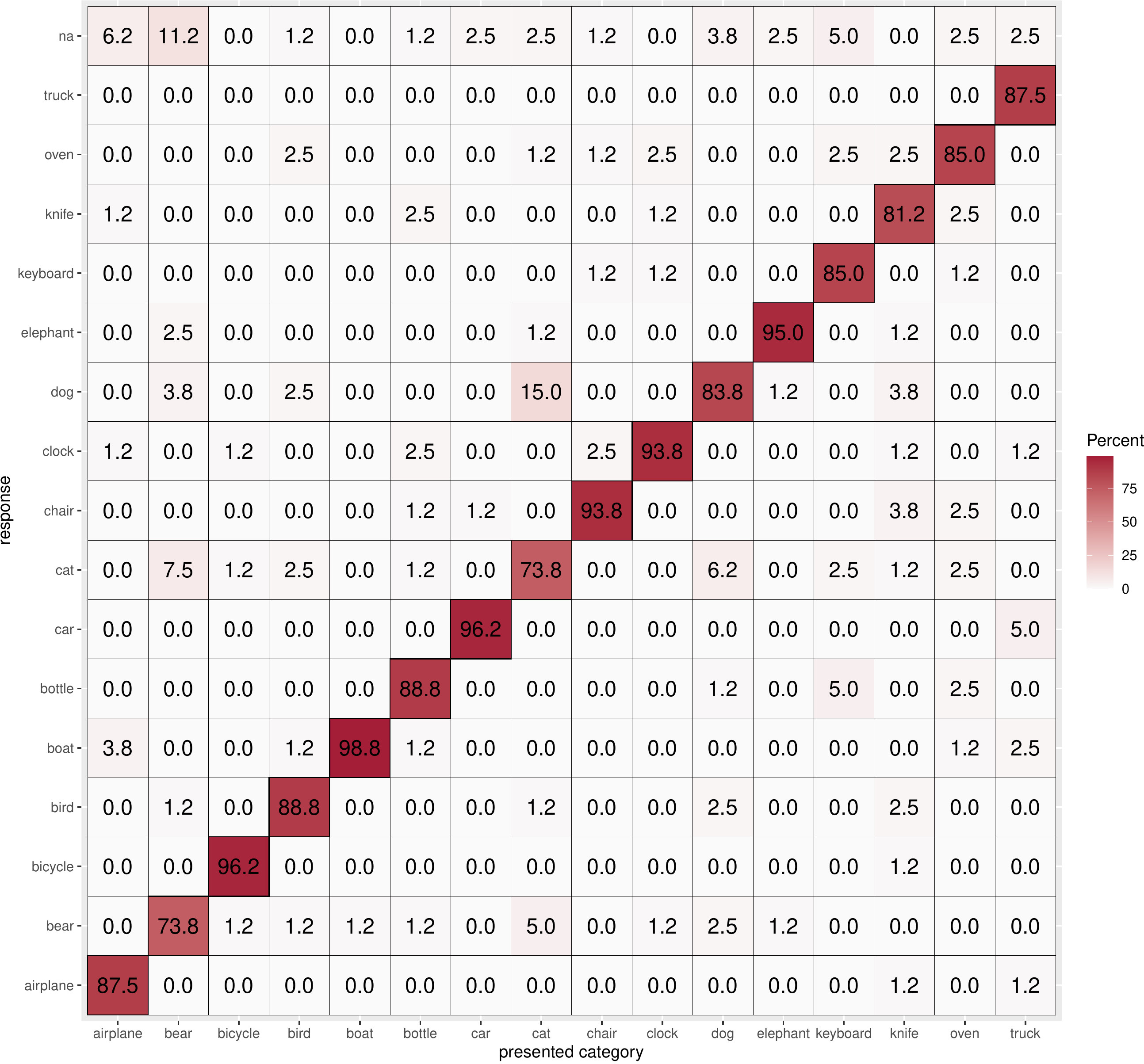}
    \end{subfigure}\hfill
    \begin{subfigure}{\figwidthVII}
        \centering
        \includegraphics[width=\linewidth]{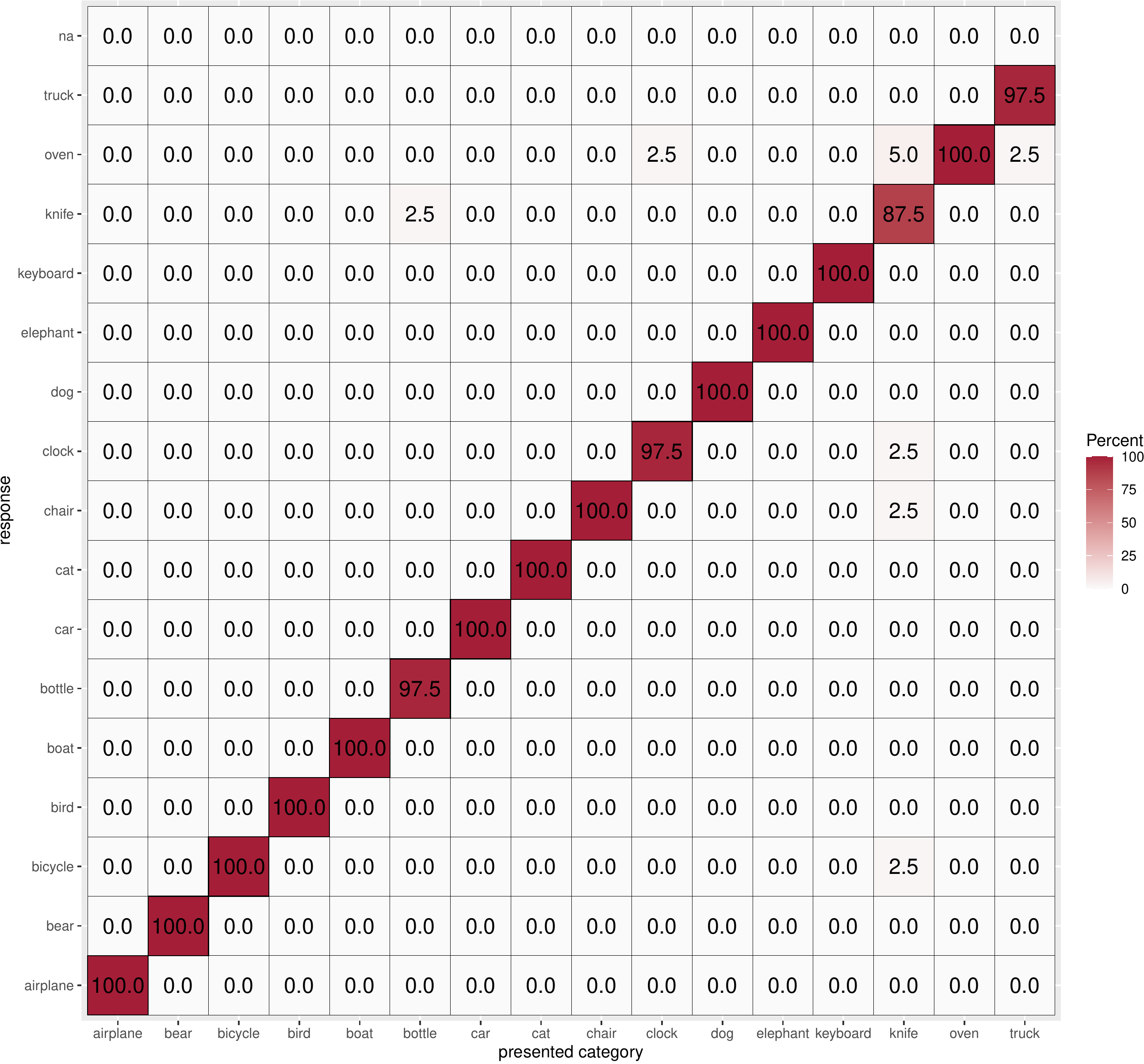}
    \end{subfigure}\hfill
    \begin{subfigure}{\figwidthVII}
        \centering
        \includegraphics[width=\linewidth]{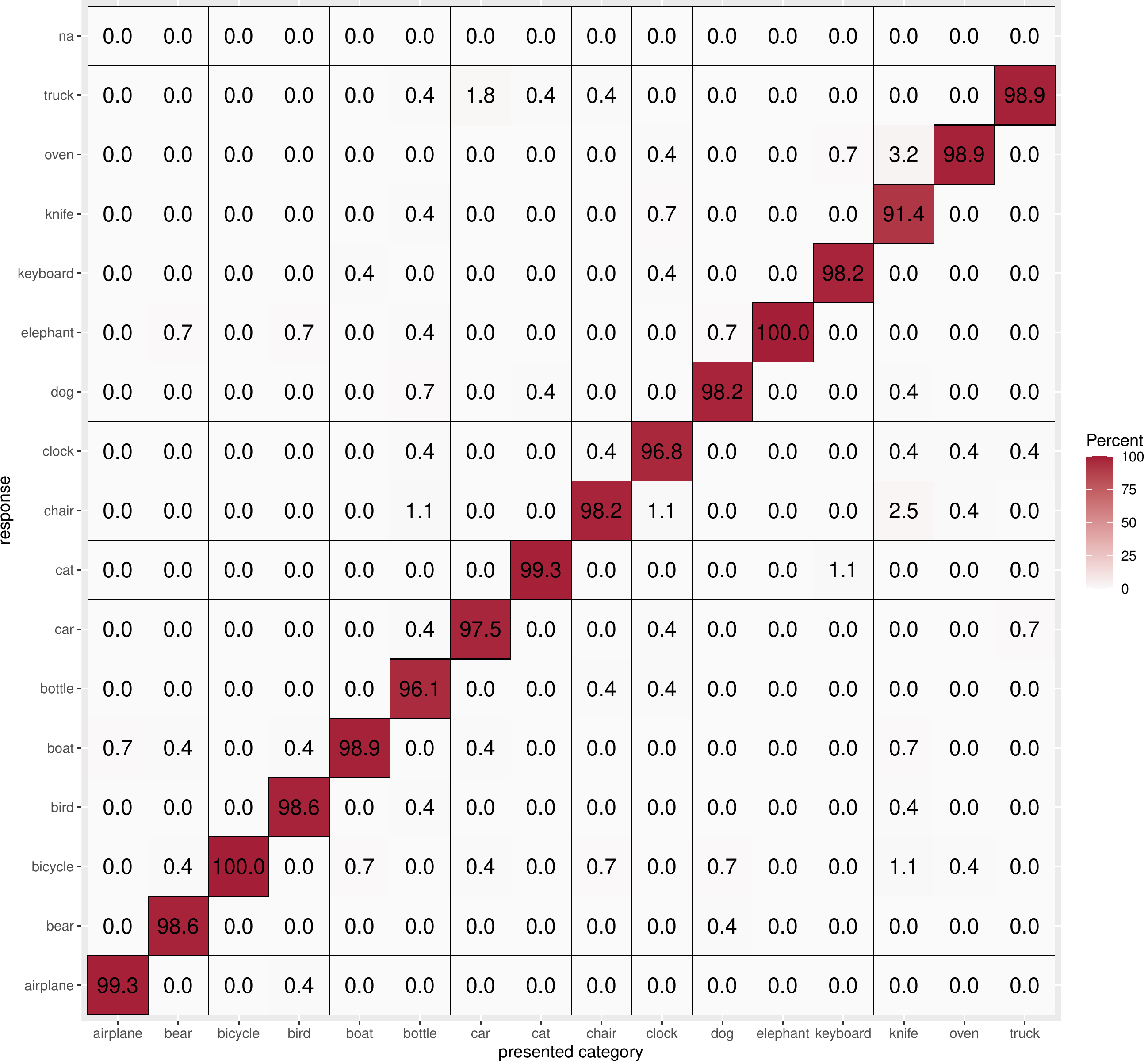}
    \end{subfigure}\hfill
    \caption{Confusion matrices for humans, ResNet-50 and CORnet-S. Different rows correspond to different experiments. Top row: cue conflict stimuli, second row: edge stimuli, third row: silhouette stimuli, last row: ImageNet stimuli.}
    \label{fig:app_confusion}
\end{figure}

\newcommand{\figwidthIII}{0.48\textwidth} %
\newcommand{\figwidthIV}{0.48\textwidth} %
\newcommand{\vSpaceI}{10pt} %

\begin{figure}[h!]
    \begin{subfigure}{\figwidthIII}
        \centering
        \includegraphics[width=\linewidth]{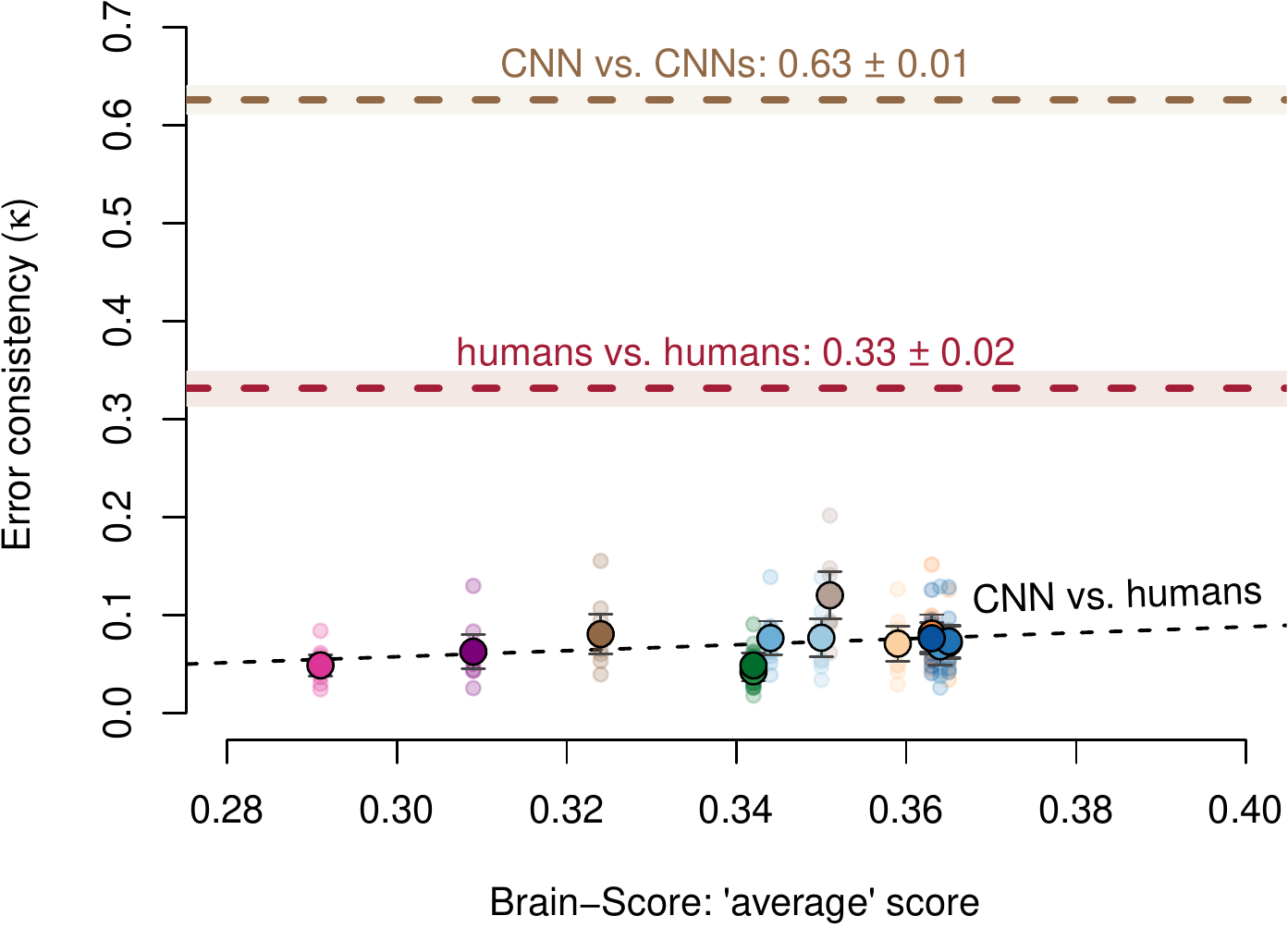}
    \end{subfigure}\hfill
    \vspace{\vSpaceI}
    \begin{subfigure}{\figwidthIV}
        \centering
        \includegraphics[width=\linewidth]{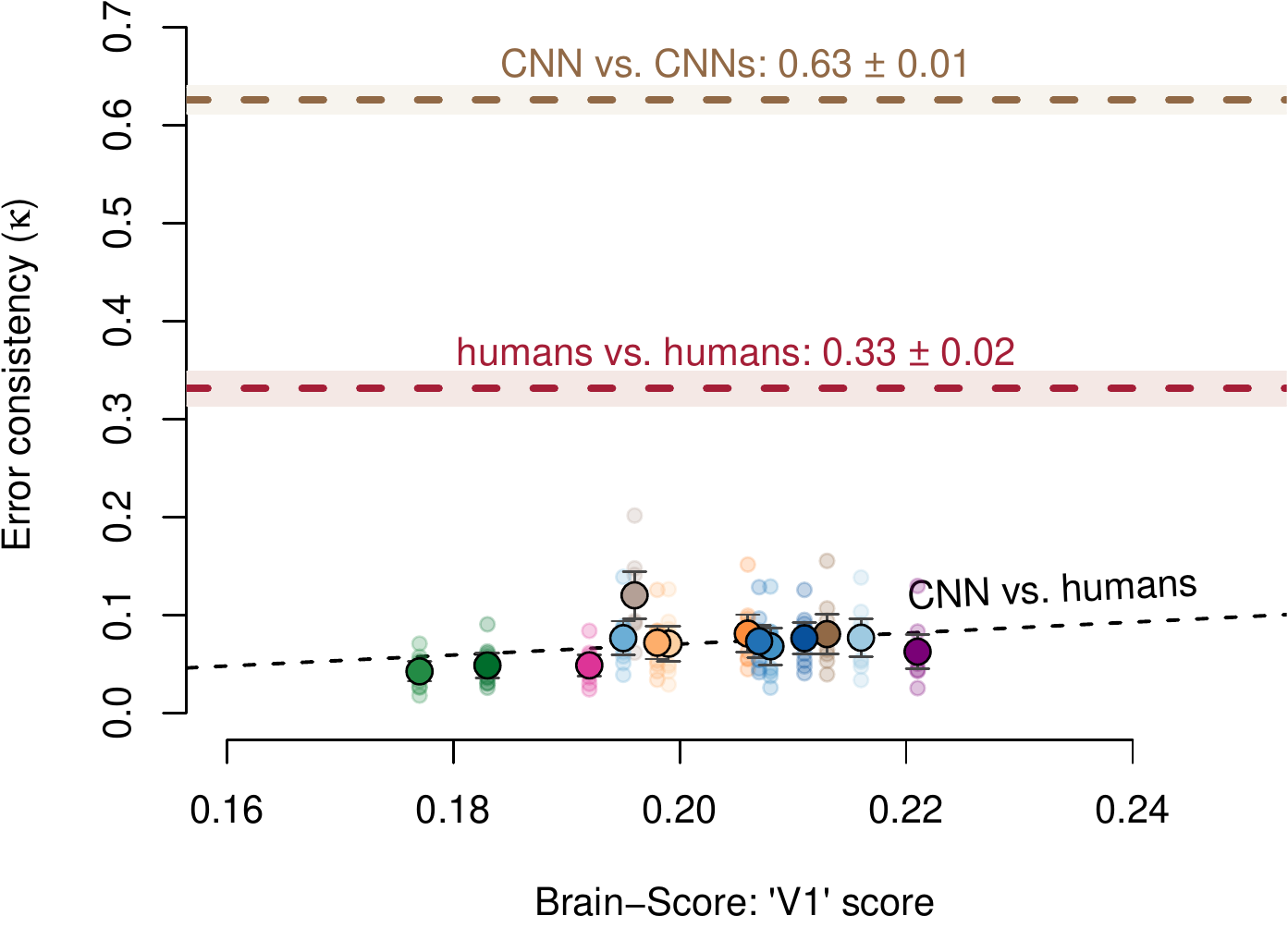}
    \end{subfigure}\hfill
    \vspace{\vSpaceI}

    \begin{subfigure}{\figwidthIII}
        \centering
        \includegraphics[width=\linewidth]{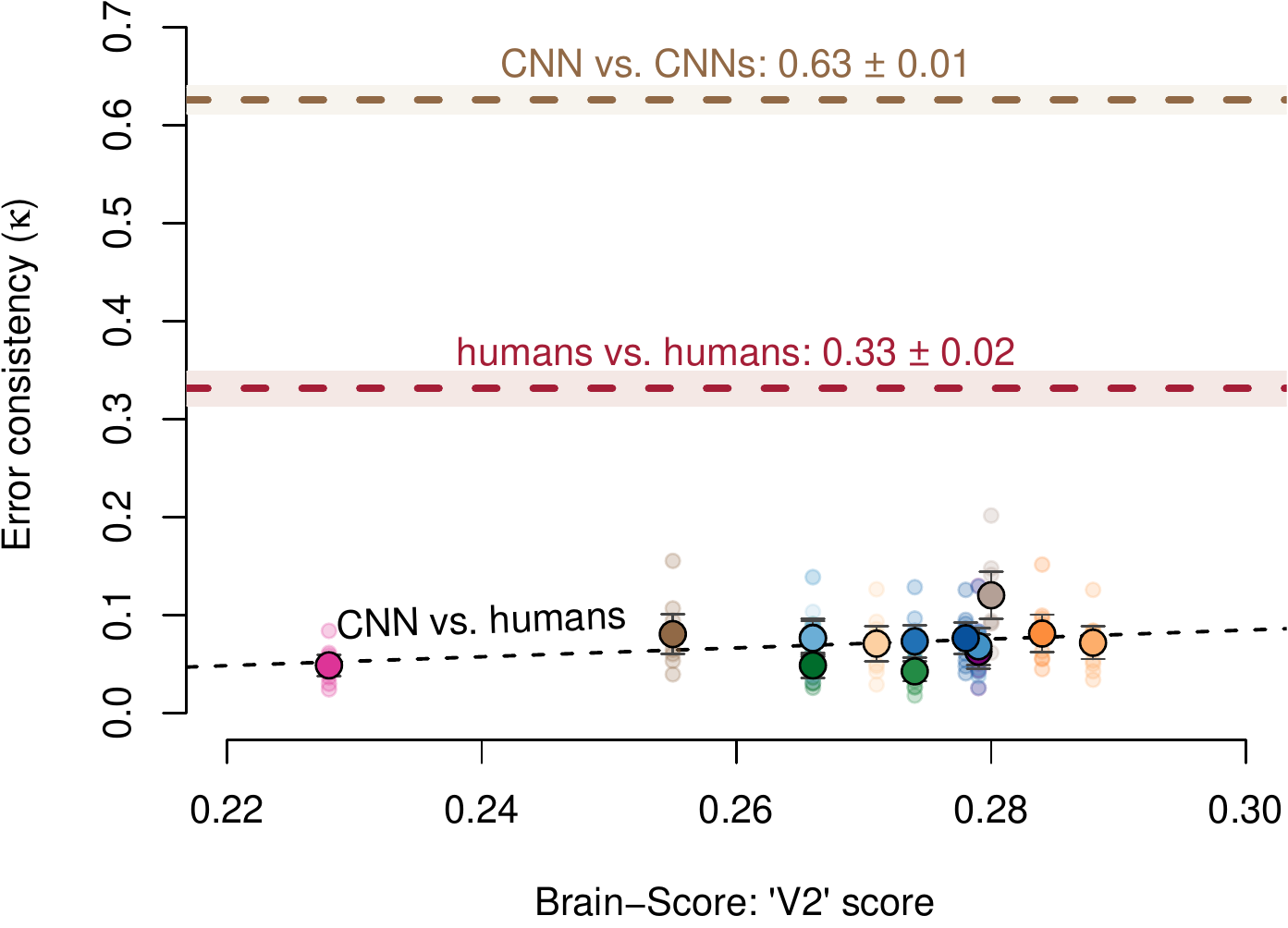}
    \end{subfigure}\hfill
    \vspace{\vSpaceI}
    \begin{subfigure}{\figwidthIV}
        \centering
        \includegraphics[width=\linewidth]{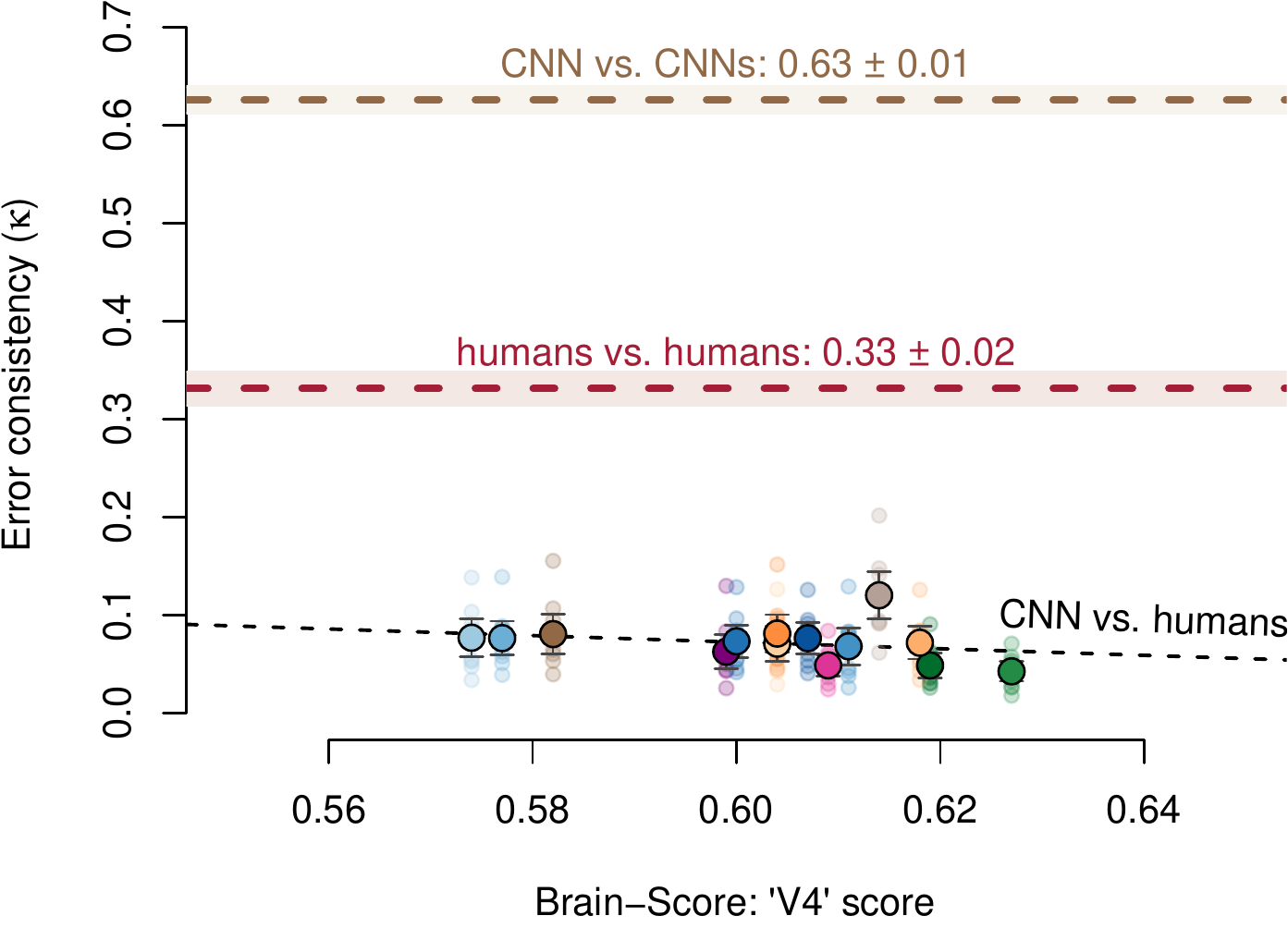}
    \end{subfigure}\hfill
    \vspace{\vSpaceI}
    
    \begin{subfigure}{\figwidthIII}
        \centering
        \includegraphics[width=\linewidth]{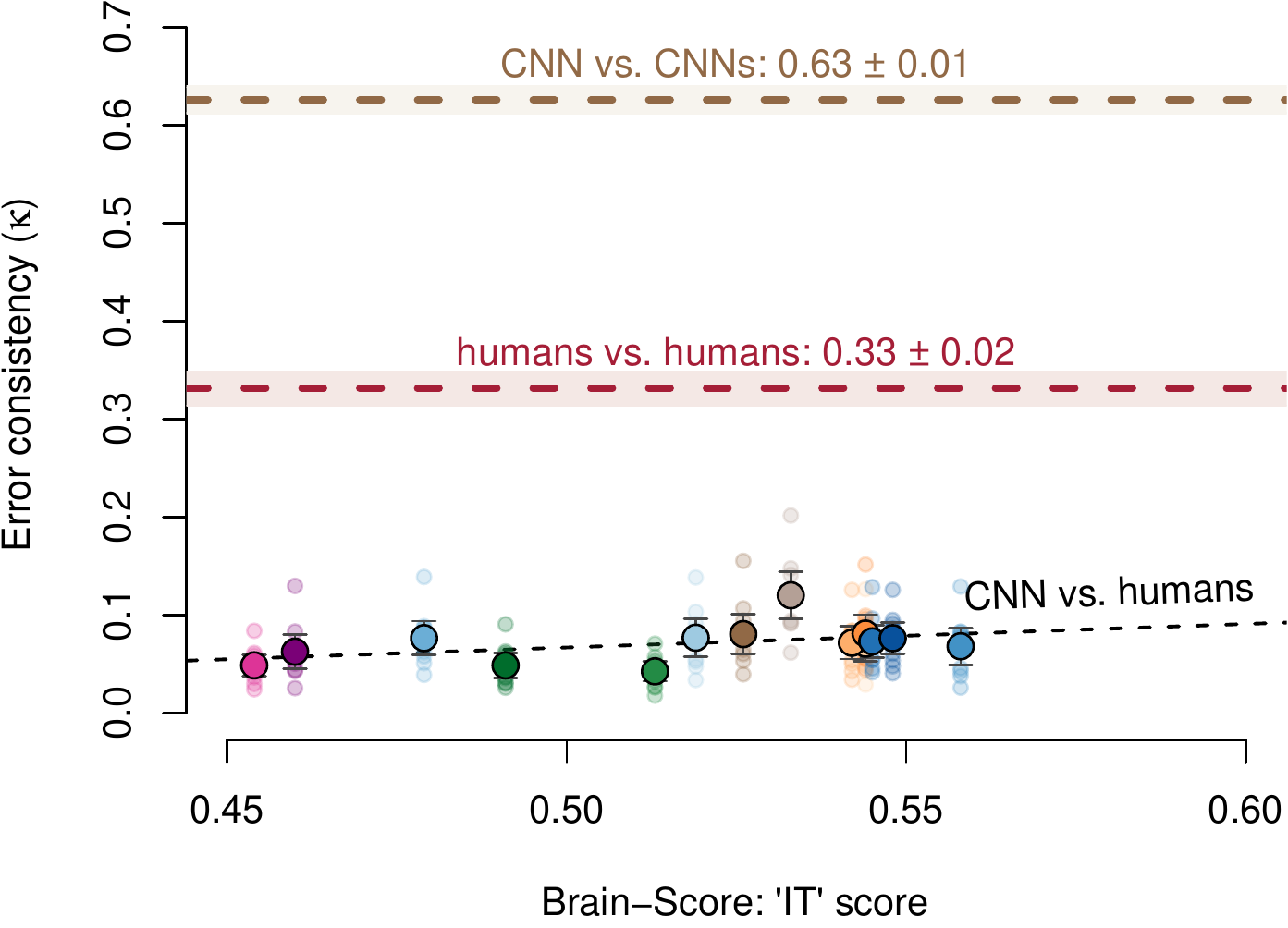}
    \end{subfigure}\hfill
    \vspace{\vSpaceI}
    \begin{subfigure}{\figwidthIV}
        \centering
        \includegraphics[width=\linewidth]{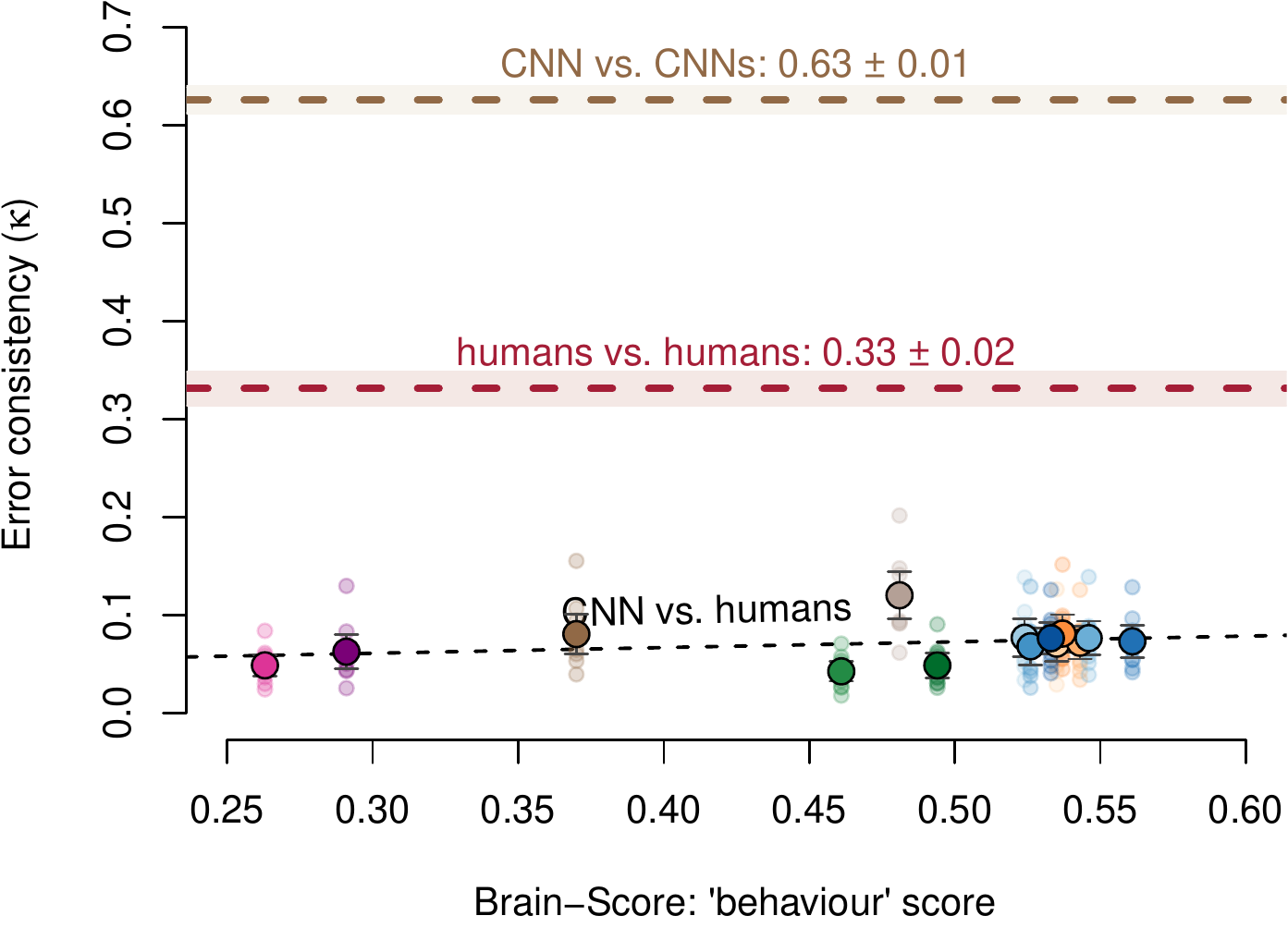}
    \end{subfigure}\hfill
    \vspace{\vSpaceI}
    \caption{Error consistency vs. \texttt{Brain-Score} metrics for PyTorch models, ``cue conflict'' stimuli.}
    \label{fig:app_brainscore_cue_conflict}
\end{figure}

\begin{figure}[h!]
    \begin{subfigure}{\figwidthIII}
        \centering
        \includegraphics[width=\linewidth]{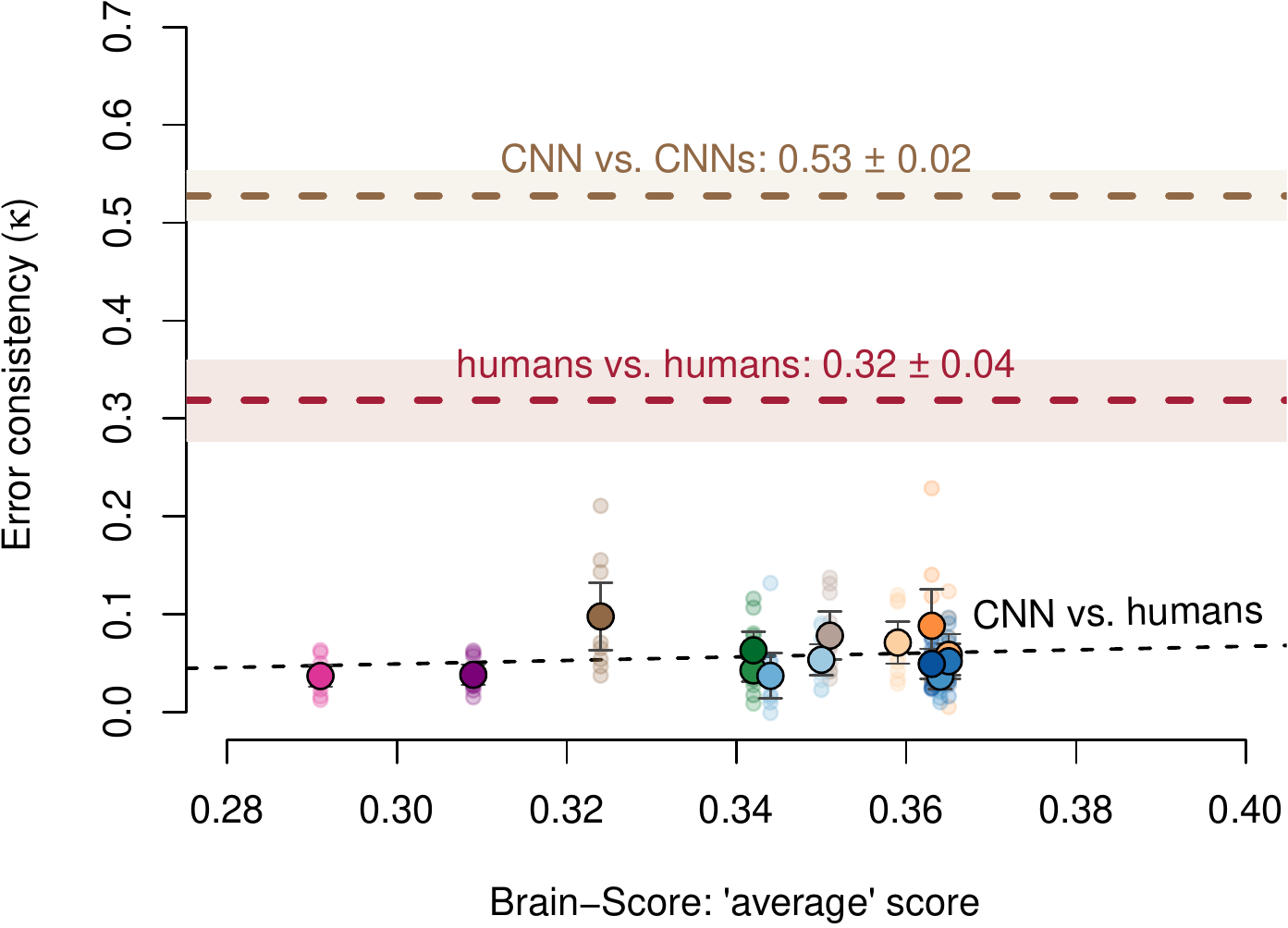}
    \end{subfigure}\hfill
    \vspace{\vSpaceI}
    \begin{subfigure}{\figwidthIV}
        \centering
        \includegraphics[width=\linewidth]{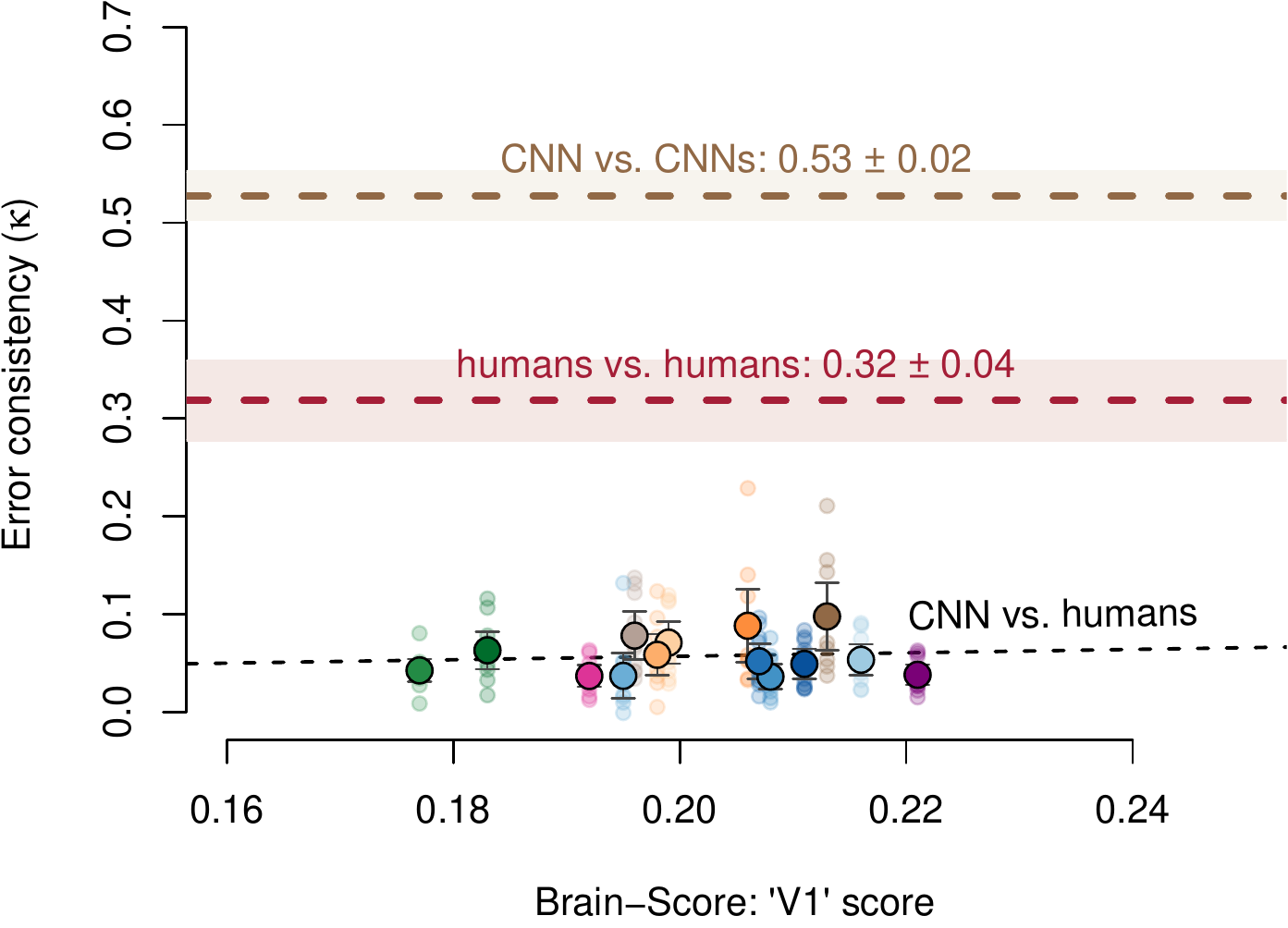}
    \end{subfigure}\hfill
    \vspace{\vSpaceI}

    \begin{subfigure}{\figwidthIII}
        \centering
        \includegraphics[width=\linewidth]{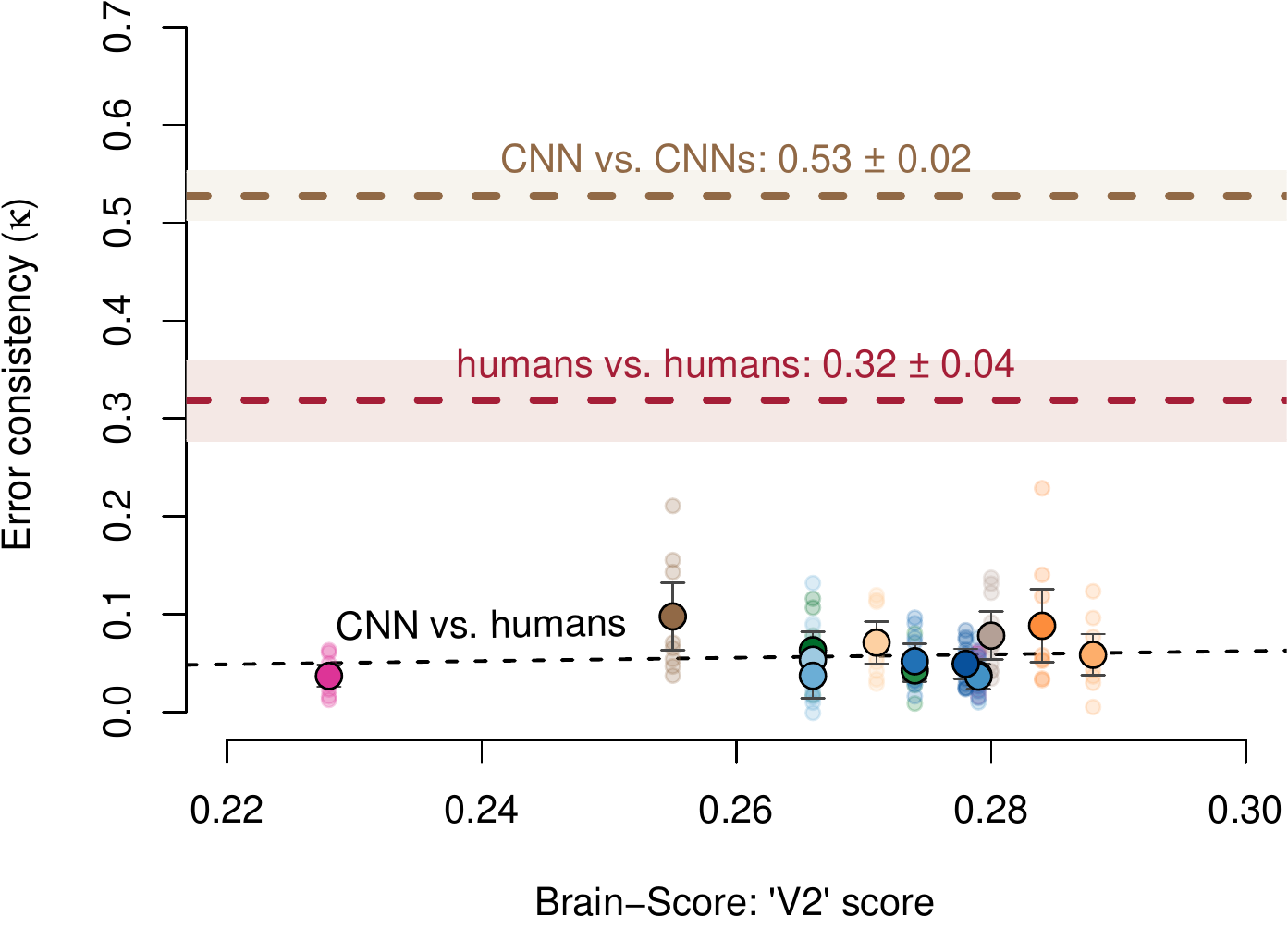}
    \end{subfigure}\hfill
    \vspace{\vSpaceI}
    \begin{subfigure}{\figwidthIV}
        \centering
        \includegraphics[width=\linewidth]{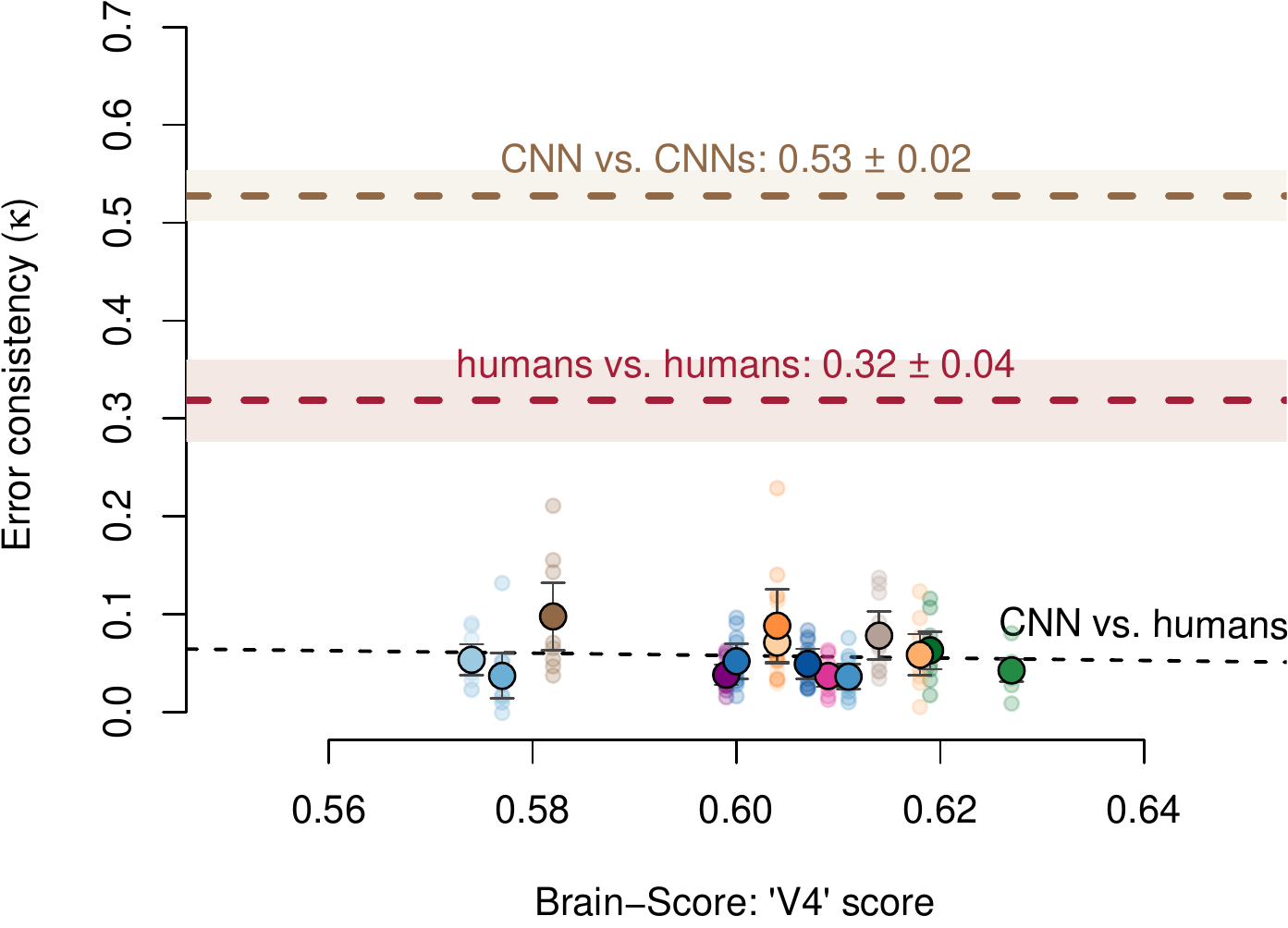}
    \end{subfigure}\hfill
    \vspace{\vSpaceI}
    
    \begin{subfigure}{\figwidthIII}
        \centering
        \includegraphics[width=\linewidth]{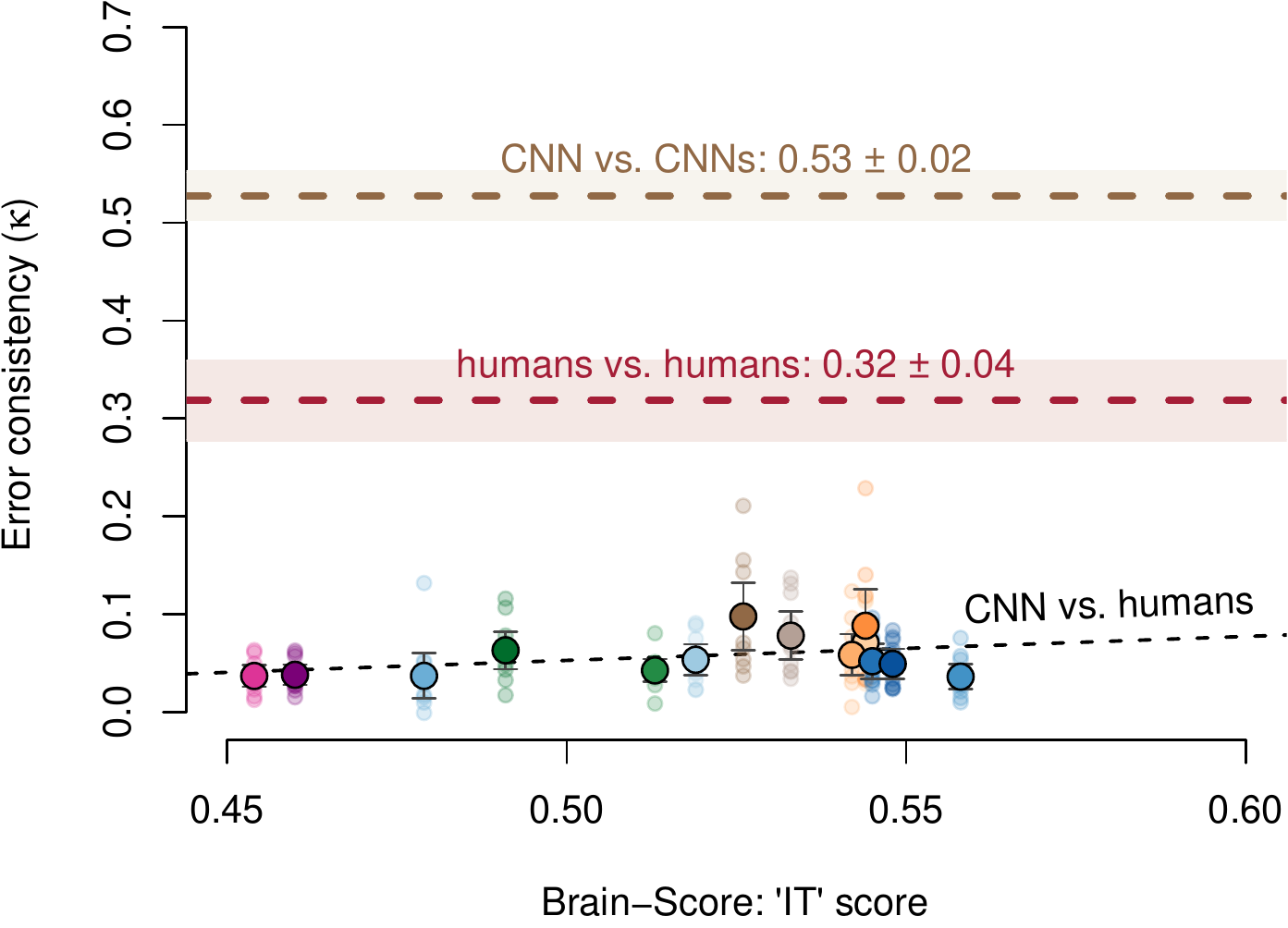}
    \end{subfigure}\hfill
    \vspace{\vSpaceI}
    \begin{subfigure}{\figwidthIV}
        \centering
        \includegraphics[width=\linewidth]{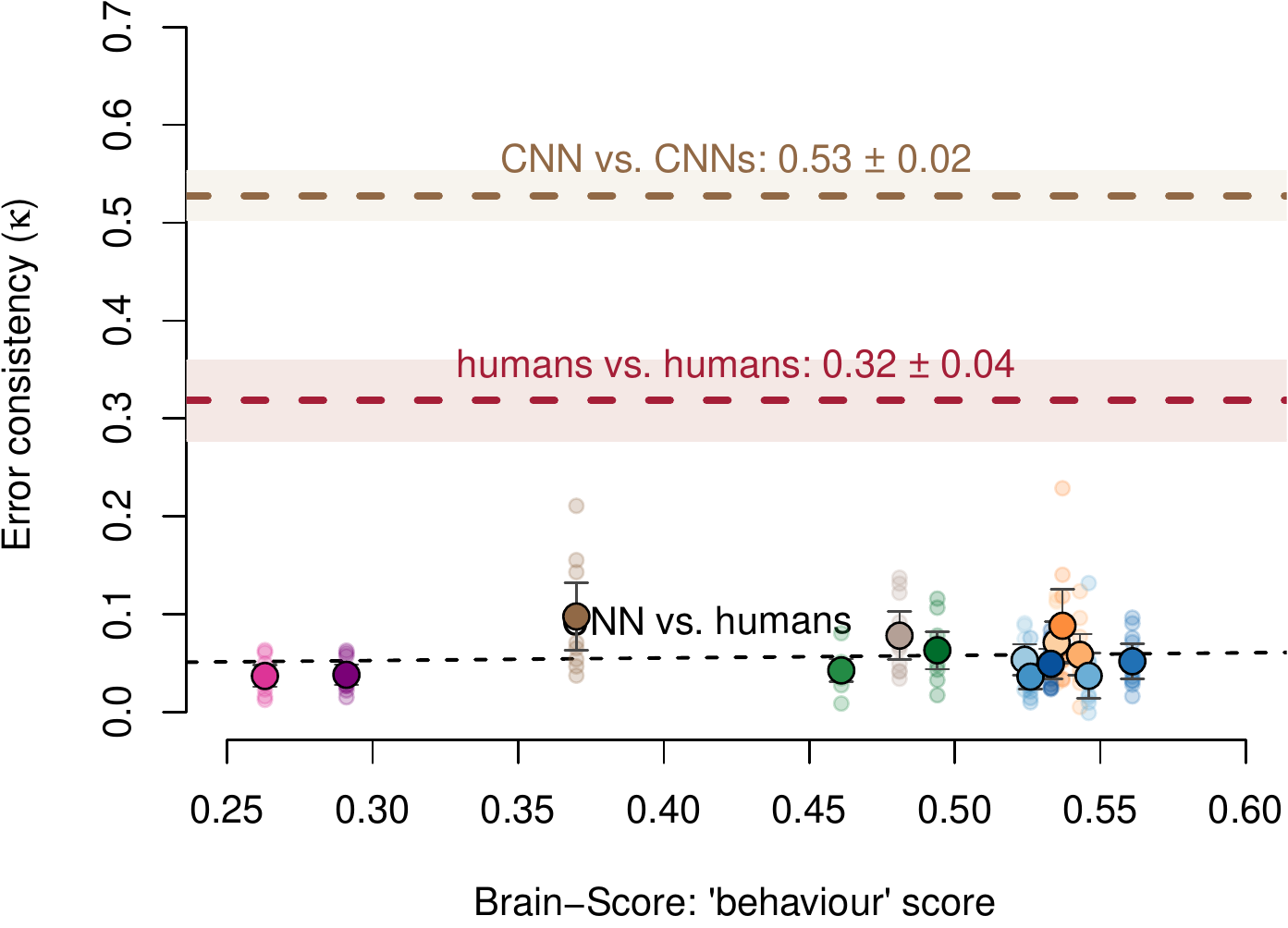}
    \end{subfigure}\hfill
    \vspace{\vSpaceI}
    \caption{Error consistency vs. \texttt{Brain-Score} metrics for PyTorch models, ``edge'' stimuli.}
    \label{fig:app_brainscore_edges}
\end{figure}

\begin{figure}[h!]
    \begin{subfigure}{\figwidthIII}
        \centering
        \includegraphics[width=\linewidth]{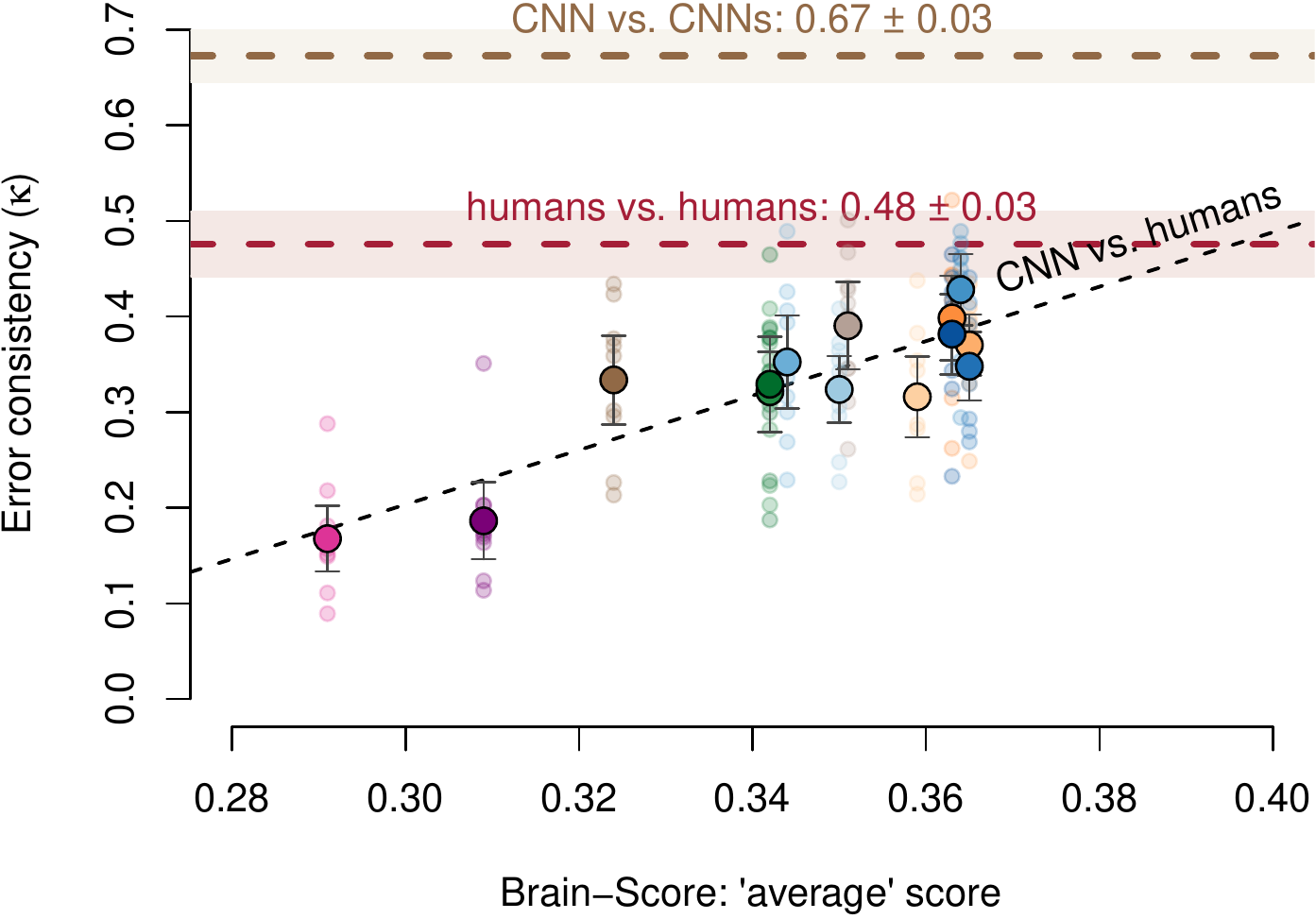}
    \end{subfigure}\hfill
    \vspace{\vSpaceI}
    \begin{subfigure}{\figwidthIV}
        \centering
        \includegraphics[width=\linewidth]{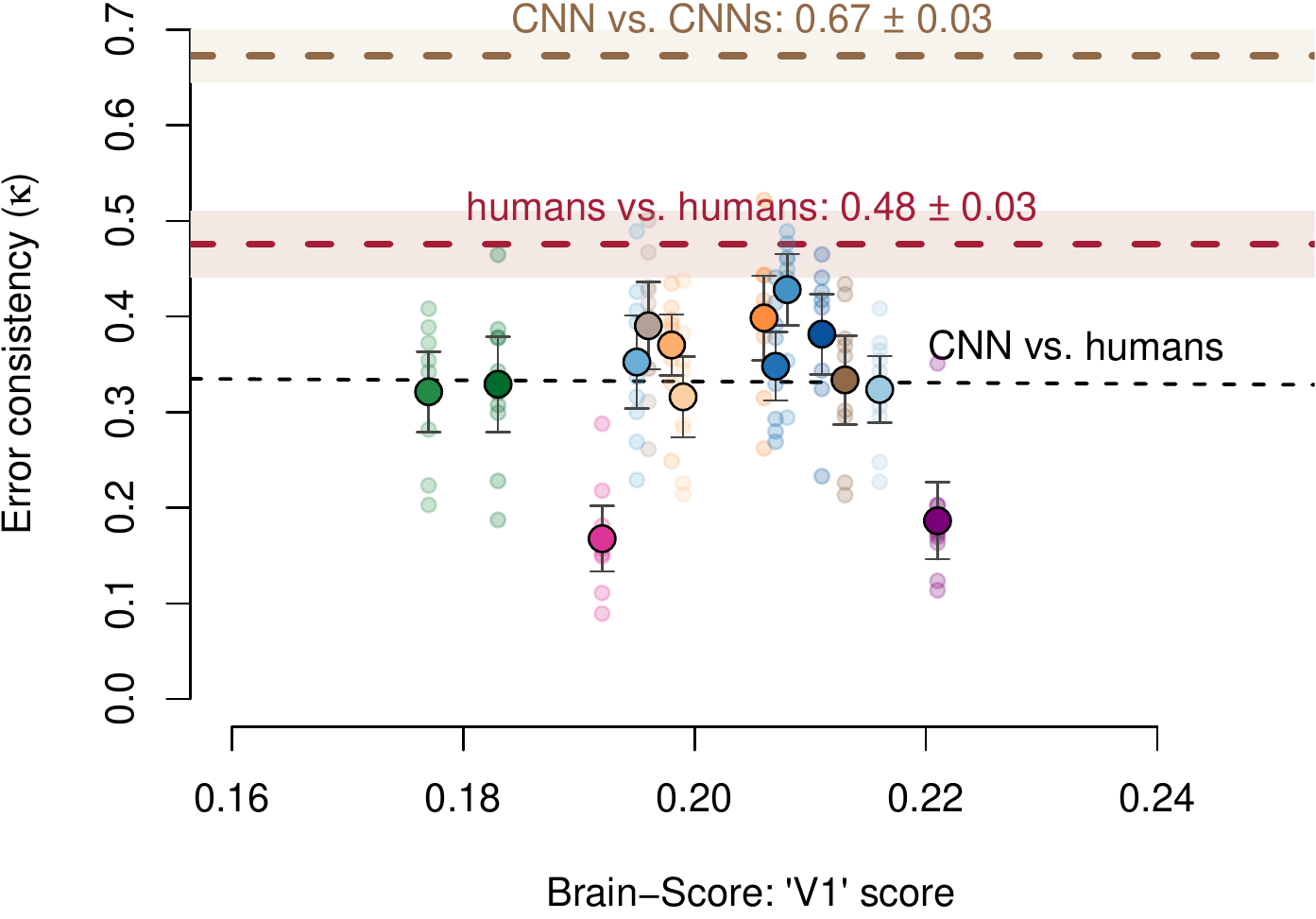}
    \end{subfigure}\hfill
    \vspace{\vSpaceI}

    \begin{subfigure}{\figwidthIII}
        \centering
        \includegraphics[width=\linewidth]{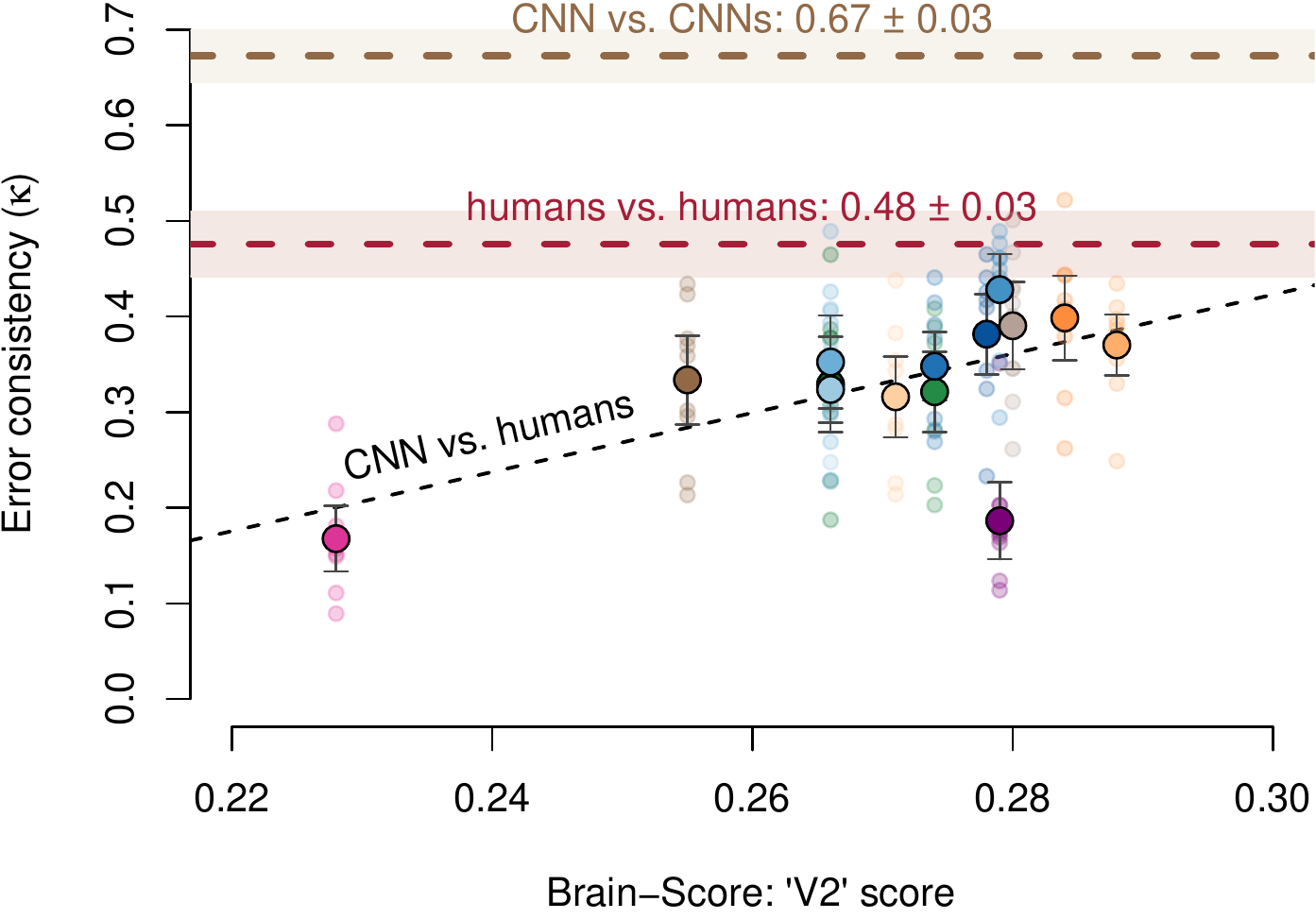}
    \end{subfigure}\hfill
    \vspace{\vSpaceI}
    \begin{subfigure}{\figwidthIV}
        \centering
        \includegraphics[width=\linewidth]{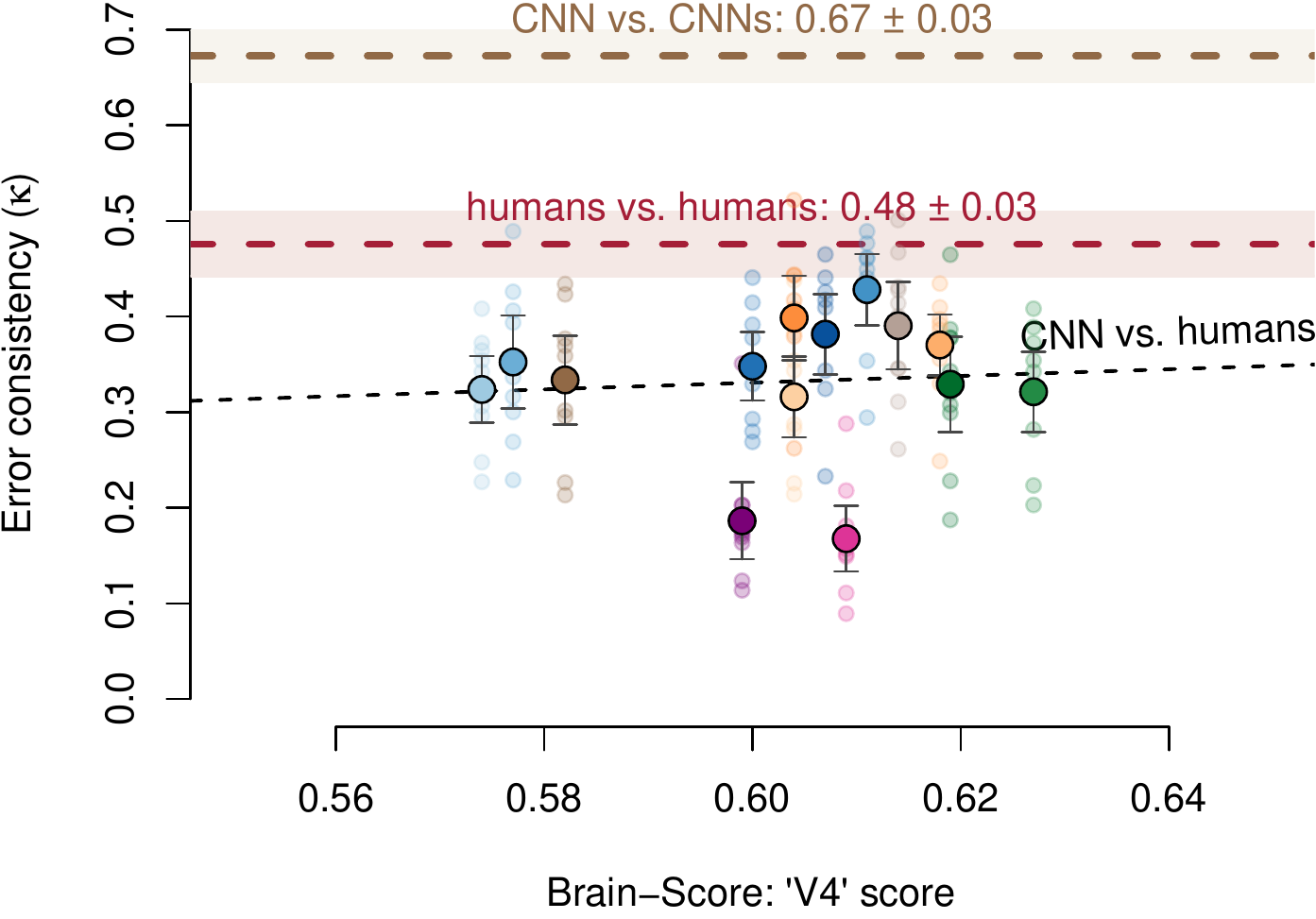}
    \end{subfigure}\hfill
    \vspace{\vSpaceI}

    \begin{subfigure}{\figwidthIII}
        \centering
        \includegraphics[width=\linewidth]{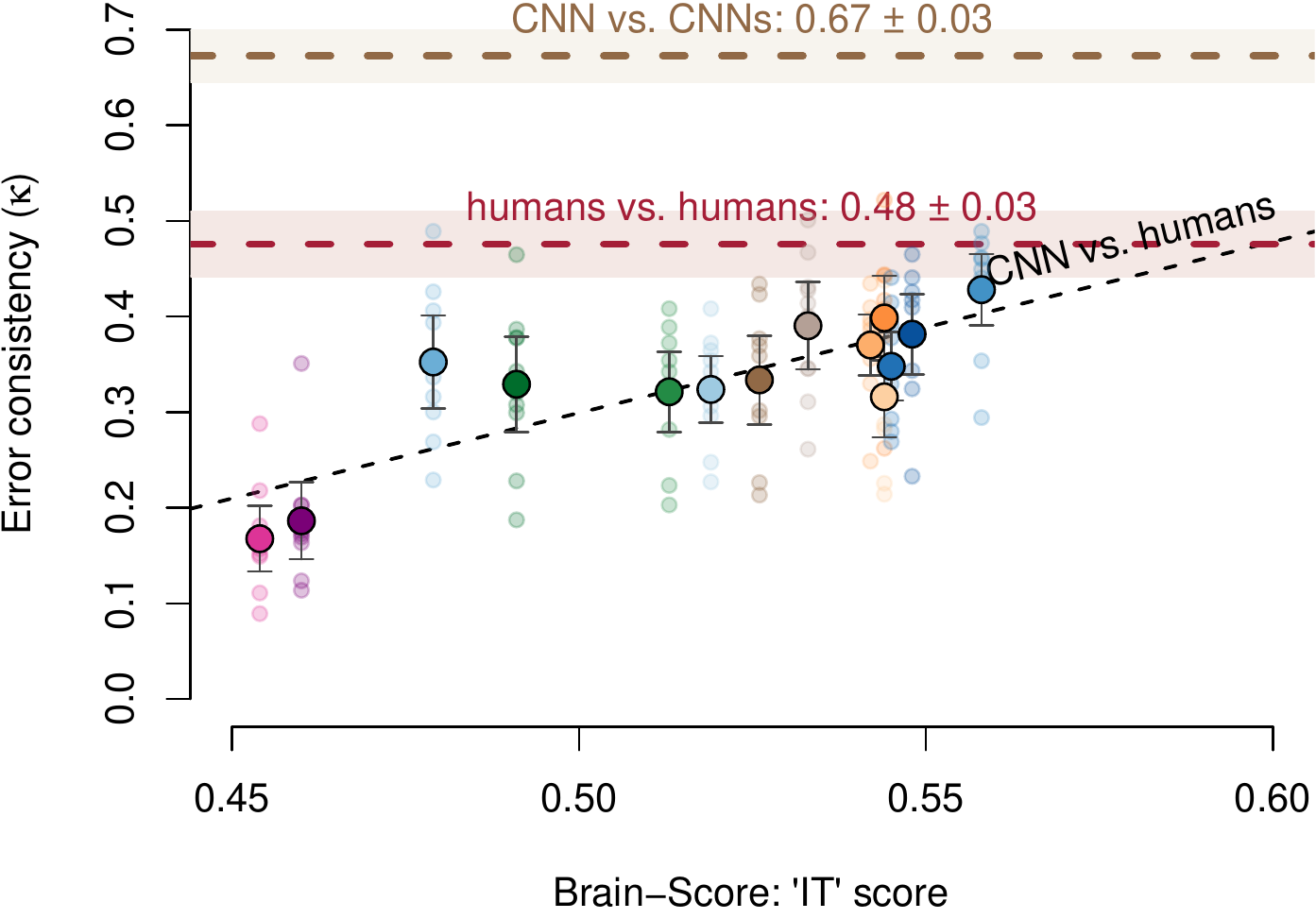}
    \end{subfigure}\hfill
    \vspace{\vSpaceI}
    \begin{subfigure}{\figwidthIV}
        \centering
        \includegraphics[width=\linewidth]{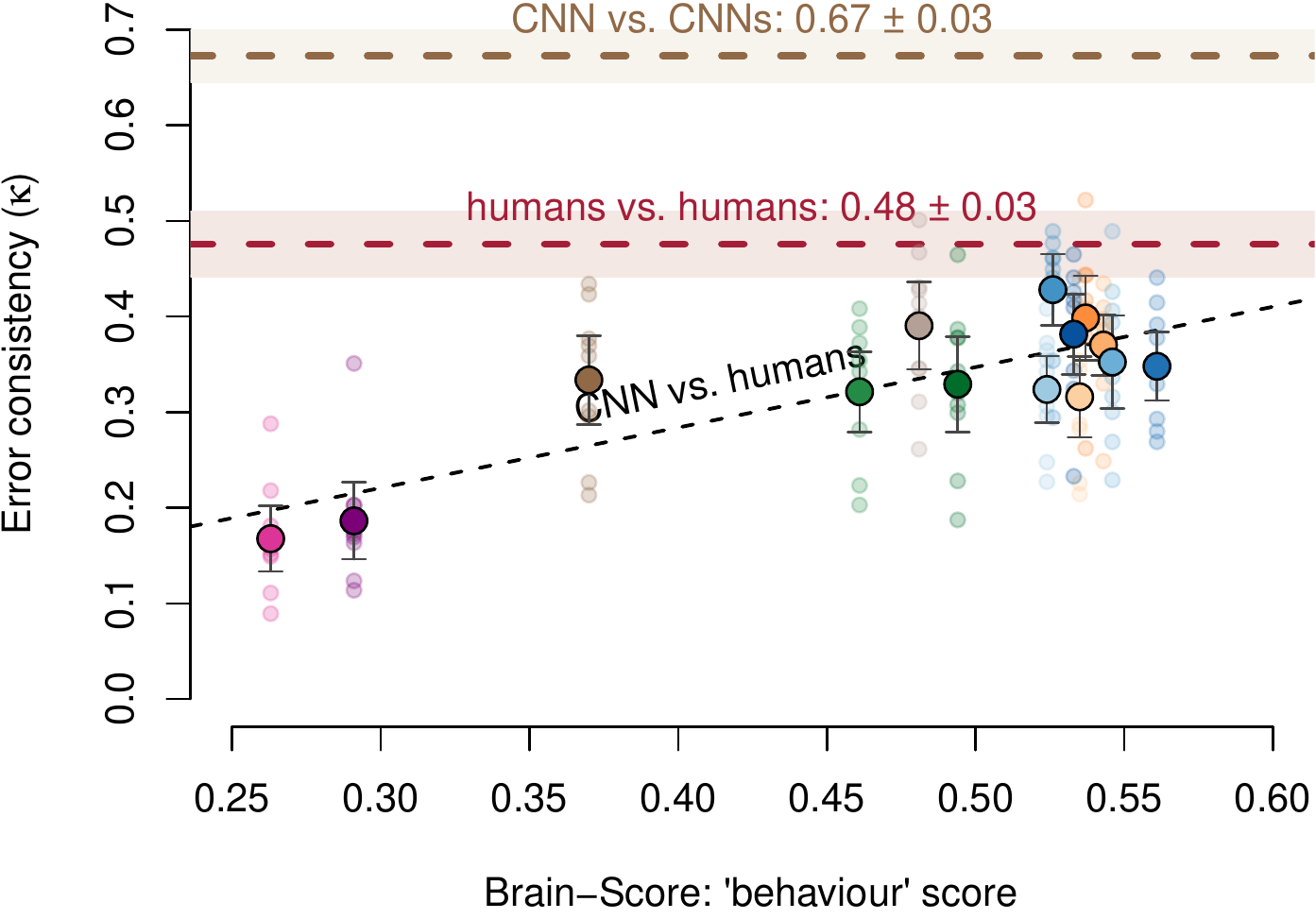}
    \end{subfigure}\hfill
    \vspace{\vSpaceI}
    \caption{Error consistency vs. \texttt{Brain-Score} metrics for PyTorch models, ``silhouette'' stimuli.}
    \label{fig:app_brainscore_silhouettes}
\end{figure}

\begin{figure}[h!]
    \begin{subfigure}{\figwidthIII}
        \centering
        \includegraphics[width=\linewidth]{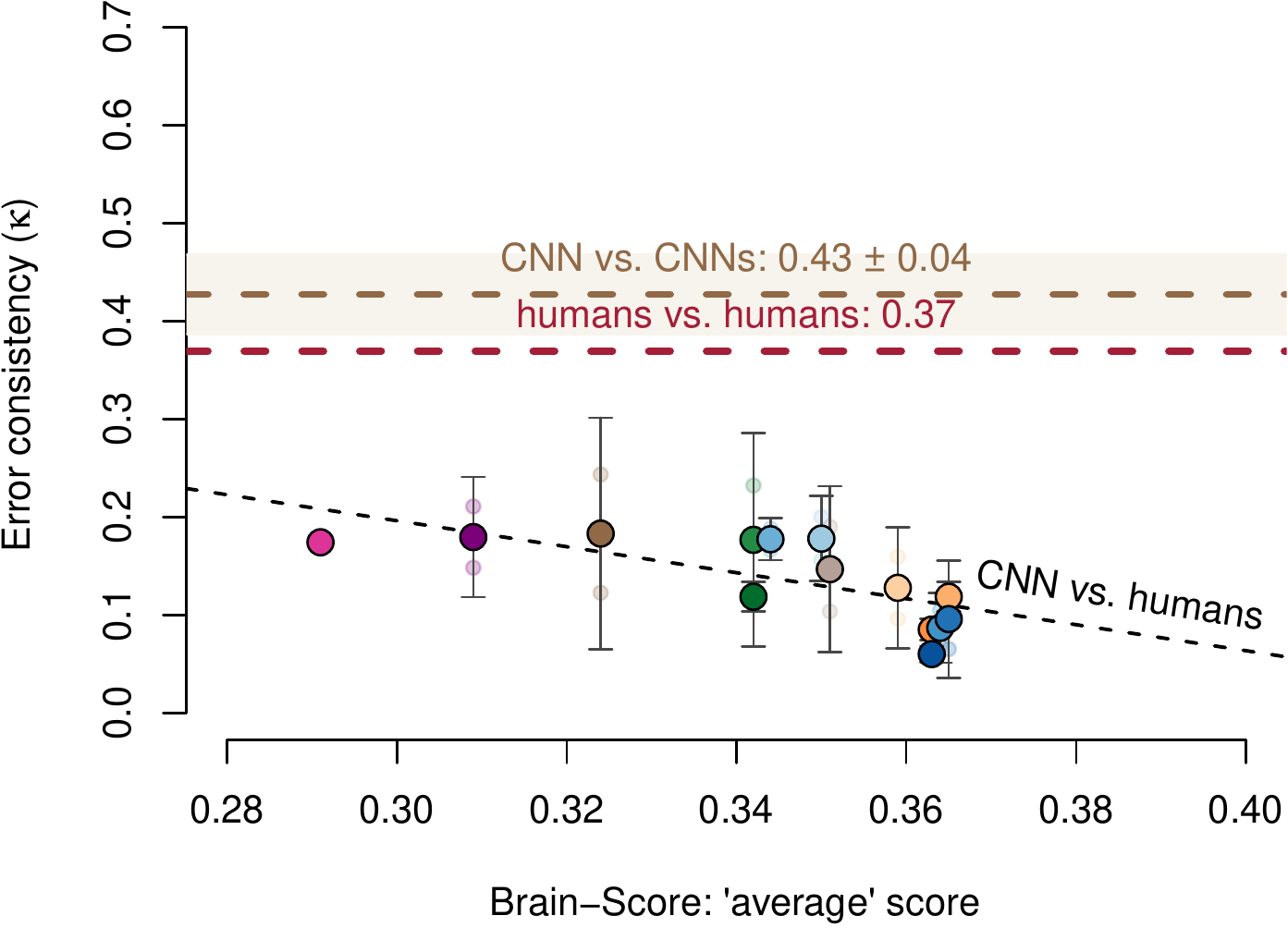}
    \end{subfigure}\hfill
    \vspace{\vSpaceI}
    \begin{subfigure}{\figwidthIV}
        \centering
        \includegraphics[width=\linewidth]{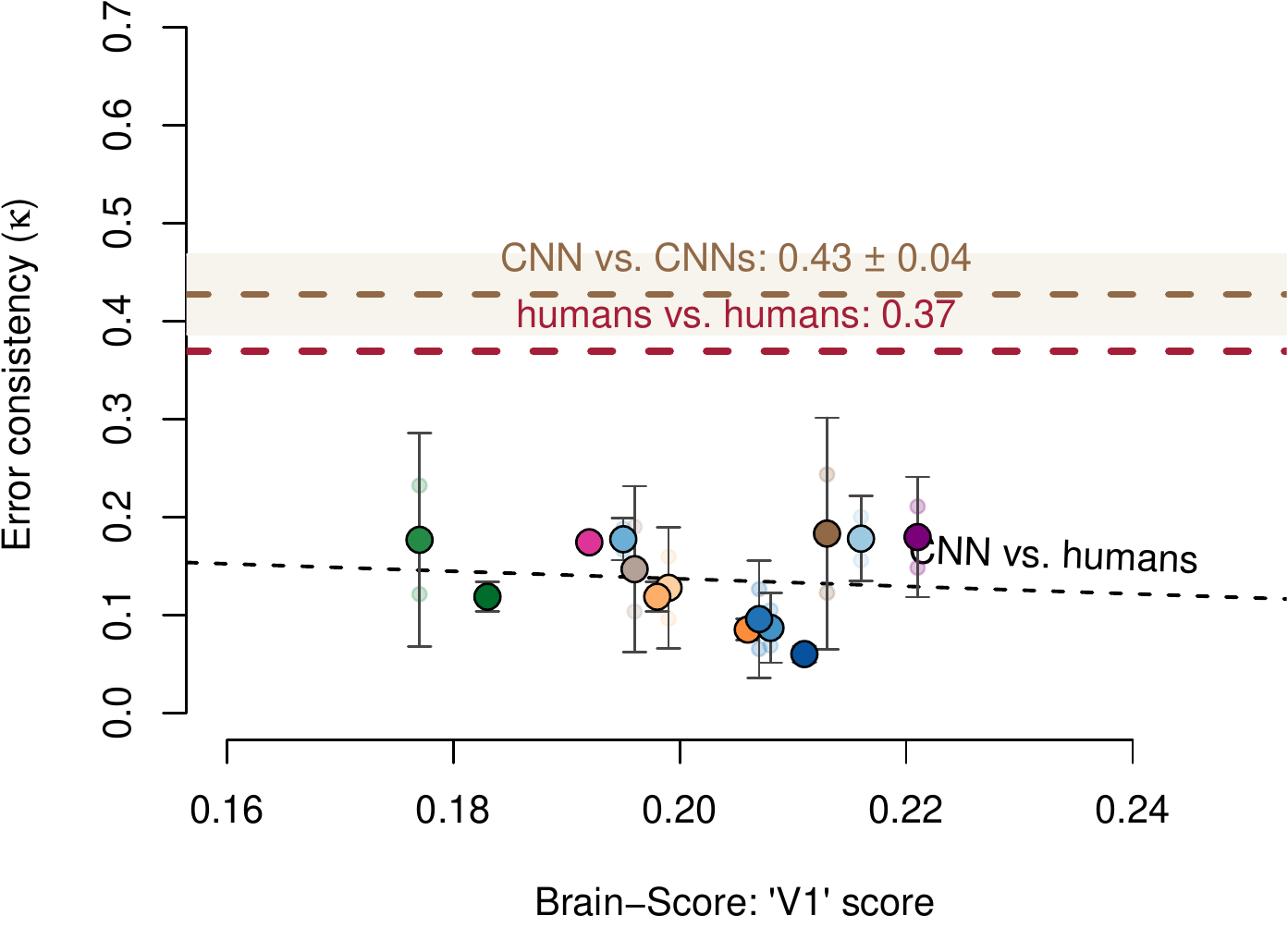}
    \end{subfigure}\hfill
    \vspace{\vSpaceI}

    \begin{subfigure}{\figwidthIII}
        \centering
        \includegraphics[width=\linewidth]{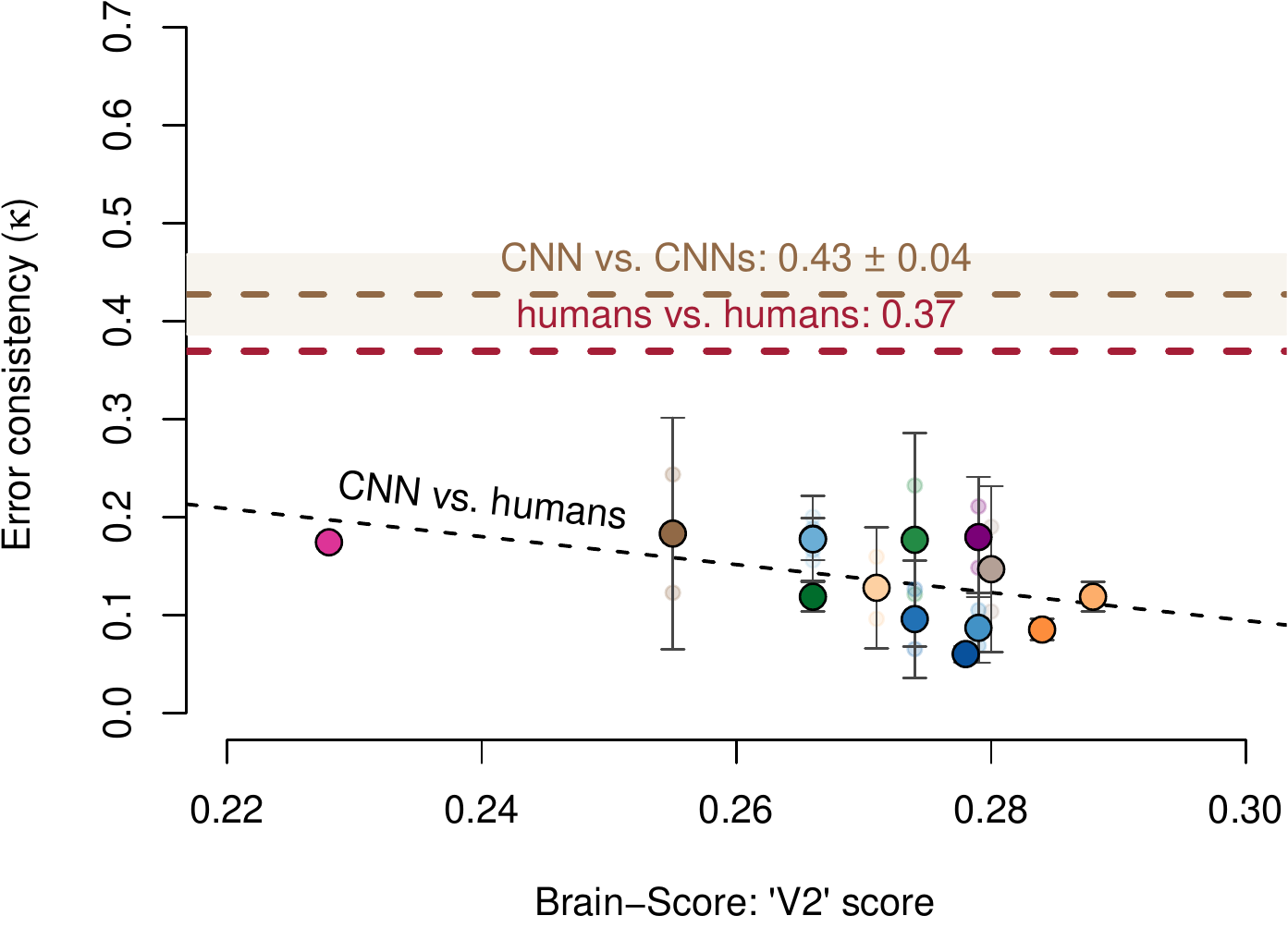}
    \end{subfigure}\hfill
    \vspace{\vSpaceI}
    \begin{subfigure}{\figwidthIV}
        \centering
        \includegraphics[width=\linewidth]{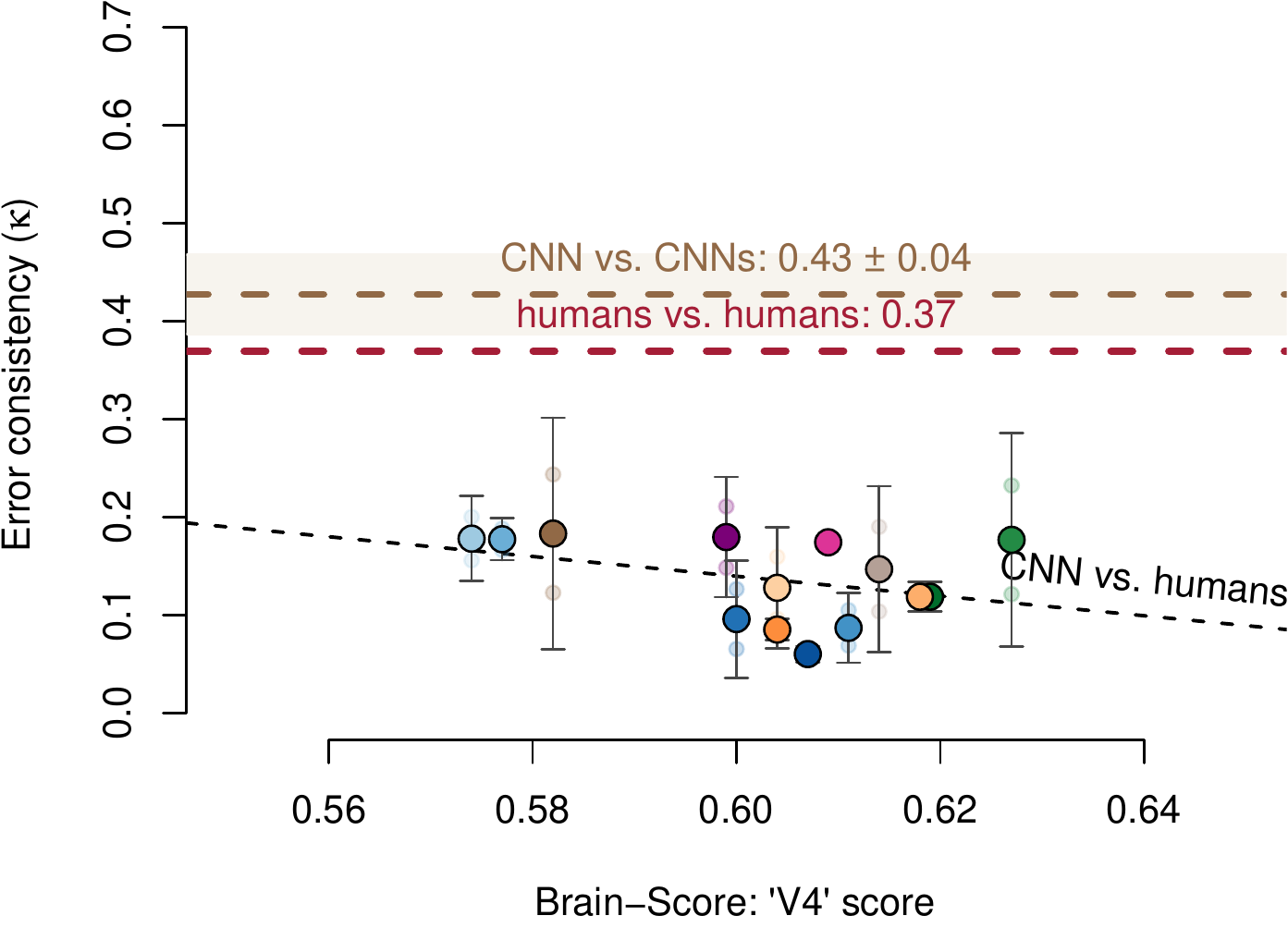}
    \end{subfigure}\hfill
    \vspace{\vSpaceI}

    \begin{subfigure}{\figwidthIII}
        \centering
        \includegraphics[width=\linewidth]{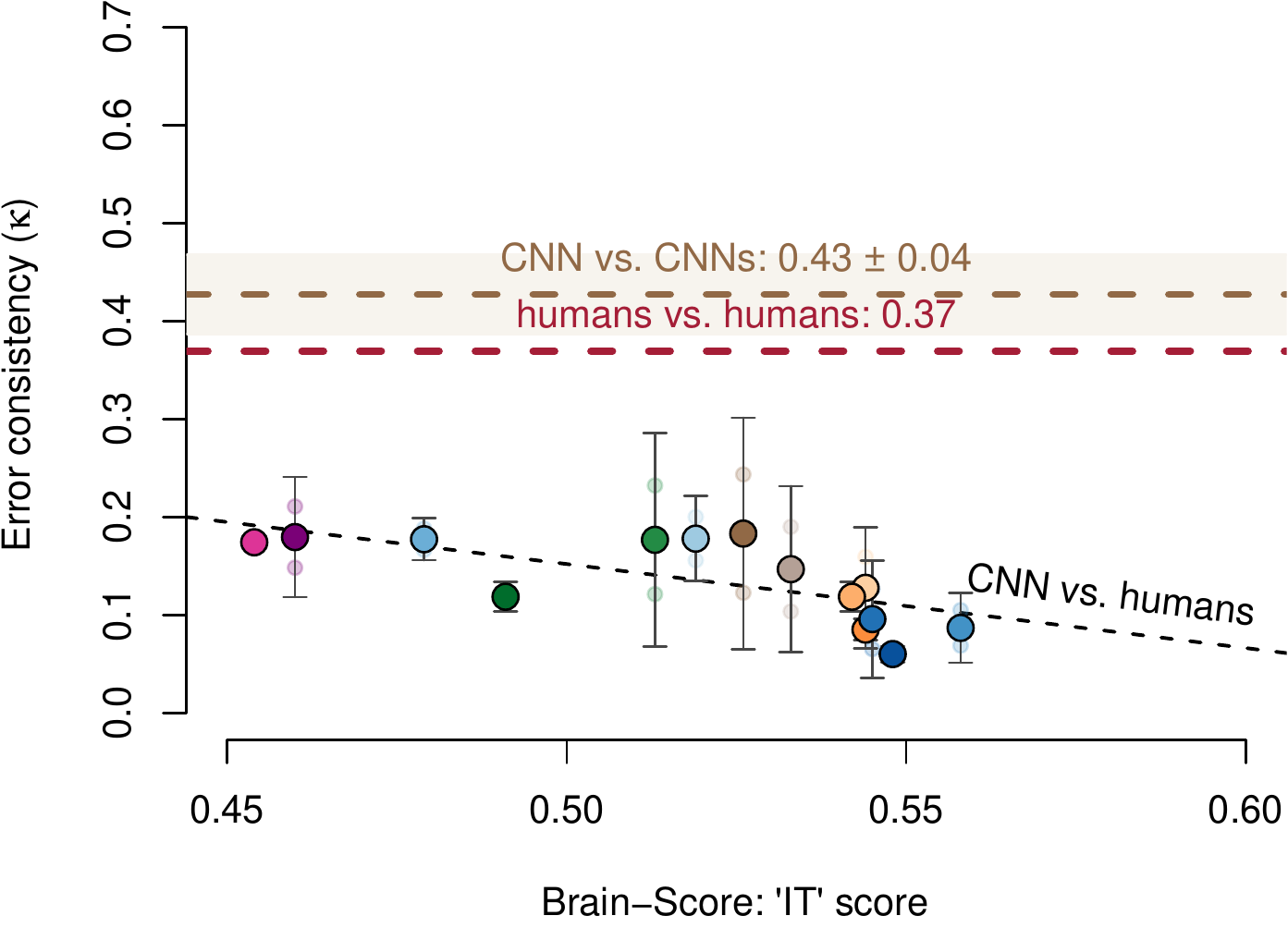}
    \end{subfigure}\hfill
    \vspace{\vSpaceI}
    \begin{subfigure}{\figwidthIV}
        \centering
        \includegraphics[width=\linewidth]{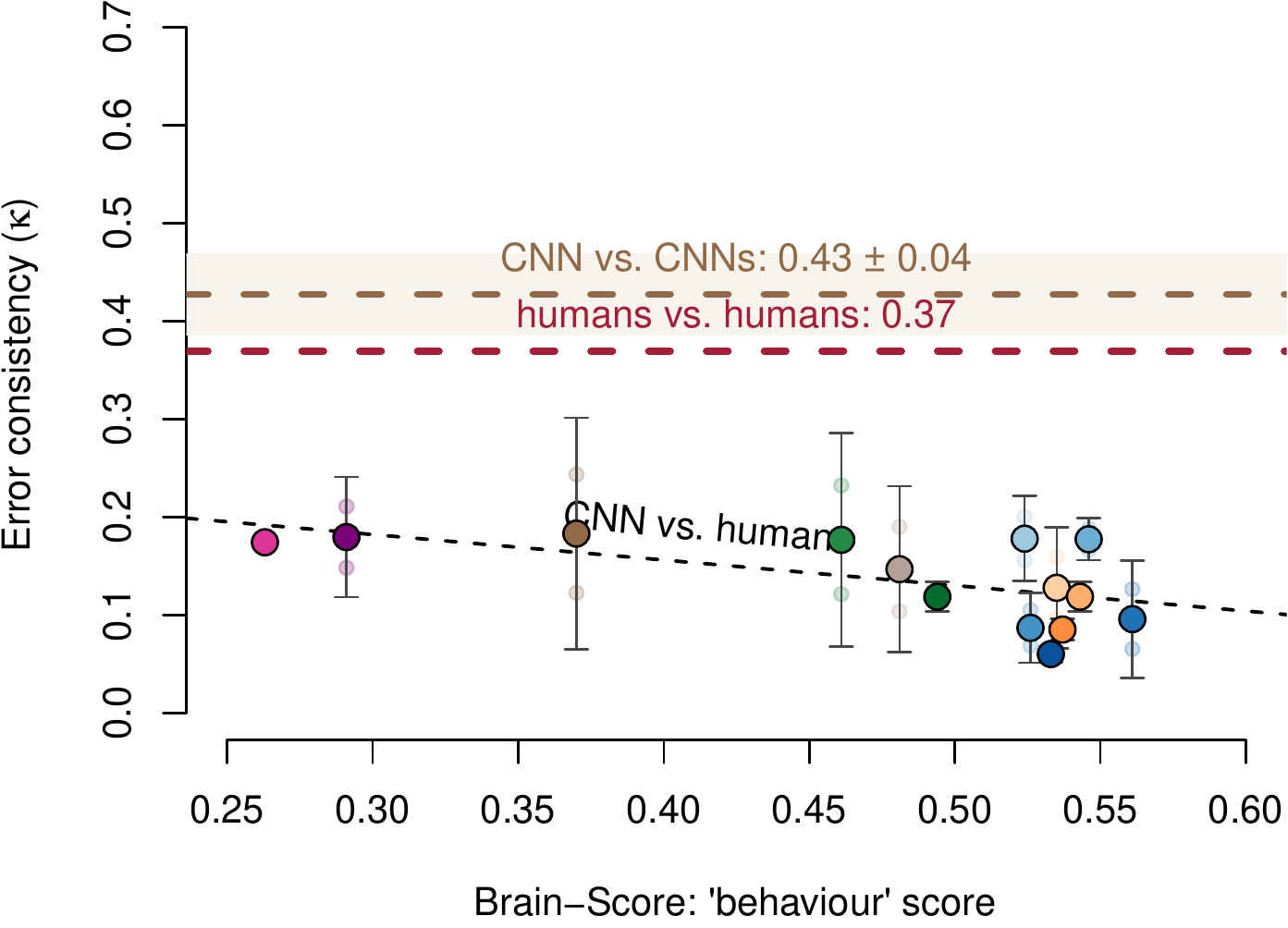}
    \end{subfigure}\hfill
    \vspace{\vSpaceI}
    \caption{Error consistency vs. \texttt{Brain-Score} metrics for PyTorch models, ``ImageNet'' stimuli.}
    \label{fig:app_brainscore_ImageNet}
\end{figure}

\end{document}